%% file: main_arxiv.tex
\documentclass[11pt]{article}

\usepackage[margin=1in]{geometry}
\usepackage{mathpazo}
\usepackage[backend=biber,style=alphabetic,natbib=true,maxnames=99,maxcitenames=2,minalphanames=3]{biblatex}
\input{header}

\title{Editing a classifier by rewriting its prediction rules}
\author{
    Shibani Santurkar\thanks{Equal contribution.} \\
  MIT \\
  \texttt{shibani@mit.edu} \\
  \and
  Dimitris Tsipras\footnotemark[1] \\
  MIT \\
  \texttt{tsipras@mit.edu} \\
  \and
  Mahalaxmi Elango \\
  MIT \\
  \texttt{melango@mit.edu} \\
  \and
  David Bau \\
  MIT \\
  \texttt{davidbau@mit.edu} \\
  \and
  Antonio Torralba \\
  MIT \\
  \texttt{torralba@mit.edu} \\
  \and
  Aleksander Madry \\
  MIT \\
  \texttt{madry@mit.edu} \\
  }
\date{}

\begin{document}
\maketitle

\begin{abstract}
\noindent
\input{sections/abstract.tex}
\end{abstract}

\section{Introduction}
\input{sections/introduction.tex}

\section{A toolkit for editing prediction rules}
\label{sec:editing}
\input{sections/editing.tex}

\section{Does editing work in practice?}
\label{sec:realworld}
\input{sections/real.tex}

\section{Large-scale synthetic evaluation}
\label{sec:discovery}
\input{sections/discovery.tex}
\label{sec:performance}
\label{sec:eval}
\input{sections/eval.tex}

\section{Beyond editing: Probing model behavior via counterfactuals}
\label{sec:concepts}
\input{sections/analysis.tex}

\section{Related work}
\label{sec:related}
\input{sections/related.tex}

\section{Conclusion}
\label{sec:conclusion}
\input{sections/conclusion}

\section{Acknowledgements}
\input{sections/ack}

\printbibliography
\clearpage

\appendix

\section{Experimental Setup} 
\input{sections/setup.tex}

\section{Additional Experiments} 
\input{sections/appendix}

\end{document}

%% file: header.tex
\usepackage{graphicx}
\usepackage{psfrag}
\usepackage{amsmath}
\usepackage{amsfonts}
\usepackage{verbatim}
\usepackage{optidef}
\usepackage{mathrsfs}
\usepackage{hyperref}
\usepackage{color}
\usepackage[ruled,vlined]{algorithm2e}
\usepackage[utf8]{inputenc}
\usepackage[T1]{fontenc}

\usepackage{subcaption}
\usepackage{booktabs}
\usepackage{tabularx}
\usepackage{multirow}

\usepackage[textsize=scriptsize]{todonotes}

\newcommand{\class}[1]{``#1''}
\newcommand{\ww}{\class{wooden wheel}}
\newcommand{\kstar}{k^\ast}
\newcommand{\vstar}{v^\ast}

%% file: sections/abstract.tex
We present a methodology for modifying the behavior of a classifier by
\emph{directly rewriting} its prediction rules.\footnote{Our code is available at 
\url{https://github.com/MadryLab/EditingClassifiers}.}
Our approach requires virtually no additional data collection and can be 
applied
to a variety of settings, including adapting a model to new environments, and
modifying it to ignore spurious features.


%% file: sections/introduction.tex
At the core of machine learning is the ability
to automatically discover prediction rules from raw
data.
However, there is mounting evidence that not all of these rules are  
reliable~\citep{torralba2011unbiased,beery2018recognition,shetty2019not,agarwal2020towards,xiao2020noise,bissoto2020debiasing,geirhos2020shortcut}.
In particular, some rules could be based on biases in the 
training data: e.g., learning to associate cows with grass since they are 
typically depicted on pastures~\citep{beery2018recognition}.
While such prediction rules may be useful in some scenarios, they will be 
irrelevant or misleading in others.
This raises the question:
\begin{center}
\emph{
    How can we most effectively modify the way in which a  given model makes its predictions?
}
\end{center}
The canonical approach for performing such post hoc modifications
 is to intervene at the data level. For example, by gathering additional data 
 that 
 better reflects the real world (e.g., 
images of cows on the beach) and 
then using it to further train the model.
Unfortunately, collecting such data can be challenging: how do we get cows
to pose for us in a variety of environments?
Furthermore, data collection is ultimately a very indirect way of 
specifying
the intended model behavior.
After all, even when data has been carefully curated to reflect a given 
real-world task, models still end up learning unintended prediction rules from 
it~\citep{ponce2006dataset,torralba2011unbiased,tsipras2020from,beyer2020are}.

\subsection*{Our contributions}
The goal of our work is to develop a toolkit that enables users to
\emph{directly modify} the prediction rules learned by an (image) 
classifier, as opposed to doing so implicitly via the data.
Concretely:

\paragraph{Editing prediction rules.}
We build on the recent work of \citet{bau2020rewriting} to develop a method for
modifying a classifier's prediction rules with
essentially \emph{no} additional data collection (Section \ref{sec:editing}).
At a high level, our method enables the user to modify the weight of a layer 
so
that the latent representations of a specific concept (e.g., snow) map to the
representations of another (e.g., road).
Crucially, this allows us to change the behavior of the classifier on all 
occurrences of
that concept, beyond the specific examples (and the corresponding 
classes) used in the editing process. 

\begin{figure}[t]
	\centering
	\begin{subfigure}[b]{\textwidth}
		\centering
		\includegraphics[width=0.95\textwidth]{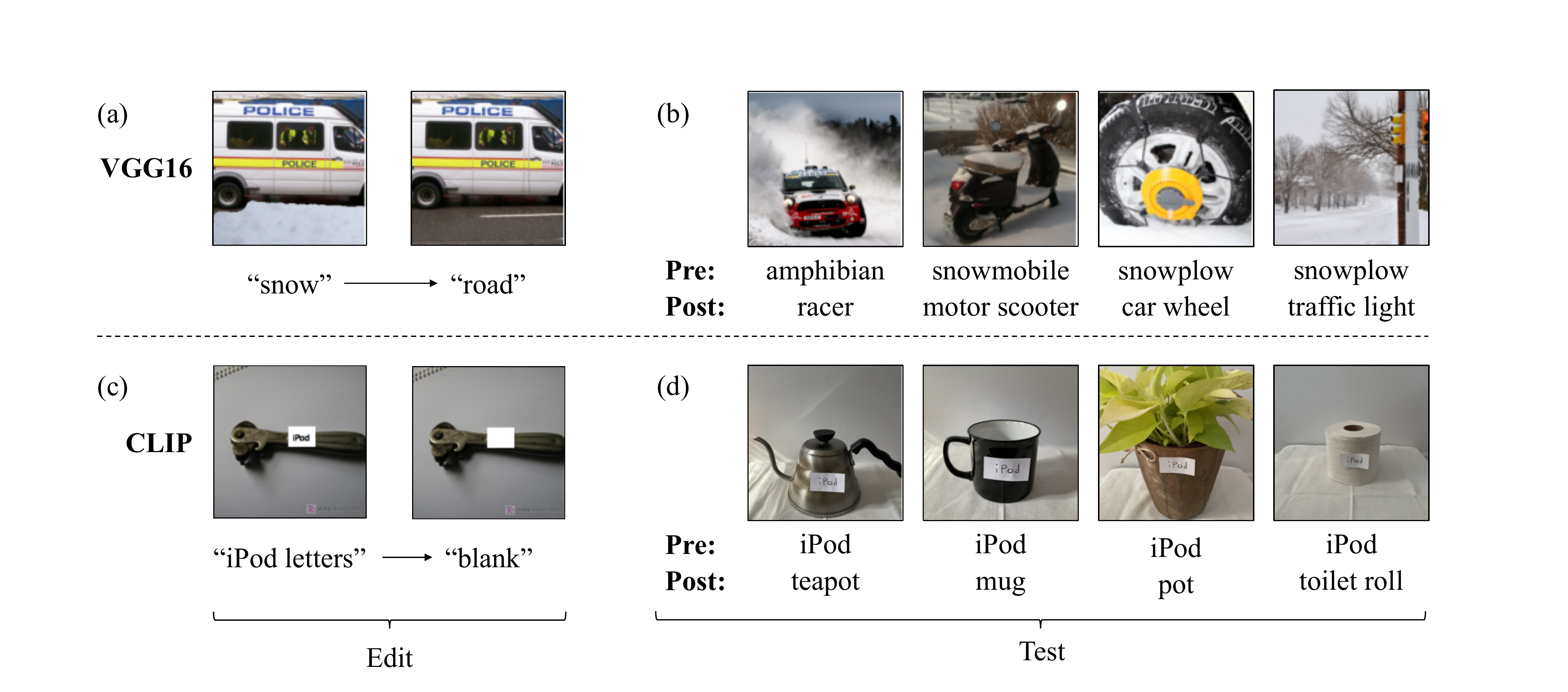}
	\end{subfigure}\hfil
	\caption{Editing prediction rules in pre-trained classifiers using a 
	\emph{single} 
	exemplar.  (a) We edit a 
	VGG16 ImageNet classifier to map the representation of the 
	concept \class{snow} to that of \class{asphalt road}.  (b) This edit 
	corrects classification errors 
	on snowy scenes corresponding to various classes.  (c) We edit a
    CLIP~\citep{radford2021learning} model such that the text 
	``iPod'' maps to a blank area.  (d) This change makes the model robust to 
	the typographic attacks from \citet{goh2021multimodal}. }
	\label{fig:intro_fig}
\end{figure}

\paragraph{Real-world scenarios.}
We demonstrate our approach in two scenarios motivated by real-world
applications (Section~\ref{sec:realworld}).
First, we focus on adapting an ImageNet classifier to a new environment: 
recognizing vehicles on snowy roads.
Second, we consider the recent ``typographic attack'' of
\citet{goh2021multimodal} on a zero-shot CLIP~\citep{radford2021learning}
classifier: attaching a piece of paper with ``iPod'' written on it to various
household items causes them to be incorrectly classified as ``iPod.''
In both settings, we find that our approach enables us to significantly 
improve model performance, using only a \emph{single, synthetic}  
example to perform the edit---cf.
Figure~\ref{fig:intro_fig}.

\paragraph{Large-scale synthetic evaluation.}
To evaluate our method at scale, 
we develop an automated
pipeline to generate a suite of varied test cases (Section~\ref{sec:discovery}).
Our pipeline revolves around identifying specific concepts (e.g., ``road" or 
``pasture'') in an existing
dataset using pre-trained instance segmentation models and then modifying
them using style transfer~\citep{gatys2016image} (e.g., to create ``snowy
road'').
We find that our editing methodology is able to consistently correct 
a significant fraction of model failures induced by these transformations.
In contrast, standard fine-tuning approaches are unable to do so given the same
data, often causing more errors than they are fixing.

\paragraph{Probing model behavior with counterfactuals.}
Moving beyond model editing, we observe that the concept-transformation 
pipeline we developed can also be viewed as a scalable approach for generating 
image 
counterfactuals.
In Section~\ref{sec:concepts}, we demonstrate how such 
counterfactuals can be
useful to gain insights into how a given model makes its predictions and
pinpoint certain spurious correlations that it has picked up.

%% file: sections/editing.tex
It has been widely observed that models pick up various context-specific
correlations in the data---e.g., using the presence of \class{road} or a \class{wheel} to predict
\class{car} (cf. Section~\ref{sec:concepts}).
Such unreliable \emph{prediction rules} (dependencies of 
predictions on 
specific input concepts) could 
hinder models when they encounter novel 
environments  (e.g., snow-covered roads), and confusing or adversarial test 
conditions (e.g., cars with wooden wheels).
Thus,  a model designer might want to modify these rules
before deploying their model.

The canonical approach to modify a classifier post hoc is to  collect 
additional data that captures the desired deployment scenario, and use
it to retrain the model.
However, even setting aside the challenges of data collection, it is not
obvious a priori how much of an effect such retraining (e.g., via fine-tuning) 
will
 have.
For instance, if we fine-tune our model on \class{cars} with wooden 
wheels, will it now recognize \class{scooters} or 
\class{trucks} with such wheels?

The goal of this work is to instead develop a more \emph{direct} way to 
modify a 
model's
behavior:  rewriting its prediction rules in a targeted manner.
For instance, in our previous example, we would ideally be able to modify the 
classifier to correctly recognize \emph{all} vehicles with wooden wheels by 
simply teaching it to treat \emph{any} wooden wheel as it would a
standard one.
Our approach is able to do exactly this.
However, before describing this approach  
(Section~\ref{sec:classifier_rewriting}), we
first provide a brief overview of recent work by~\citet{bau2020rewriting} which
forms its basis.

\subsection{Background: Rewriting generative models}
\label{sec:gan_rewriting}
\citet{bau2020rewriting} developed an approach for rewriting a deep generative 
model: specifically, enabling a user to replace all 
occurrences 
of one selected
object (say, \class{dome}) in the generated images with another 
(say, \class{tree}), without changing the model's behavior in other contexts.
Their approach is motivated by the observation that, using a handful of 
example images, we can identify a vector in the model's representation space 
that encodes a specific 
high-level concept~\citep{kim2018interpretability,bau2020rewriting}.
Building on this, \citet{bau2020rewriting} treat each layer of the 
model as an \emph{associative memory}, which maps such a concept vector 
at each spatial location in its input  (which we 
will
refer to as the \emph{key}) to another concept vector in its output 
(which we will 
call the \emph{value}).

In the simplest case, one can think of a linear layer with weights $W \in 
\mathbb{R}^{mxn}$ 
transforming the key $k \in \mathbb{R}^n$  to the value $v\in 
\mathbb{R}^m$.
In this setting, \citet{bau2020rewriting} formulate the rewrite operation as 
modifying 
the 
layer weights
from 
$W$ 
to $W'$ so that $\vstar =
W' \kstar$, where $\kstar$ corresponds to the old concept that we want to 
replace, and $\vstar$ to the new concept.
For instance, to replace \class{domes} with \class{trees} in the 
generated images, we
would modify the layer so that the key $\kstar$ for \class{dome} maps to the 
value $\vstar$ for \class{tree}.
Consequently, when this value is fed into the downstream 
layers
of the network it would result in a \emph{tree} in the 
final image.
Crucially, this update should change the model's behavior for \emph{every }
instance of the  concept encoded in $\kstar$---i.e., all ``domes'' in the 
images should now be ``trees''.

To extend this approach to typical deep generative models, two challenges 
remain: (1) handling
non-linear layers, and (2) ensuring that 
the edit doesn't significantly 
hurt model behavior on other concepts.
With these considerations in mind, \citet{bau2020rewriting} 
propose making the following rank-one updates to the parameters $W$ of an 
arbitrary non-linear layer $f$:
\begin{equation}
    \min_\Lambda\quad  \sum_{(i,j) \in S} \left\|\vstar_{ij} 
    -f(\kstar_{ij}; W')\right\| \qquad
    \text{s.t.}\quad  W' = W + \Lambda (C^{-1}d)^\top.
    \label{eq:final}
\end{equation}
Here, $S$ denotes the set of spatial locations in representation space for a 
single image
corresponding to the concept of interest, $d$ is the top eigenvector of 
the keys $\kstar_{ij}$ at locations $(i,j)\in S$ and $C=\sum_d k_d
{k_d}^\top$ captures the second-order statistics for other keys $k_d$.
Intuitively, the goal of this update is to modify the layer 
parameters to rewrite the desired key-value mapping in the most minimal way.
We refer the reader to \citet{bau2020rewriting} for further details.

\begin{figure}[!t]
	\centering
	\begin{subfigure}[b]{0.64\textwidth}
		\centering
		\includegraphics[height=5cm]{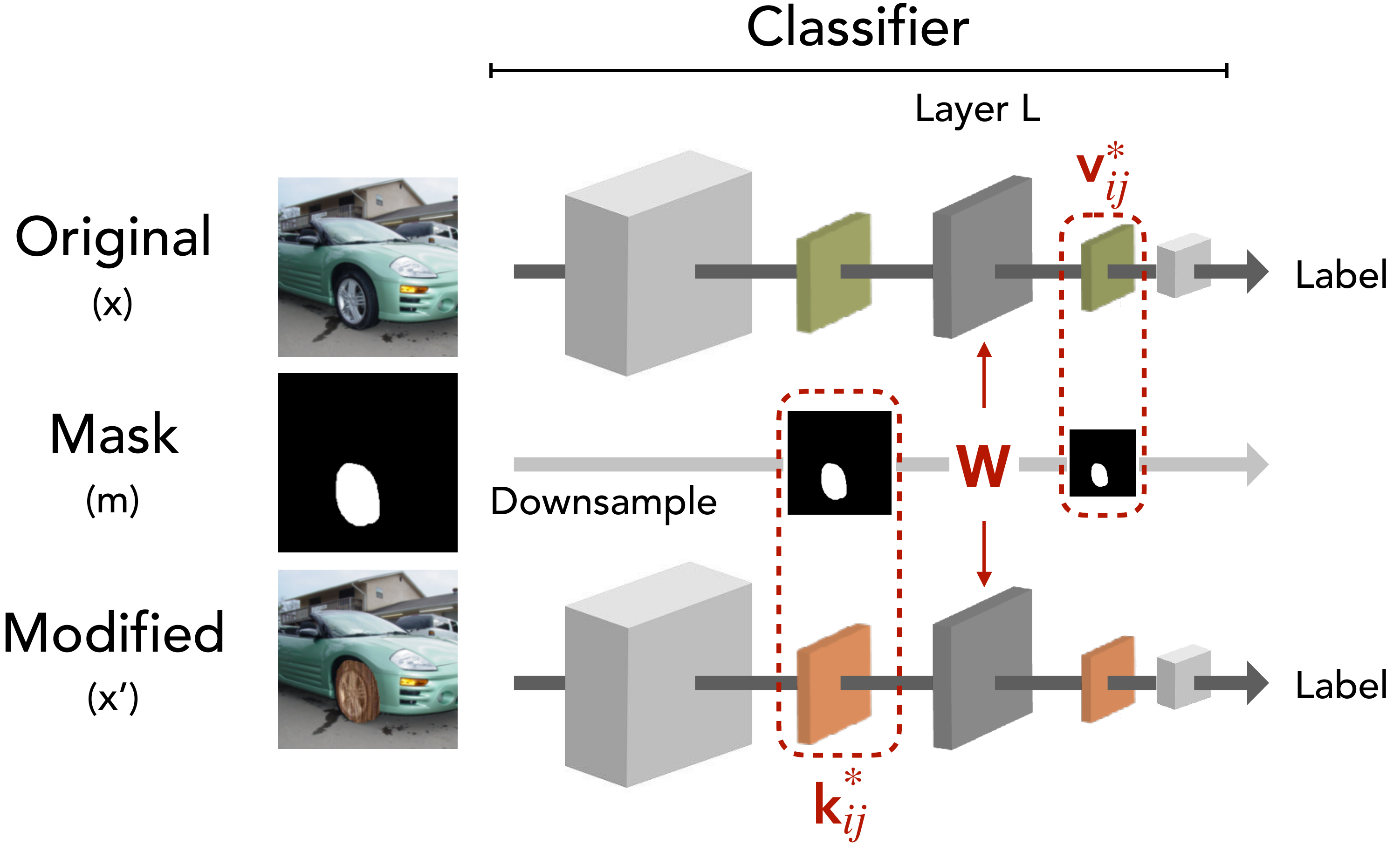}
		\caption{Edit}
	\end{subfigure}
	\hfil
	\begin{subfigure}[b]{0.33\textwidth}
		\centering
		\includegraphics[height=4.4cm]{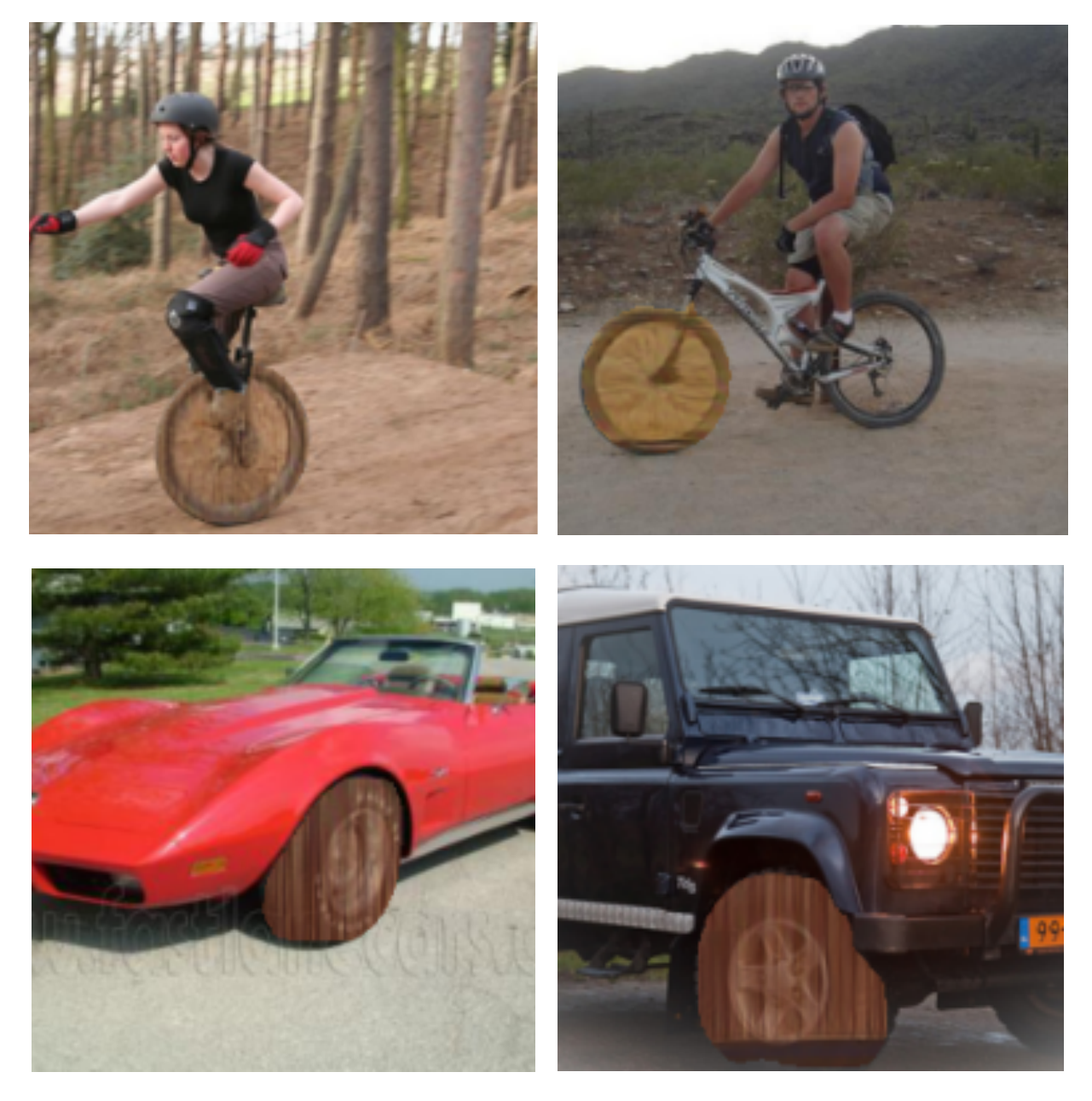}
		\caption{Test}
		\label{fig:editing_pipeline_test}
	\end{subfigure}
	\caption{Overview of our pipeline for directly editing the prediction-rules of 
	a classifier. 
	The edit in (a) seeks to modify the network to perceive wooden wheels 
	as standard ones, using a small set of exemplar images (say from class 
	\class{car}).
	To achieve this, we first obtain the keys $\kstar_{ij}$  corresponding to 
	the 
	new concept (here, \class{wooden wheel}), and the values $\vstar_{ij}$ 
	corresponding to the original concept (here, \class{standard wheel}) in the 
	input and output representation space of a layer $L$ respectively.
	We then update the weights $W$ of the layer to enforce 
	this new key-value association \eqref{eq:final}. (b) To test our method, we measure the improvement 
	in model performance on test instances  containing the 
	new concept---here, images of other
	vehicles with \class{wooden wheels}.}
	\label{fig:editing_pipeline}
\end{figure}

\subsection{Editing classifiers}
\label{sec:classifier_rewriting}
We now shift our attention to the focus of this work: editing classifiers.
To describe our approach, we use the task of enabling 
classifiers to detect vehicles with \class{wooden wheels} as a running 
example.
At a high level, we would like to apply the approach described in 
Section~\ref{sec:gan_rewriting} to modify a chosen 
(potentially
non-linear) layer $L$ of the network to rewrite the relevant key-value 
association.
But, we need to first determine what these relevant keys and values are.

Let us start with a single image $x$, say, from
class \class{car}, that contains the concept \class{wheel}.
Let the location of the \class{wheel} in the image be denoted by a 
binary mask $m$.\footnote{Such a  mask can either be obtained manually or
automatically via instance segmentation (cf.  Section~\ref{sec:discovery}).}
Then, say we have access to a transformed version of $x$---namely, 
$x'$---where the 
\class{car} has a \class{wooden wheel} (cf. Figure~\ref{fig:editing_pipeline}).
This $x'$ could be created by manually replacing the 
wheel, or by applying an automated procedure such as that in
Section~\ref{sec:discovery}.
In the rest of our study, we refer to a single ($x$, $x'$) pair as an 
\emph{exemplar}.

Intuitively, we want the classifier to perceive the \ww{} in the transformed
image $x'$ as it does the standard wheel in the original image $x$.
To achieve this, we must map the keys for wooden wheels to the 
value corresponding to their standard counterparts.
In other words, the relevant keys ($\kstar$)  correspond to the network's
representation of the concept in the \emph{transformed} image ($x'$) 
directly \emph{before} layer $L$.
Similarly, the relevant values ($\vstar$) that we want to map these keys to 
correspond 
to the network's representation of the concept in the \emph{original} images  
($x$)
directly \emph{after} layer $L$.
(The pertinent spatial regions in the representation space are simply 
determined by 
downsampling the mask to the appropriate dimensions.)
Finally, the model edit is performed by feeding the resulting key-value pairs
into the optimization problem \eqref{eq:final} to determine the updated layer
weights $W'$---cf. Figure~\ref{fig:editing_pipeline} for an illustration of the
overall process.
Note that this approach can be easily extended to use multiple exemplars 
$(x_k, x'_k)$ by simply expanding $S$ to include the union of relevant spatial 
locations (corresponding the concept of interest) across these exemplars.

%% file: sections/real.tex
To test the effectiveness of our approach, we start by considering two
scenarios motivated by real-world concerns: (i) adapting classifiers to 
handle novel weather conditions, and (ii) making models robust to 
typographic attacks~\cite{goh2021multimodal}.
In both cases, we edit the model using a single
\emph{exemplar}, i.e., a single image that we manually annotate and modify.
For comparison, we also consider two variants of fine-tuning using the same
exemplar:
(i) \emph{local} fine-tuning, where we only train the weights of a single layer
$L$ (similar to our editing approach); 
and (ii) \emph{global} fine-tuning, where we also train all other layers between
layer $L$ and the output of the model.
It is worth noting that unlike fine-tuning, our editing approach does not 
utilize class labels in any way.
See Appendix~\ref{app:setup} for experimental details.

\subsection{Tackling new environments: Vehicles on snow}
\label{sec:snow}
Our first use-case is adapting pre-trained classifiers to image 
subpopulations that are under-represented in the training 
data.
Specifically, we focus on the task of recognizing vehicles 
under heavy 
snow conditions---a setting that could be pertinent to self-driving cars---using
a VGG16 classifier trained on
ImageNet-1k.
To study this problem, we collect a set of real photographs from road-related
ImageNet classes using Flickr (details in Appendix~\ref{app:real_setup}).
We then rewrite the model's
prediction rules to map ``snowy roads'' to \class{road}.
To do so, we create an exemplar by manually annotating 
the concept ``road'' in an ImageNet image from a 
\emph{different} class (here, \class{police van}), and the manually replace it 
 with snow texture obtained from Flickr.
We then apply our editing methodology (cf. 
Section~\ref{sec:editing}), using this single \emph{synthetic} snow-to-road 
exemplar---see 
Figure~\ref{fig:intro_fig}. 

\begin{figure}[!t]
	\centering
	\hfill
	\begin{subfigure}[b]{0.48\textwidth}
		\centering
		\includegraphics[width=1\columnwidth]{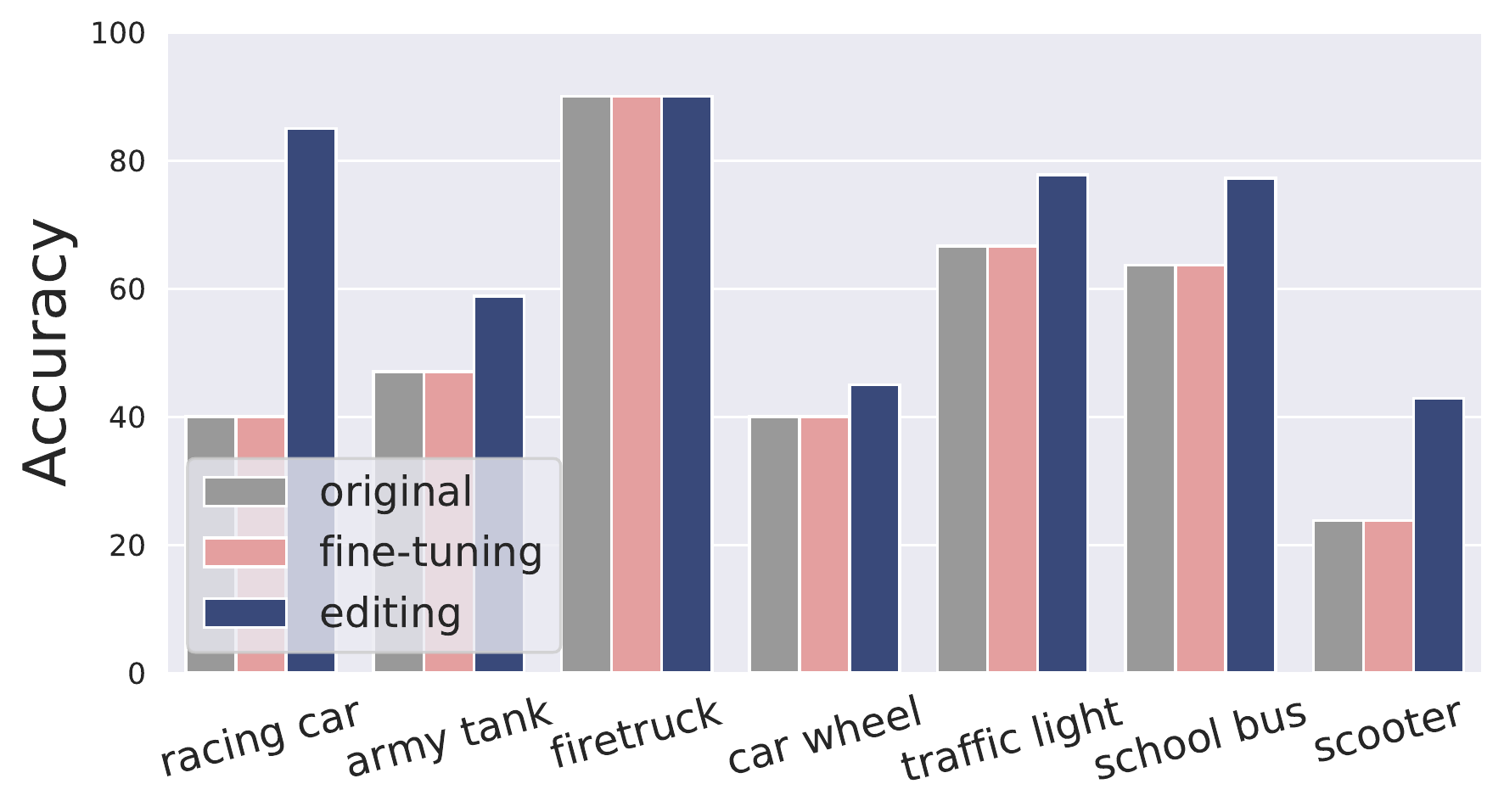}
		\caption{Vehicles in snowy weather}
		\label{fig:snow}
	\end{subfigure}
	\hfill
	\begin{subfigure}[b]{0.48\textwidth}
		\centering
		\includegraphics[width=1\columnwidth]{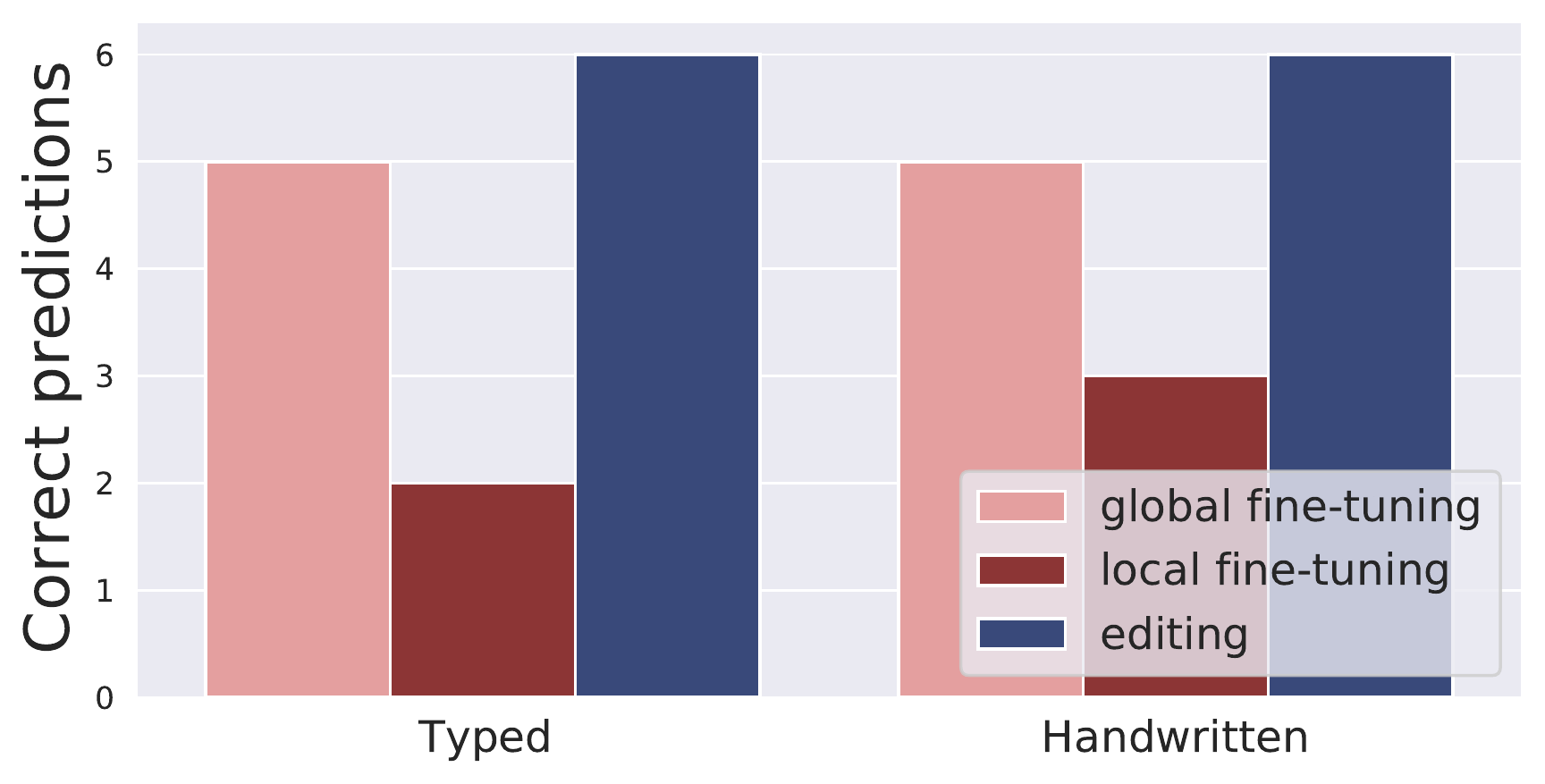}
		\caption{Typographic attacks (6 images)}
		\label{fig:clip}
	\end{subfigure}
	\hfill
	\caption{(a) Adapting a pre-trained ImageNet VGG16 classifier to images of 
		vehicles on snowy roads with a single exemplar. Fine-tuning (both local
        and global) does not improve accuracy, while editing to map \class{snowy 
			road}$\to$\class{road} leads to a consistent improvement across multiple classes. (b) Improving the robustness of 
		CLIP-ResNet-50~\citep{radford2021learning} models to typographic attacks 
		\citep{goh2021multimodal}. Editing the model to map the text 
		\class{iPod}$\to$\class{blank} using a single 
		exemplar---either 
		based on hand-written text on a physical teapot
		or from pasting typed text on an image of a \class{can opener}---completely corrects this vulnerability. While global fine-tuning can also improve model performance in this setting, it requires 
		more careful hyperparameter tuning and typically hurts model performance 
		in other contexts (Appendix Figure~\ref{fig:app_ft_clip_failure}).
        Here, hyperparameters (cf. Appendix 
        Table~\ref{tab:real_hparams}) are chosen based on the large-scale
        synthetic study in Section~\ref{sec:performance}.
    }
\end{figure}

In Figure~\ref{fig:snow}, we measure the error rate of the model on the new 
test set (vehicles in snow) before and after performing the 
rewrite.
We find that our edits significantly improve the model's error rate on these 
images, despite the
fact that we only use a single \emph{synthetic} exemplar (i.e., not a 
{real} 
\class{snowy 
road} photograph).
Moreover, Figure~\ref{fig:snow} demonstrates that our method indeed 
changes the way that the model
processes a concept (here ``snow'') in a way that generalizes beyond the
specific class used during 
editing (here, the exemplar was a \class{police van}) .
In contrast, fine-tuning the model under the same setup does not 
improve its performance on these inputs.

One potential concern is the impact of this process on the model's accuracy
on other ImageNet classes that contain snow
(e.g., \class{ski}).
On the 246 (of 50k) ImageNet test images that contain snow (identified using
an MS-COCO-trained instance segmentation model~\citep{chen2017deeplab}),
the model's accuracy pre-edit is 92.27\% and post-edit is
91.05\%---i.e., only 3/246 images are rendered incorrect by the
edit.
This indicates that the classifier is not disproportionately affected by the
edit.

\subsection{Ignoring a spurious feature: Typographic attacks}
Our second use-case is modifying a model to ignore a spurious feature.
We focus on the recently-discovered 
typographic attacks from 
\citet{goh2021multimodal}: simply attaching a piece of paper with the 
text ``iPod'' on it is enough to make a zero-shot 
CLIP~\citep{radford2021learning} classifier incorrectly classify an assortment of
objects to be iPods. 
We reproduce these attacks on the ResNet-50 variant of the
model---see Appendix Figure~\ref{fig:ipod_full} for an illustration.

To correct this behavior, we rewrite the model's prediction rules to
map the text ``iPod'' to ``blank''.
For the choice of our transformed input $x'$, we consider two variants: 
either a real photograph of a \class{teapot} with the typographic attack (Appendix 
Figure~\ref{fig:ipod_full}); or an ImageNet image of a \class{can opener}
(randomly-chosen) with the typed text ``iPod'' pasted on it 
(Figure~\ref{fig:intro_fig}).
The original image $x$ for our approach is obtained by replacing the 
handwritten/typed text with a white mask---cf. Figure~\ref{fig:intro_fig}.
We then use this single training exemplar to perform 
the model edit.

In both cases, we find that editing is able to fix 
\emph{all} 
the errors 
caused by the typographic attacks, see Figure~\ref{fig:clip}.
Interestingly, global fine-tuning also helps to correct 
many of these errors (potentially by adjusting class biases), albeit less reliably (for specific hyperparameters).
However, unlike editing, fine-tuning also ends up damaging the 
model behavior in other 
scenarios---e.g., the model now spuriously associates the text ``iPod'' with 
the 
target class used for fine-tuning and/or has lower accuracy on normal 
\class{iPod} images from the test set (Appendix 
Figure~\ref{fig:app_ft_clip_failure}).

%% file: sections/discovery.tex
The analysis of the previous section demonstrates that editing
can improve model performance in realistic settings.
However, due to the practical constraints of real-world data 
collection, this analysis was restricted to a relatively small test set.
To corroborate the generality of our approach, we 
now develop a pipeline 
to automatically construct diverse rule-editing test cases.
We then perform a large-scale evaluation and
ablation of our editing method on this testbed.

\begin{figure}[!t]
	\centering
\includegraphics[width=0.8\columnwidth]{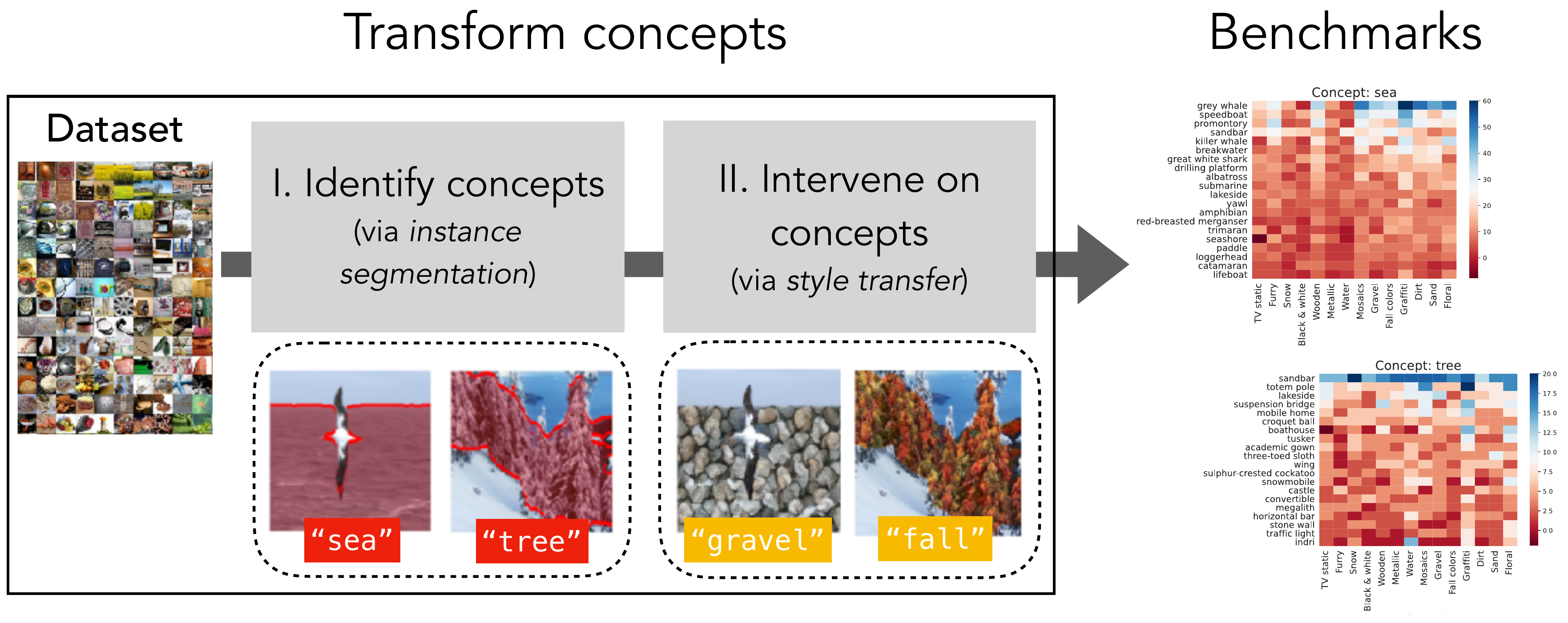}
	\caption{Creating large-scale test sets for model rewriting.
		Given a standard dataset, we first identify salient concepts within the 
		corresponding images using instance segmentation, and then apply a 
		realistic transformation to each of these concepts using style 
		transfer~\citep{gatys2016image}.
		We can then evaluate the effectiveness of a model rewriting technique 
		based on the extent to which it can alleviate the model's sensitivity (i.e., 
		drop in accuracy) to such concept-level transformations.}
	\label{fig:discovery_pipeline}
\end{figure}

\subsection{Synthesizing concept-level transformations}
\label{sec:counterfactuals}
At a high level, our goal is to automatically create a test set(s) in
which 
a specific 
concept---possibly relevant to the detection of multiple 
dataset
classes---undergoes a realistic transformation.
To achieve this without additional data collection, we transform all instances 
of the concept of interest within \emph{existing} datasets.
For example, the ``vehicles-on-snow'' scenario of Section~\ref{sec:snow} 
can be synthetically
reproduced by identifying  the images within a standard dataset  (say 
ImageNet~\citep{deng2009imagenet,russakovsky2015imagenet}) that 
contain 
segments of road and transforming
these to make them resemble snow.
Concretely, our pipeline (see Figure~\ref{fig:discovery_pipeline} for an 
illustration and Appendix~\ref{app:discovery_setup} for details), which takes 
as input an existing dataset, consists of the following two steps:

\begin{enumerate}
	\item \textbf{Concept identification:}
        In order to identify concepts within the dataset images in 
        a scalable manner, we leverage pre-trained instance segmentation 
        models.
        In particular, using state-of-the-art segmentation models---trained on
        MS-COCO~\citep{lin2014microsoft} and 
        LVIS~\citep{gupta2019lvis}---we are able to automatically generate  
        concept 
        segmentations for a range of high-level concepts (e.g., ``grass'', 
        ``sea'', ``tree'').
	
	\item \textbf{Concept transformation:}
		We then transform the detected concept (within dataset images) in a 
		consistent manner using existing methods for style
        transfer~\citep{gatys2016image,ghiasi2017exploring}.
        This allows us to preserve  
        fine-grained image features and realism, while still exploring a range of 
        potential transformations for a single concept.
        For our analysis, we manually curate a set of realistic 
        transformations (e.g.,
        \class{snow} and \class{graffiti}). 
\end{enumerate}

Note that this concept-transformation pipeline does not require \emph{any} 
additional training or data
annotation.
Thus it can be directly applied to new datasets, as long as we have access to a
pre-trained segmentation model for the concepts of interest.

%% file: sections/eval.tex
\subsection{Creating a suite of editing tasks}
We now utilize the concept transformations described above to create a 
benchmark for evaluating model rewriting methods.
Intuitively, these transformations can capture invariances that the 
model should ideally have---e.g., recognizing vehicles correctly 
even when they have wooden wheels.
In practice however, model accuracy on one or more classes (e.g., ``car'', 
``scooter'') may degrade under these 
transformations. 
The goal of rewriting the model would thus be to fix these failure modes in a
data-efficient manner.
In this section, we evaluate our editing methodology---as well as the 
fine-tuning approaches discussed in Section~\ref{sec:realworld}---along this 
axis.

Concretely, we focus on vision classifiers---specifically,
VGG~\citep{simonyan2015very} and ResNet~\citep{he2015residual} models trained on
the ImageNet~\citep{deng2009imagenet,russakovsky2015imagenet} and
Places-365~\citep{zhou2017places} datasets (cf.  Appendix~\ref{app:models}).
Each test set is constructed using the concept-transformation pipeline 
discussed above, based on a chosen concept-style pair (say
\class{wheel}-\class{wooden})~\footnote{Note that some of these cases 
might 
not
    be suitable for editing, i.e., when the transformed concept is critical for
    recognizing the label of an input image (e.g., transforming concept
    \class{dog} in images of class \class{poodle}).  We thus manually exclude
    such concept-class pairs from our analysis---cf. 
    Appendix~\ref{app:filtering}.}. It consists of $N$ exemplars (pairs of 
    original and transformed images, 
$(x,
x')$) that belong to a single (randomly-chosen) target class in the dataset.
All other transformed images containing the concept, including those from
classes
other than the target one, are used for validation and testing (30-70
split). 
We create two variants of the test set: one using the same style image as the
exemplars (i.e., same wooden texture) for the transformation; and another 
using
held-out style images (i.e., other wooden textures).


 To evaluate the impact of a 
method, we measure the change in model accuracy 
on the transformed examples
(e.g., vehicles with \ww{s} in Figure~\ref{fig:editing_pipeline_test}).
If the method is effective, then it should recover some of the incorrect
predictions caused by the transformations.
We only focus on the subset of examples $D$ that were correctly classified
before the transformation, since we cannot expect to correct mistakes that do
not stem from the transformation itself.
Concretely, we measure the change in the number of mistakes 
made by the model on the
transformed examples: 
\begin{equation}
    \text{\% errors corrected} := \frac{N_{pre}(D) - 
N_{post}(D)}{N_{pre}(D)}
\label{eq:improvement}
\end{equation}
where $N_{pre/post}(D)$ denotes the number of transformed examples 
misclassified by the model before and after the rewrite, respectively.  
Note that this metric can range from 100\% when rewriting leads to perfect 
classification on the transformed examples, to even a negative value when
the rewriting process causes more mistakes that it fixes.

In each case, we select the best hyperparameters---including 
the choice of the layer to 
modify---based on the validation set performance 
(cf. Appendix~\ref{app:hyperparameters_setup}). 
To quantify the effect of the modification on overall model behavior, we also 
measure the change in its (standard) test set performance. 
Since we are interested in rewrites that do not significantly hurt the
overall model performance, we only consider hyperparameters that do not
cause a large accuracy drop ($\leq$0.25\%).
We found that the exact accuracy threshold did not have significant impact on
the results---see Appendix
Figures~\ref{fig:app_tradeoff_editing_imagenet_vgg}-\ref{fig:app_tradeoff_editing_places_resnet}
for a demonstration of the full accuracy-effectiveness trade-off.

\subsection{The effectiveness of editing}
\begin{figure}[!t]
	\begin{subfigure}[b]{0.5\textwidth}
		\centering
		\includegraphics[width=1\columnwidth]{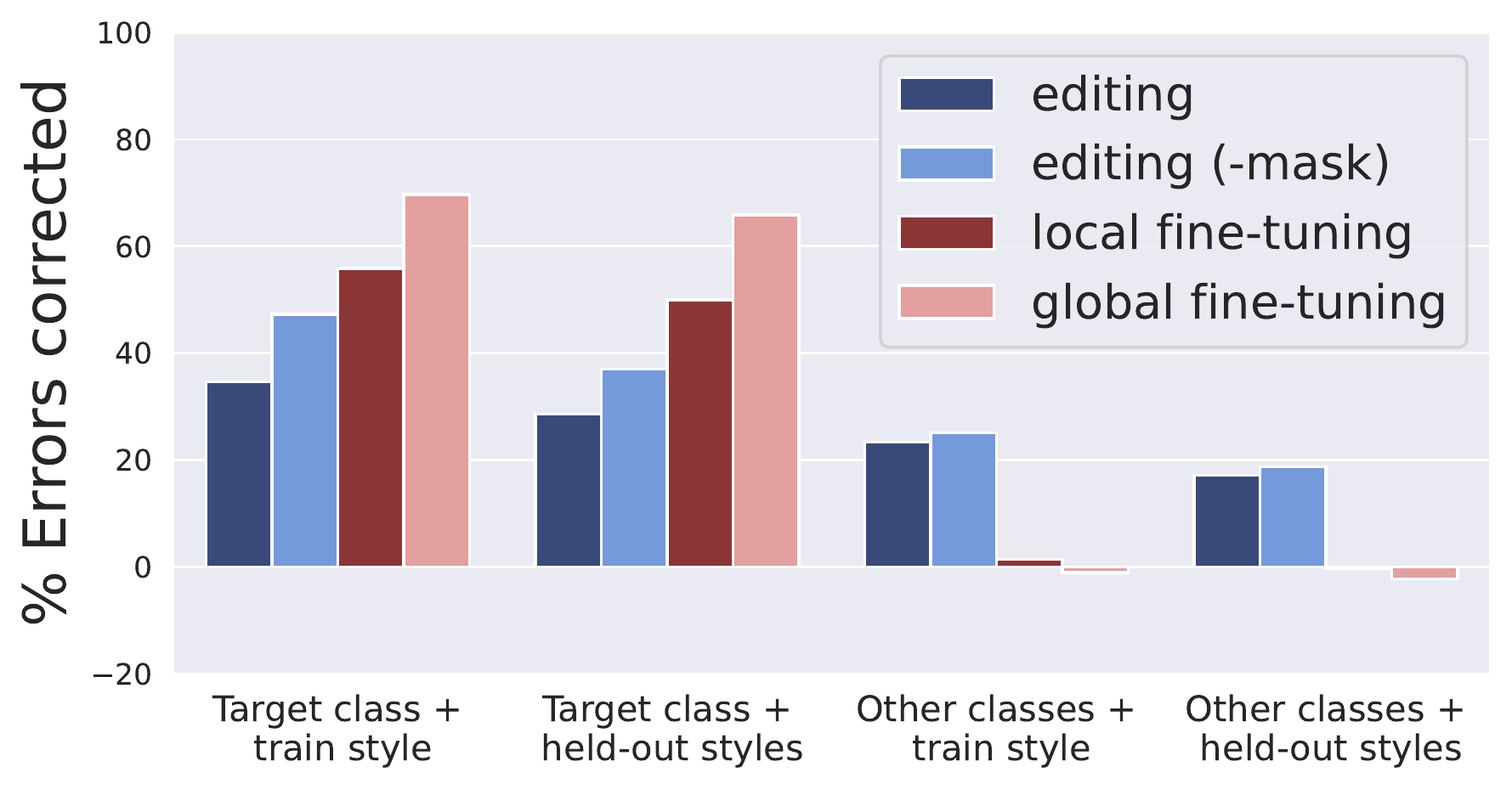}
		\caption{ImageNet-trained VGG16 (concepts from COCO)}
		\label{fig:performance_a}
	\end{subfigure}
	\begin{subfigure}[b]{0.5\textwidth}
		\centering
		\includegraphics[width=1\columnwidth]{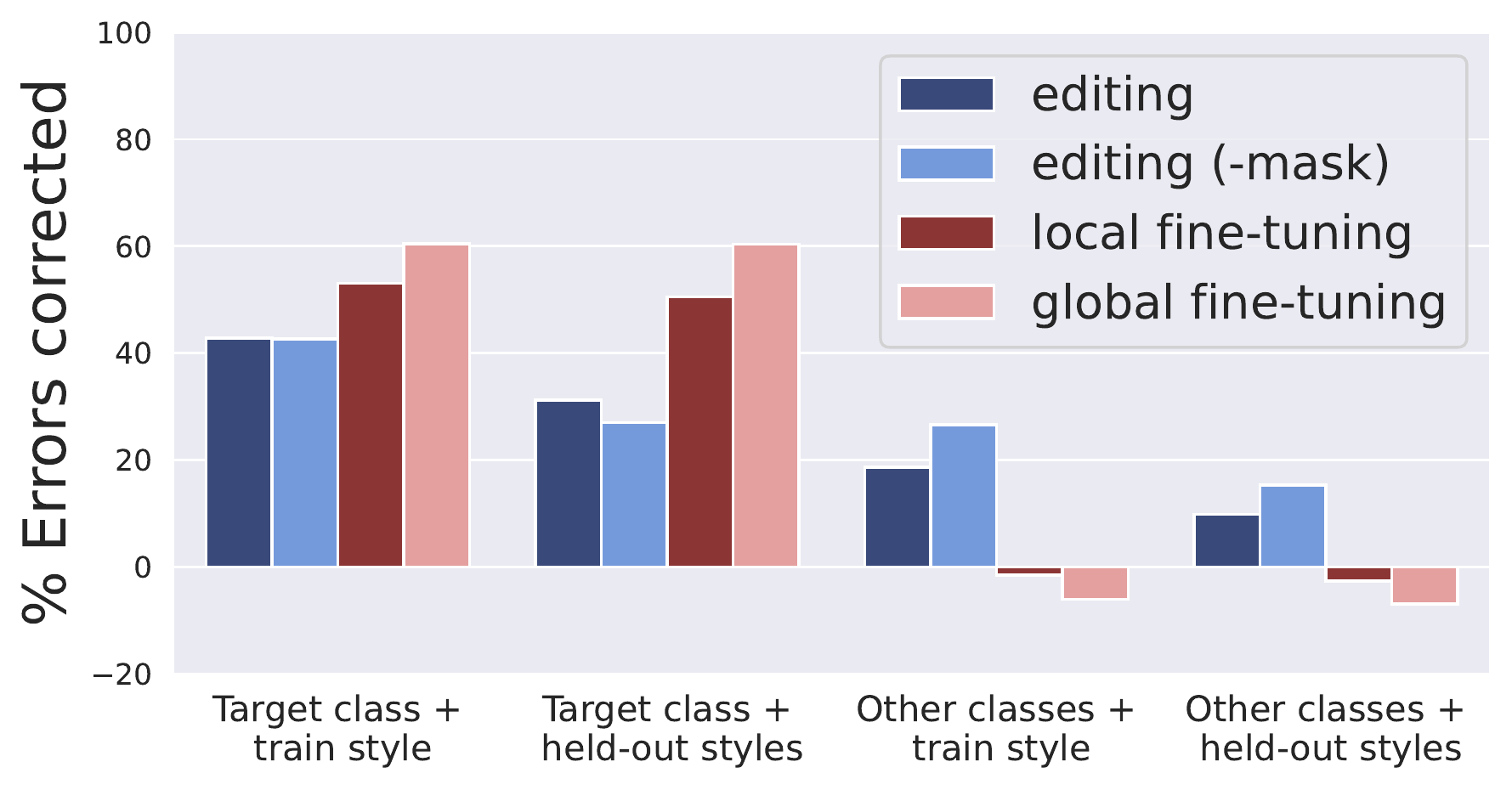}
		\caption{Places-trained ResNet-18 (concepts from LVIS)}
		\label{fig:performance_b}
	\end{subfigure}
	\caption{Editing vs. fine-tuning, averaged over concept-style 
		pairs.
		We find that both methods (and their variants) are fairly successful at 
		correcting 
		misclassifications on the target class (examples of which are used to 
		perform the rewrite).
		This holds even when the transformation applied during 
		testing is different from the one present in the train exemplars (e.g., a 
		different texture of \class{wood}).
		However, crucially, only the improvements induced by editing generalize to other 
		classes where the transformed concept is present. Fine-tuning 
		fails  in this setting---typically, causing more errors than it fixes. See 
		Appendix 
		Figures~\ref{fig:app_performance_imagenet_3}-\ref{fig:app_performance_places_10}
		for other experimental settings.}
	\label{fig:performance}
\end{figure}

Recall that a key desideratum of our prediction-rule edits is that they should  
\emph{generalize}.
That is, if we modify the way that our model treats a specific concept, we want
this modification to apply to \emph{every} occurrence of that concept.
For instance, if we edit a model to enforce that \class{wooden wheels} should be
treated the same as regular \class{wheels} in the context of \class{car} images, we want
the model to do the same when encountering other vehicles with \class{wooden
wheels}.
Thus, when analyzing performance  in
Figure~\ref{fig:performance}, we consider inputs belonging to the 
class 
used to 
perform the edit
separately.

\paragraph{Editing.}
We find that editing is able to consistently correct mistakes in a manner that
\emph{generalizes across classes}.
That is, editing is able to reduce errors in non-target classes, often by more
than 20 percentage points, even when performed using only three exemplars
from the target class.
Moreover, this improvement extends to transformations using 
different variants of the style (e.g., textures of \class{wood}), other
than those present in exemplars used to perform the modification.

In Appendix~\ref{app:ablation}, we conduct ablation studies to get a better sense of the key algorithmic factors driving performance.
Notably, we find that imposing the editing constraints \eqref{eq:final} on the
entirety of the image---as opposed to only focusing on key-value pairs that
correspond to the concept of interest as proposed in 
\citet{bau2020rewriting}---leads to even better performance (cf.
`-mask' in Figure~\ref{fig:performance}).
We hypothesize that this has a regularizing effect as it constrains the 
weights to preserve the original mapping between keys and values in 
regions that do not contain the concept.

\paragraph{Fine-tuning.}
Our baseline is the canonical fine-tuning 
approach, i.e., directly minimizing the cross-entropy loss on the new data (in 
this 
case the transformed images) with respect to the target label.
Similar to Section~\ref{sec:realworld}, we consider both the local and global 
variants of fine-tuning.
We find that while these approaches are able to correct model errors on
transformed inputs of the {target class} used to perform the modification, they
typically \emph{decrease} the model's performance on \emph{other}
classes---i.e., they cause more errors than they fix.
Moreover, even when we allow a larger drop in the model's accuracy, or use more
training exemplars, their performance  often {becomes} \emph{worse} on inputs
from other classes (Appendix
Figures~\ref{fig:app_tradeoff_editing_imagenet_vgg}-\ref{fig:app_tradeoff_editing_places_resnet}).\\

We present examples of errors corrected (or not) by editing and fine-tuning in 
Appendix Figure~\ref{fig:app_errors_corrected}, and provide a per-concept/style break down in Appendix Figures~\ref{fig:app_per_concept} 
and~\ref{fig:app_per_style}.

%% file: sections/analysis.tex
\begin{figure}[!t]
	\centering
	\includegraphics[width=0.95\textwidth]{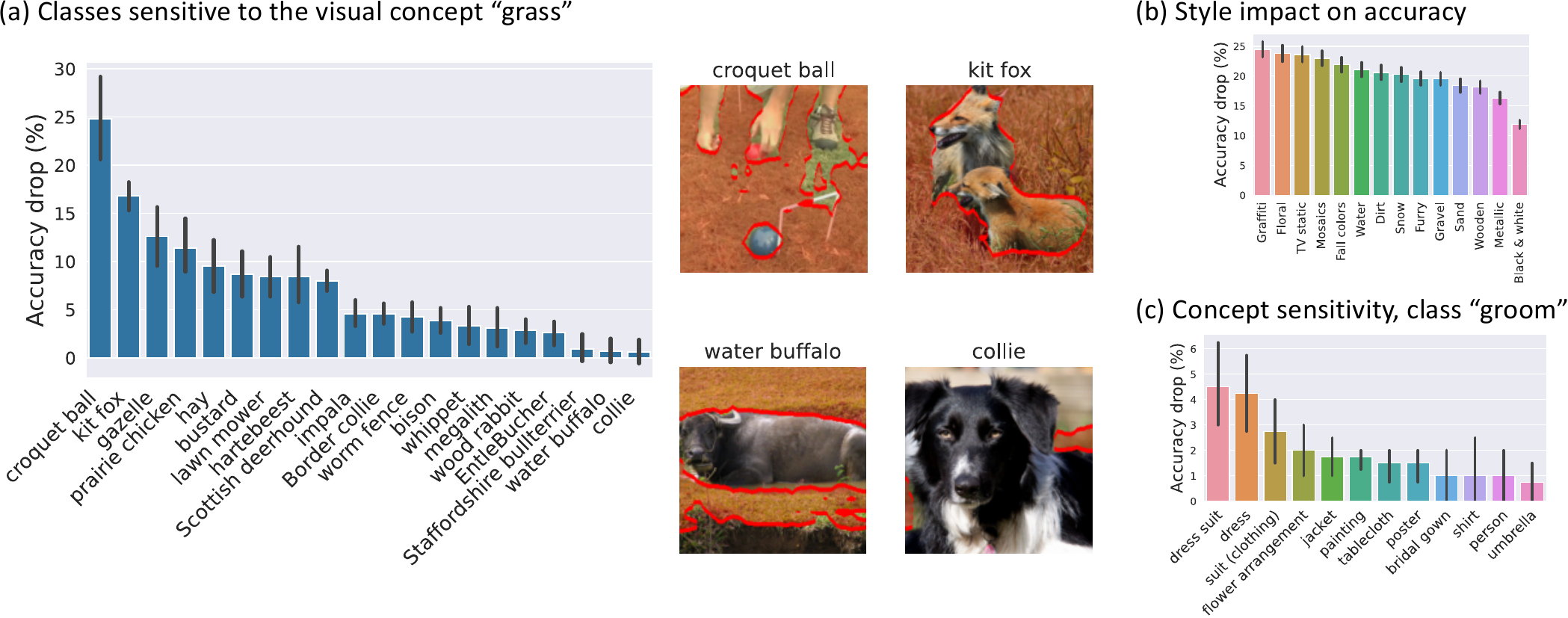}
	\caption{Model sensitivities diagnosed using our pipeline in a VGG16 classifier trained on ImageNet.
	(a) Classes for which the model relies on the concept ``grass'': e.g., a 
	\class{croquet ball} is not accurately recognized
	if \class{grass} is not present, while \class{collie}s are not affected.  (We 
	present the twenty classes for which the visual concept is most often.)  
	(b) Applying different transformations (i.e., styles used for style transfer) 
	to visual concepts reduces accuracy by 
	varying amounts.
	(c) Visual concepts that cause 
	accuracy losses for a given class can highlight prediction rules: e.g.,
	the class ``groom'' is sensitive to the presence of ``dress.''}
	\label{fig:errors_examples}
\end{figure}

In the previous section, we developed a scalable pipeline for creating
concept-level transformations which we used to evaluate model 
rewriting methods.
Here, we put forth another related use-case of that pipeline: 
debugging models to identify their (learned)  prediction rules. 

In particular, observe that the resulting transformed inputs (cf. 
Figure~\ref{fig:discovery_pipeline}) can be viewed 
as \emph{counterfactuals}---a
primitive commonly used in causal inference~\citep{pearl2010causal} and
interpretability~\citep{goyal2019explaining,goyal2019counterfactual,bau2020understanding}.
Counterfactuals can be used to identify the features that a model 
uses to make its prediction on a given input.
For example, to understand whether the model relies on the presence of a
\class{wheel} to recognize a \class{car}  image, we can evaluate how the 
its prediction changes when just the wheel in the image is transformed.
Based on this view, we now repurpose our concept transformations from
Section~\ref{sec:discovery} to discover a given classifier's prediction rules.

\paragraph{The effect of specific concepts.} As a point of start, we study how
sensitive the model's prediction  is to a given high-level concept---in terms of
the accuracy drop caused by transformation of said concept.
For instance, in Figure~\ref{fig:errors_examples}a, we find that the 
accuracy of a VGG16 ImageNet classifier drops by 25\% 
on images of \class{croquet ball} when \class{grass}
is transformed, whereas its accuracy on \class{collie} does not change.
In line with previous 
studies~\citep{zhang2007local,ribeiro2016why,rosenfeld2018elephant,barbu2019objectnet,xiao2020noise},
 we also 
find that 
background concepts, such as \class{grass}, \class{sea} and \class{sand},  
have a large effect on model 
performance.
We can contrast this measure of influence across concepts for a 
single model (Appendix Figure~\ref{fig:app_variation_across_concepts}), 
and across 
architectures 
for a single concept (Appendix~\ref{app:debugging}).
Finally, we can also examine the effect of specific transformations to a single
concept---cf. 
Figure~\ref{fig:errors_examples}b.

\paragraph{Per-class prediction rules.}
If instead we restrict our attention to a single class, we can pinpoint the set 
of concepts that the model relies on to detect this class.
It turns out that aside from the main image object, ImageNet classifiers also 
heavily depend on 
commonly co-occurring 
objects~\citep{stock2018convnets,tsipras2020from,beyer2020are} in the 
image---e.g., ``dress'' for 
the class ``groom'', ``person'' for the class ``tench'' (\emph{sic}), and ``road'' for the class ``race car'' (cf. 
Figure~\ref{fig:errors_examples}c and Appendix
Figure~\ref{fig:app_debugging_classes}). 
We can also examine which transformations hurt 
model performance on the given class most substantially---e.g., we find that 
making ``plants'' ``floral'' hurts accuracy on 
the class \class{damselfly} 15\% more than making them 
 ``snowy''. \newline

Overall, this pipeline provides a scalable path for model designers to analyze 
the invariances (and sensitivities) of their models with respect to 
various natural 
transformations of interest.
We thus believe that this primitive holds promise for future interpretability  
and robustness studies.

%% file: sections/related.tex
\noindent\textbf{High-level concepts in latent representations.}
There has been increasing interest in explaining the inner workings of deep 
classifiers through high-level concepts: e.g., by identifying individual 
neurons~\citep{erhan2009visualizing,zeiler2014visualizing,olah2017feature,bau2017network,engstrom2019learning}
  or activation vectors~\citep{kim2018interpretability,zhou2018interpretable,chen2020concept} 
that correspond to  human-understandable features.
The effect of these features on model predictions can be analyzed by either
inspecting the  downstream
weights of the model~\citep{olah2018building,wong2021leveraging} or through counterfactual
tests: e.g., via synthetic data~\citep{goyal2019explaining}; by swapping
features between individual images~\citep{goyal2019counterfactual}; or by
silencing sets of neurons \citep{bau2020understanding}. 
A parallel line of work aims to learn models that operate on data
representations that explicitly encode high-level concepts: either by learning
to predict a set of attributes~\citep{lampert2009learning,koh2020concept} or by
learning to segment inputs~\citep{losch2019interpretability}.
In our work, we identify concepts by manually selecting the relevant pixels in a
handful of images and measure the impact of manipulating these features
on model performance on a new test set.

\paragraph{Model interventions.}  Direct manipulations of latent
representations inside generative models have been used to create
human-understandable changes in synthesized
images~\citep{bau2019gan,jahanian2019steerability,goetschalckx2019ganalyze,shen2020interpreting,harkonen2020ganspace,wu2020stylespace}.
Our work is inspired by that line of work as well as a recent finding that
parameters of a generative model can be directly changed to alter generalized
behavior~\citep{bau2020rewriting}.  Unlike previous work, we edit classification models, changing rules that govern predictions rather than image synthesis.
Concurrently with our work, there has been a series of methods proposed for
editing factual knowledge in language models~\citep{mitchell2021fast,
de2021editing, dai2021knowledge}.

\paragraph{Ignoring spurious features.}
Prior work on preventing models from relying on spurious correlations is based on constraining model predictions to satisfy certain invariances.
Examples include: training on counterfactuals (either by adding or removing objects from 
scenes~\citep{shetty2018adversarial,shetty2019not,agarwal2020towards} or
    having human annotators edit text input~\citep{kaushik2019learning}), learning representations that are
simultaneously optimal across domains~\citep{arjovsky2019invariant},
ensuring comparable performance across subpopulations~\citep{sagawa2019distributionally}, or enforcing consistency across inputs that depict the same entity~\citep{heinze2017conditional}.
In this work, we focus on a setting where the model designer is aware of
 undesirable correlations learned by the model and we provide 
the tools to
rewrite them directly.

\paragraph{Model robustness.}
A long line of work has been devoted to 
discovering and correcting failure modes of models. 
These studies focus on simulating variations in testing conditions that can arise during  deployment, including: adversarial or natural input corruptions~\citep{szegedy2014intriguing,fawzi2015manitest,fawzi2016robustness, engstrom2019rotation,ford2019adversarial,hendrycks2019benchmarking, kang2019testing}, changes in the data collection process~\citep{saenko2010adapting,torralba2011unbiased,
	khosla2012undoing,tommasi2014testbed,recht2018imagenet}, or variations in the data subpopulations present~\citep{beery2018recognition,oren2019distributionally,sagawa2019distributionally,santurkar2020breeds,koh2020wilds}.
Typical approaches for improving robustness in these contexts include robust
optimization~\citep{madry2018towards,yin2019fourier,sagawa2019distributionally}
and data augmentation schemes~\citep{lopes2019improving,hendrycks2019augmix,zhang2021does}.
Our rule-discovery and editing pipelines can be viewed as complementary to 
this work as they allows us to preemptively adjust the model's prediction rules in anticipation of deployment conditions.

\paragraph{Domain adaptation.}
The goal of domain adaptation is to adapt a model to a specific deployment environment using (potentially unlabeled) samples from it. This is typically achieved by either fine-tuning the model on the new  domain~\citep{donahue2014decaf,razavian2014cnn,kumar2020understanding}, learning the correspondence between the source and target domain, often in a latent representation space~\citep{ben2007analysis,saenko2010adapting,ganin2014unsupervised,courty2016optimal,gong2016domain}, or updating the model's batch normalization statistics~\citep{li2016revisiting, burns2021limitations}.
These approaches all require a non-trivial amount of data from the target
domain. The question of adaptation from a handful of samples has been
explored~\citep{motiian2017few}, but in a setting that requires samples across all target classes. In contrast, our method allows for generalization to new (potentially unknown) classes with even a single example.

%% file: sections/conclusion.tex
We developed a general toolkit for performing targeted post hoc 
modifications to vision classifiers.
Crucially, instead of specifying the desired behavior \emph{implicitly} via the
training data, our method allows users to \emph{directly} edit the model's
prediction rules.
By doing so, our approach makes it easier for users to encode their prior
knowledge and preferences during the model debugging process.
Additionally, a key benefit of this technique is that it fundamentally changes 
how the model processes a given concept---thus making it possible to 
edit its behavior beyond the specific class(es) used for editing.
Finally, our edits do not require any additional data collection: they can be
guided by as few as a single (synthetically-created) exemplar. 
We believe that this primitive opens up new avenues to interact with and correct
our models before or during deployment.

\subsection*{Limitations and broader impact.}
Even though our methodology provides a general tool for model editing,
performing such edits does require manual intervention and domain expertise.
After all, the choice of what concept to edit---and its implications on the
robustness of the model---lies with the model designer.
For instance, in the vehicles-on-snow example, our objective was to have the
model recognize any vehicle on snow the same way it would on a regular
road---e.g., to adapt a system to different weather conditions.
However, if our dataset contains classes for which the presence snow is
absolutely essential for recognition, this might not be an appropriate edit to perform.

Moreover, direct model editing is a
departure from the standard way in which models are trained, and
may have broader implications. While we have shown how it can be used to cause
beneficial changes in pre-trained models, direct control of
prediction rules could also make it easier for adversaries to introduce
vulnerabilities into the model (e.g., by manipulating model behavior on a
specific population demographic). Overall, direct model editing makes it clearer
than ever that our models are a reflection of the goals and biases of we who
create them—not only through the training tasks we choose, but now also through
the rules that we rewrite.

%% file: sections/ack.tex
We thank the anonymous reviewers for their helpful comments and feedback.

Work supported in part by the NSF grants CCF-1553428 and CNS-1815221, the DARPA
SAIL-ON HR0011-20-C-0022 grant, Open Philanthropy, a Google PhD fellowship, and
a Facebook PhD fellowship. This material is based upon work supported by the
Defense Advanced Research Projects Agency (DARPA) under Contract No.
HR001120C0015.

Research was sponsored by the United States Air Force Research Laboratory and
the United States Air Force Artificial Intelligence Accelerator and was
accomplished under Cooperative Agreement Number FA8750-19-2-1000. The views and
conclusions contained in this document are those of the authors and should not
be interpreted as representing the official policies, either expressed or
implied, of the United States Air Force or the U.S. Government. The U.S.
Government is authorized to reproduce and distribute reprints for Government
purposes notwithstanding any copyright notation herein.

%% file: sections/setup.tex
\label{app:setup}

\subsection{Datasets}
\label{app:datasets}

For the bulk of our experimental analysis (Sections~\ref{sec:discovery}
and~\ref{sec:concepts}) we use the
ImageNet-1k~\citep{deng2009imagenet,russakovsky2015imagenet} and
Places-365~\citep{zhou2017places} datasets which contain images from 1,000 and
365 categories respectively.
In particular, both prediction-rule discovery and editing are performed on  samples from the standard test sets to avoid overlap with the training 
data used to develop the models.

\paragraph{Licenses.} Both datasets were collected by scraping online 
image hosting
engines and, thus, the images themselves belong to their individuals who
uploaded them.
Nevertheless, as per the terms of agreement for each dataset
\footnote{\url{http://places2.csail.mit.edu/download.html}}
\footnote{\url{https://www.image-net.org/about.php}}, these images can be 
used for non-commercial research purposes.

\subsection{Models}
\label{app:models}
Here, we describe the exact architecture and training process for each model we
use.
For most of our analysis, we utilize two canonical, yet relatively
diverse model architectures for our study: namely, VGG~\citep{simonyan2015very}
and ResNet~\citep{he2016deep}.
We use the standard PyTorch
implementation~\footnote{\url{https://pytorch.org/vision/stable/models.html}}
and train the models from scratch on the ImageNet and Places365 
datasets.
The accuracy of each model on the corresponding test set is provided in
Table~\ref{tab:app_accuracy}.

\paragraph{ImageNet classifiers.} We study: (i) a VGG16 variant with batch 
normalization
and (ii) a ResNet-50.
Both models are trained using standard hyperparameters: SGD for 90 
epochs with an initial learning rate 
of 0.1 that drops by a factor of 10 every 30 epochs.
We use a momentum of 0.9, a weight decay 
of $10^{-4}$ and a batch size of 256 for the VGG16 and 512 for the 
ResNet-50.

\paragraph{Places365 classifiers.} We study: (i) a VGG16 
and (ii) a ResNet-18.
Both models are
trained for 131072 iterations using SGD with a single-cycle learning rate 
schedule peaking at 2e-2 and descending to 0 at the end of training.
We use a momentum 0.9, a weight 
decay 5e-4 and a batch size of 256 for both models.

\paragraph{CLIP.} For the typographic attacks of 
Section~\ref{sec:realworld}, we use the ResNet-50
models trained via CLIP~\citep{radford2021learning}, as provided in the original
model repository.\footnote{\url{https://github.com/openai/CLIP}}

\begin{table}[h!]
    \setlength{\tabcolsep}{15pt}
    \renewcommand{\arraystretch}{1.3}
	\centering
	\begin{tabular}{lcc} 
        & \multicolumn{2}{c}{Test Accuracy (\%)} \\
		Architecture $\backslash$ Dataset & ImageNet & Places \\
		\hline
		VGG         & 73.70 & 54.02 \\ 
		ResNet      & 75.77 & 54.24 \\
		CLIP-ResNet & 59.84 & - \\
	\end{tabular}
    \vspace{1em}
	\caption{Accuracy of each model architecture on the datasets used in our
    analysis.}
	\label{tab:app_accuracy}
\end{table}

\subsection{Compute}
\label{app:compute}
Our experiments were performed on our internal cluster, comprised mainly of
NVIDIA 1080Ti GTX GPUs.
For the rule-discovery pipeline, we only need to evaluate models on images where
the corresponding object is present, which allows us to perform the evaluation
on a single style in less than 8 hours on a single GPU (amortized over concepts
and classes).
For the model rewriting process, each instance of editing or fine-tuning takes a
little more than a minute, since it only operates on a section of the model and
using a handful of training examples.

\subsection{Model rewriting}
\label{app:model_rewriting}
Here, we describe the training setup of our model 
editing process, as well as the fine-tuning baseline.
Recall that these rewrites are performed with respect to a single 
concept-style 
pair.

\paragraph{Layers.} We consider a layer to be a block of 
convolution-BatchNorm-ReLU, similar to \citet{bau2020rewriting} and
rewrite the weights of the convolution.
For ResNets (which were not previously studied), we 
must also account for skip connections.
In particular, note that the effect of a rewrite to a layer 
inside any residual block will be attenuated (or canceled) by the 
skip connection.
To avoid this, we only rewrite the final layer within each residual block---i.e., 
focus on the convolution-BatchNorm-ReLU right before a skip connection, 
and include the skip connection in the output of the layer.
Unless otherwise specified, we perform rewrites to layers $[8,10,11,12]$ for 
VGG models, $[4,6,7]$ for ResNet-18, and $[8,10,14]$ for
ResNet-50 models.
We tried earlier layers in our initial experiments, but found that both methods
perform worse.

\subsubsection{Editing}
\label{app:editing_setup}

We use the ADAM optimizer with a fixed learning rate to perform the 
optimization in \eqref{eq:final}.
We grid over different learning rate-number of step pairs:
$[(10^{-3}, 10k)$, $(10^{-4}, 20k)$, $(10^{-5}, 40k)$,
$(10^{-6}, 80k)$, $(10^{-7}, 80k)]$.
The second order statistics (cf. Section~\ref{sec:editing})
are computed based on the keys for the entire test set.

\subsubsection{Fine-tuning}
\label{app:ft_setup}
When fine-tuning a single layer (local fine-tuning), we optimize the weights of
the convolution of that particular layer.
Instead, when we fine-tune a suffix of the model (global fine-tuning), we
optimize all the trainable parameters including and after the chosen layer.
In both cases, we use SGD, griding over different learning rate-number of 
step pairs:
$[(10^{-2}, 500)$, $(10^{-3}, 500)$, $(10^{-4}, 500)$, $(10^{-5}, 800)$,
$(10^{-6}, 800)]$.
We verified that in all cases the optimal performance of the method was achieved
for hyperparameters strictly within that range and thus performing more steps
would not provide any benefits.

\subsection{Real-world test cases}
\label{app:real_setup}
In Section~\ref{sec:realworld} we study two real-world applications of our
model rewriting methodology. Below, we outline the data-collection process 
for each case as well as the hyperparameters used.

\paragraph{Vehicles on snow.} We manually chose a subset of Imagenet 
classes that frequently contain \class{roads}, identified using our
prediction-rule discovery pipeline in Section~\ref{sec:discovery}.
In particular, we focus on the classes: \class{racing car}, 
\class{army
tank}, \class{fire truck}, \class{car wheel}, \class{traffic light},
\class{school bus}, and \class{motor scooter}.
For each of these classes, we searched
Flickr\footnote{\url{https://www.flickr.com/}} using the query
``$<$class name$>$ on snow'' and manually selected the images that clearly
depicted the class and actually contained snowy roads.
We were able to collect around 20 pictures for each class with the exception
of \class{traffic light} where we only found 9.

\begin{figure}[!h]
	\centering
		\includegraphics[width=.9\textwidth]{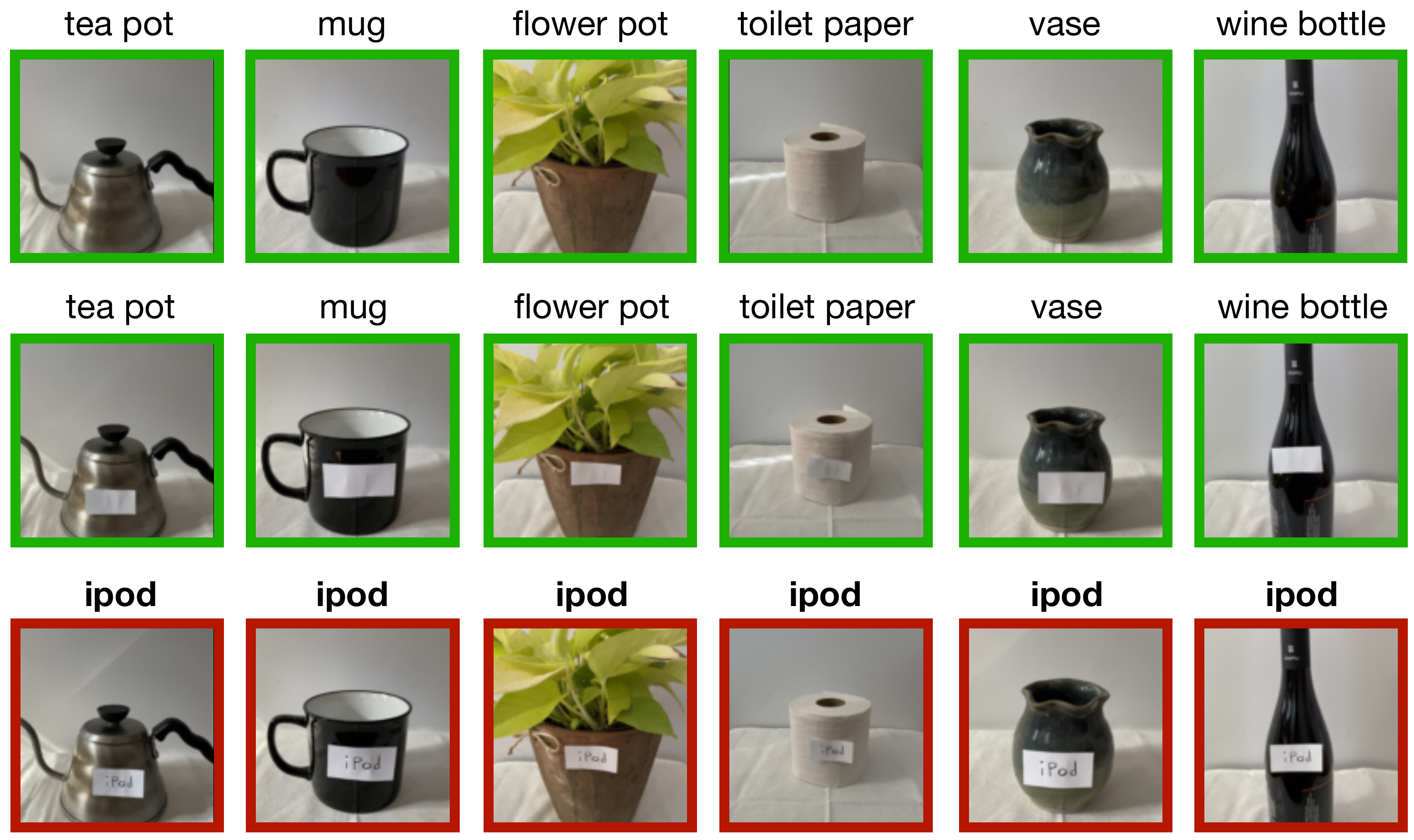}
        \caption{Typographic attacks on CLIP: We reproduce the results of 
        \citet{goh2021multimodal} by taking photographs of household objects 
        with a paper containing handwritten text ``iPod'' attached to them 
        (third row). We see that these attacks consistently fool the 
        zero-shot CLIP classifier (ResNet-50)---compare the predictions (shown 
        in the title) for the first and third row. In contrast, if we instead
        use a blank piece of paper (second row), the model predicts correctly.}
	\label{fig:ipod_full}
\end{figure}

\paragraph{Typographic attacks.} We picked six household objects corresponding
to ImageNet classes, namely: \class{teapot}, \class{mug}, \class{flower pot},
\class{toilet tissue}, \class{vase}, and \class{wine bottle}.
We used a smartphone camera to photograph each of these objects against a plain
background.
Then, we repeated this process but after affixing a piece of paper with the text
``iPod'' handwritten on it, as well as when affixing a blank piece of
paper---see Figure~\ref{fig:ipod_full}.

\paragraph{Hyperparameters.} Since our manually collected test sets are
rather small, we decided to avoid tuning hyperparameters on them as this would
require holding out a non-trivial number of data points.
Instead, we inspected the results of the large-scale synthetic evaluation and
manually picked values that performed consistently well which we list in
Table~\ref{tab:real_hparams}.
Nevertheless, we found that the results would be quite similar if we tuned
hyperparameters directly on these test sets.

\begin{table}[h!]
    \setlength{\tabcolsep}{10pt}
    \renewcommand{\arraystretch}{1.3}
	\centering
	\begin{tabular}{ll|cccc} 
        Model & Method & \# steps & Step size & Layer & Mask \\
		\hline
        \multicolumn{1}{l}{VGG16}
             & Editing & 20,000 & 1e-4 & 12 (last) & No \\
             & Fine-tuning & 400 & 1e-4 & 12 (last) & N/A \\
		\hline
         \multicolumn{1}{l}{CLIP-ResNet-50}
             & Editing & 20,000 & 1e-4 & 14 (last) & No \\
             & Fine-tuning & 800 & 1e-5 & 14 (last) & N/A 
	\end{tabular}
    \vspace{1em}
	\caption{Hyperparameters chosen for evaluating on the real-world test cases.}
	\label{tab:real_hparams}
\end{table}

\subsection{Synthetic evaluation}
We now describe the details of our evaluation methodology, namely, how we 
transform inputs, how we chose which concept-style pairs to use for testing and
how we chose the hyperparameters for each method.

\subsubsection{Creating concept-level transformations}
\label{app:discovery_setup}
\begin{figure}[!t]
	\centering
	\begin{subfigure}[b]{0.48\textwidth}
		\centering
		\includegraphics[width=1\columnwidth]{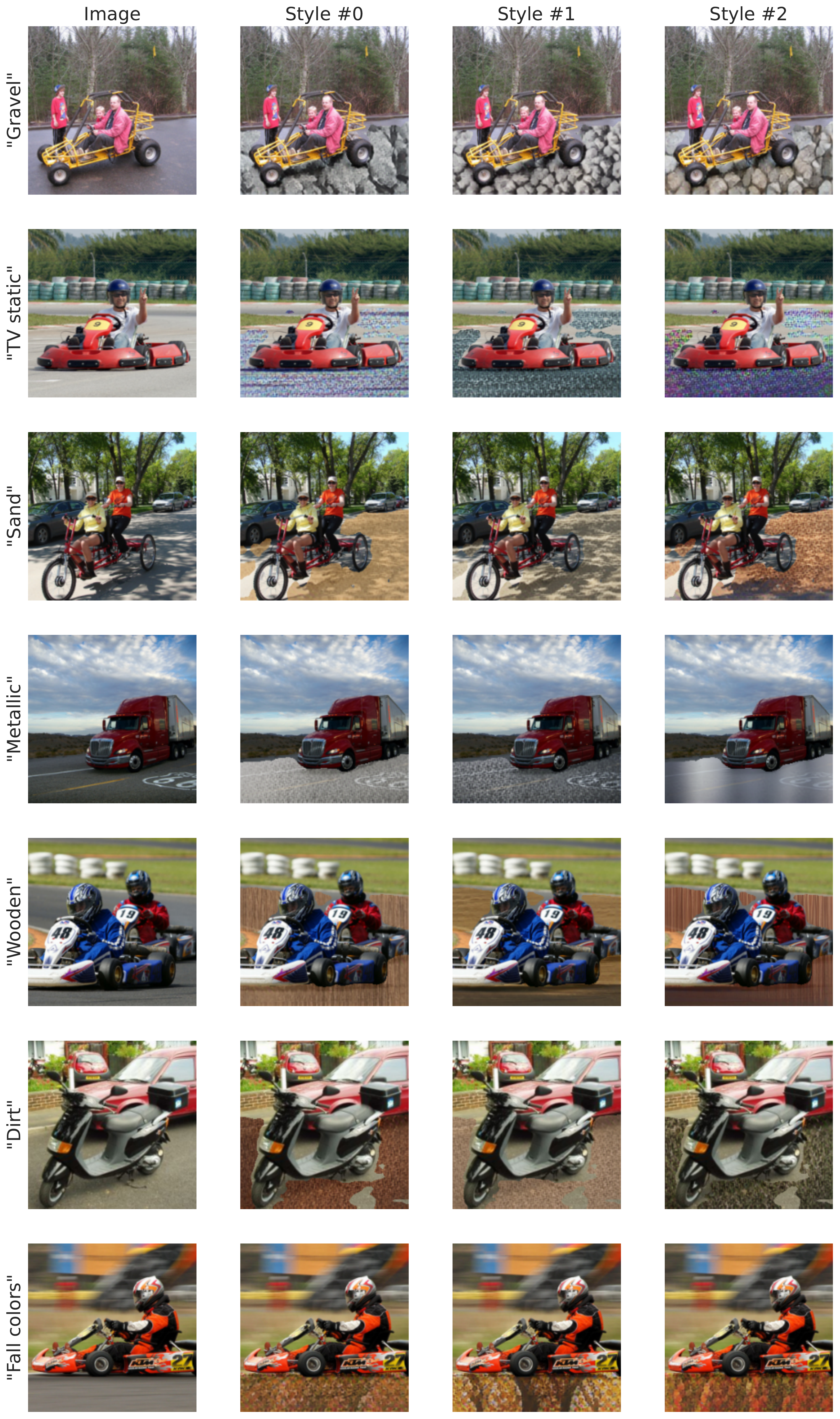}
		\caption{}
	\end{subfigure}
	\hfill
	\begin{subfigure}[b]{0.48\textwidth}
		\centering
		\includegraphics[width=1\columnwidth]{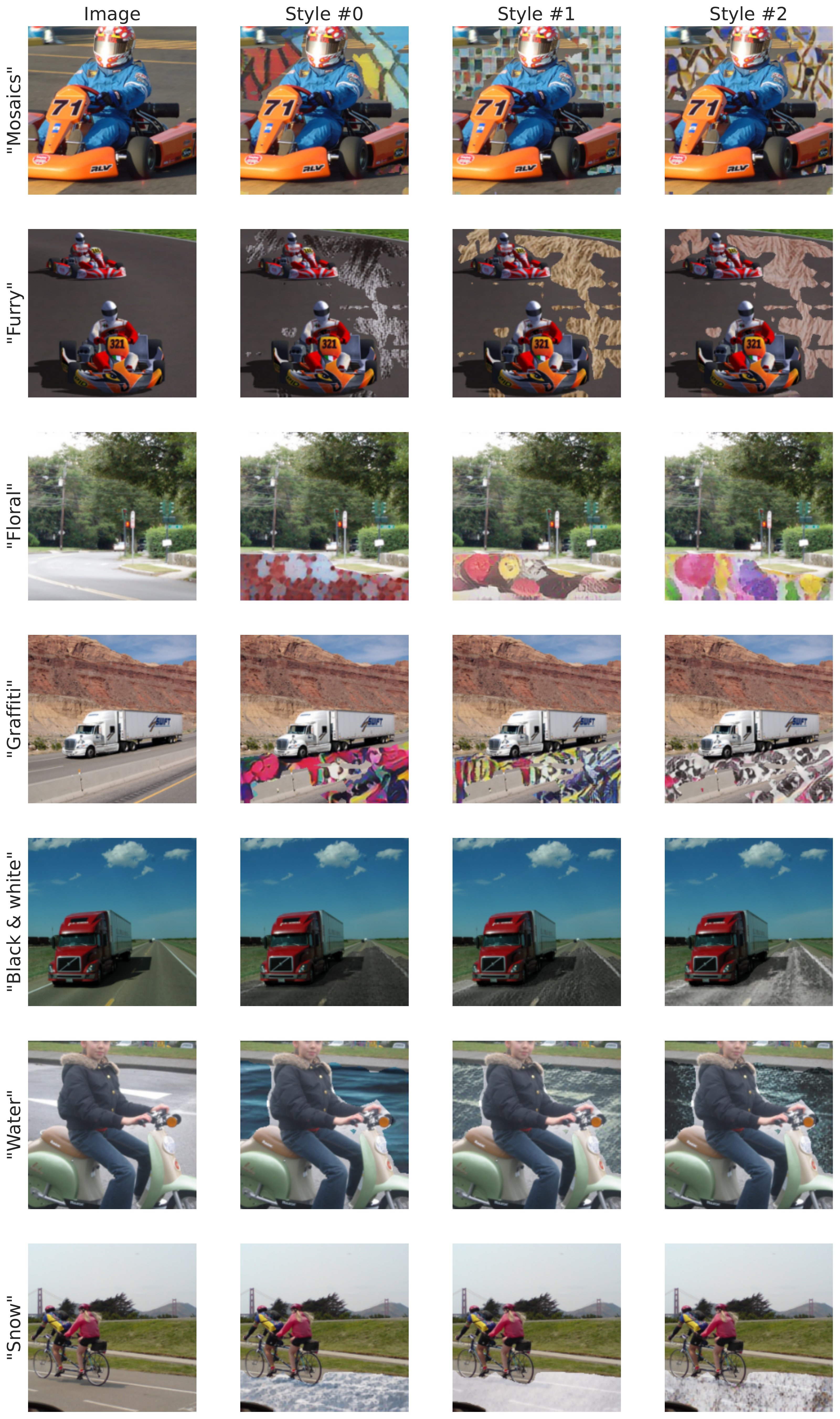}
		\caption{}
	\end{subfigure}
	\caption{Illustration of concept-level transformations in ImageNet: We 
		transform the concept ``road'' in images belonging to various 
		classes via style transfer. Each row (within (a) and (b)) depicts the 
		stylization of a single image with respect to the style described in the 
		label (e.g., ``gravel''). We collect three examples per style, 
		which are then split across training and testing.}
	\label{fig:app_style_examples}
\end{figure}

Recall that our pipeline for transforming concepts  consists of two steps:
concept detection and concept transformation (Section~\ref{sec:discovery}).
We describe each step below and provide examples in
Figure~\ref{fig:app_style_examples}.

We detect concepts using pre-trained object detectors trained on
MS-COCO~\citep{lin2014microsoft} and LVIS~\citep{gupta2019lvis}.
For MS-COCO, we use a model with a ResNet-101
backbone\footnote{\url{https://github.com/kazuto1011/deeplab-pytorch}}
which is trained on
COCO-Stuff\footnote{\url{https://github.com/nightrome/cocostuff}} annotations
and can detect 182 concepts~\citep{chen2017deeplab}.
For LVIS, we use a pre-trained model from the
Detectron~\citep{girshick2018detectron} model
zoo\footnote{\url{https://github.com/facebookresearch/detectron2/blob/master/MODEL_ZOO.md}},
which can detect 1230 classes.
We only consider a prediction as valid for a specific pixel if the model's
predicted probability is at least 0.80 for the COCO-based model and 0.15 for the
LVIS-based model (chosen based on manual inspection).
Moreover, we treat a concept as present in a specific image if it present in at
least 100 pixels (image size is 224$\times$224 for ImageNet and 256$\times$256
for Places).

In order to transform concepts, we utilize the fast style transfer 
methodology of
\citet{ghiasi2017exploring} using their pre-trained
model\footnote{\url{https://tfhub.dev/google/magenta/arbitrary-image-stylization-v1-256/2}}.
This allows us to quickly apply the same style to a large number of images which
is ideal for our use-case.
Specifically, we manually choose 14 styles (illustrated in
Figure~\ref{fig:app_style_examples}) and choose 3 images for each.
This allows us to perform the concept-level transformation in several ways and
evaluate how sensitive our model is to the exact style used.

All the pre-trained models used are open-sourced and available for
non-commercial research.

\subsubsection{Selecting concept-style pairs}
\label{app:filtering}

\paragraph{Concept selection.}
Recall that our transformation pipeline from Section~\ref{sec:discovery}
identifies concepts which, when transformed in a certain manner hurts model 
accuracy on one or more classes.
We first filter these concepts (automatically) to identify ones that are 
particularly salient in the model's prediction-making process. 
In particular, we focus on concepts which
simultaneously: (a) affect at least 3 classes; (b) are present in at least 20\%
percent of the test images of each class; and (c) cause a drop of at least 15\%
among these images.
This selection results in a test bed where we can meaningfully observe
differences in performance between approaches.

At the same time, we need to also ensure that the rewriting task we are
solving is meaningful.
For instance, if we replace all instances of \class{dog} with a stylized
version, then distinguishing between a \class{terrier} and a \class{poodle}
can become challenging (or even impossible).
Moreover, we cannot expect the performance of the model to improve on 
other dog breeds if we modify it to treat a stylized dog as a \class{terrier}.
To eliminate such test cases, we manually filter the concept-class pairs 
flagged by our prediction-rule discovery pipeline.
In particular, we removed those where the detected concept overlapped 
significantly with the class object itself.
In other words, if the concept detected is essential for correctly recognizing
the class of the image, we exclude it from our analysis.
Typical examples of excluded concept-class pairs on ImageNet include broad
animal categories (e.g., \class{bird} or \class{dog}) for classes corresponding
to specific breeds (e.g., \class{parrot}) or the concept \class{person} which
overlaps with classes corresponding to articles of clothing (e.g.,
\class{suit}).

\paragraph{Style selection.} We consider a subset of 8 styles for our analysis:
``black and white'', ``floral'', ``fall 
colors'', ``furry'', ``graffiti'', ``gravel'', ``snow'' and ``wooden''.
While performing editing with respect to a single concept-style pair---say 
``wheel''-``wooden''---we randomly select one wooden texture to create 
train exemplars and hold out the other two for testing (described as held-out 
styles 
in the figures).

\subsubsection{Hyperparameter selection}
\label{app:hyperparameters_setup}
As discussed in Appendix~\ref{app:model_rewriting}, for a particular 
concept-style pair, we grid over different 
hyperparameters 
pertaining to the rewrite (via editing or 
fine-tuning)---in particular the layer that is modified, as well as training 
parameters such as the learning rate.
For our evaluation, we then choose a single set of hyperparameters (per 
concept-style pair).
At a high level, our objective is to find hyperparameters that improve model 
performance on transformed examples, while also ensuring that the test 
accuracy of the model does not drop below a
certain threshold.
To this end, we create a validation set per concept-style pair with 30\% of the 
examples containing this concept (and transformed using the same style as 
the train exemplars).
We then use the performance on that subset \eqref{eq:improvement} to 
choose the best set of
hyperparameters. 
If all of the hyperparameters considered cause accuracy to drop below the
specified threshold, we choose to not perform the edit at all.
We then report the performance of the method on the test set (the other 
70\% of samples containing this concept).

\subsection{Data ethics}
\label{app:ethics}
Since we manually collected all the data necessary for our analysis in
Section~\ref{sec:realworld}, we were able to filter them for offensive content. 
Moreover, we made sure to only collect images that are available under a
Creative Commons license (hence allowing non-commercial use with proper
attribution).

For the rest of our analysis, we relied on publicly available datasets that are
commonly used for image classification.
Unfortunately, due to their scale, these datasets have not been thoroughly
filtered for offensive content or identifiable information.
In fact, improving these datasets along this axis is an active area of
work\footnote{\url{https://www.image-net.org/update-mar-11-2021.php}}.
Nevertheless, since our research did not involve redistributing these datasets
or presenting them to human annotators, we did not perceive any additional
risks that would result from our work.

%% file: sections/appendix.tex
\input{sections/app_editing}

\clearpage
\subsection{Discovering Prediction-rules}
\label{app:debugging}
Here, we expand on our analysis in Section~\ref{sec:concepts} so as to 
characterize the effect of concept-level transformations on classifiers.

\begin{figure}[!h]
	\centering
	\begin{subfigure}[b]{0.9\textwidth}
		\centering
		\includegraphics[width=4.5cm]{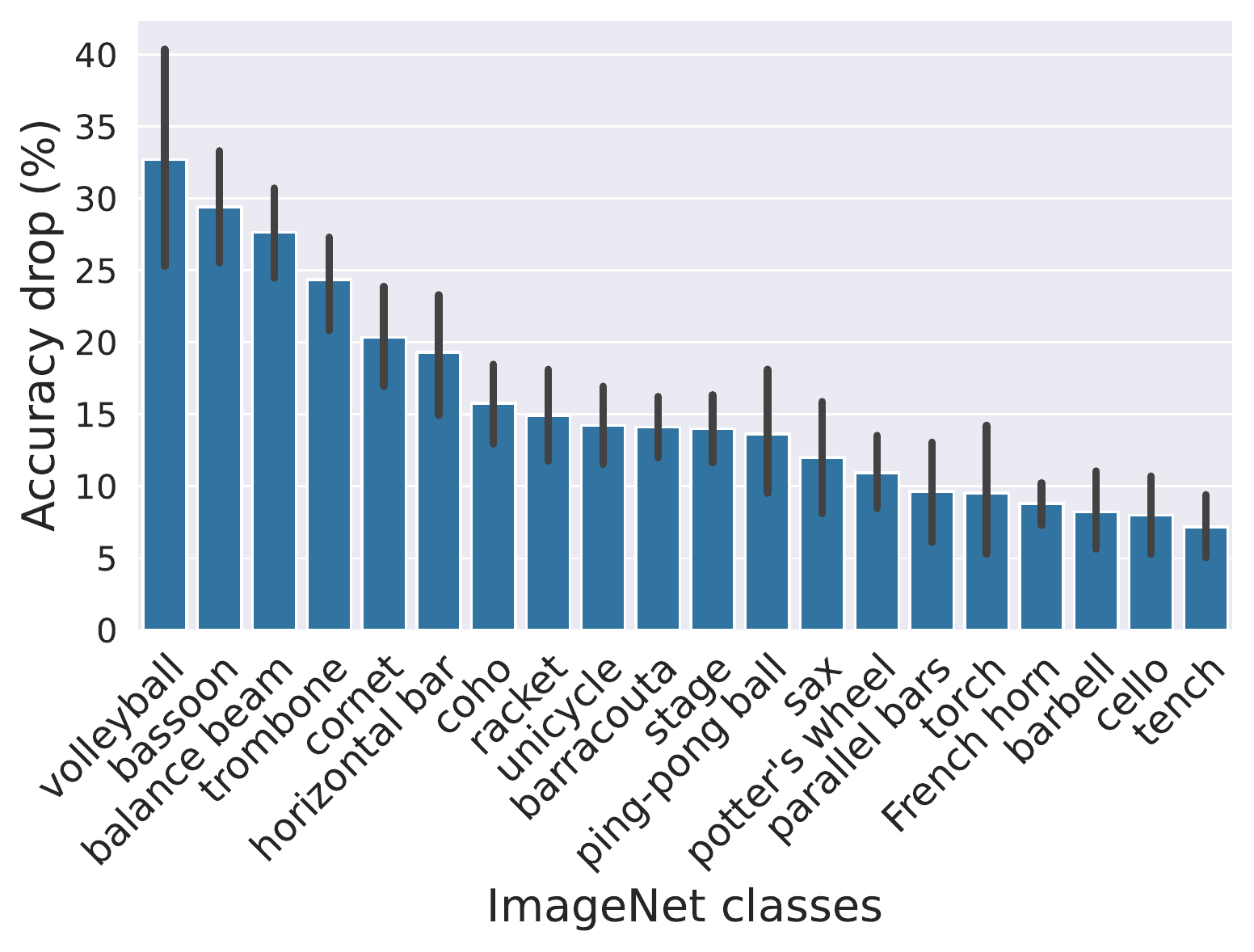}
		 \hfil
		\includegraphics[width=4.5cm]{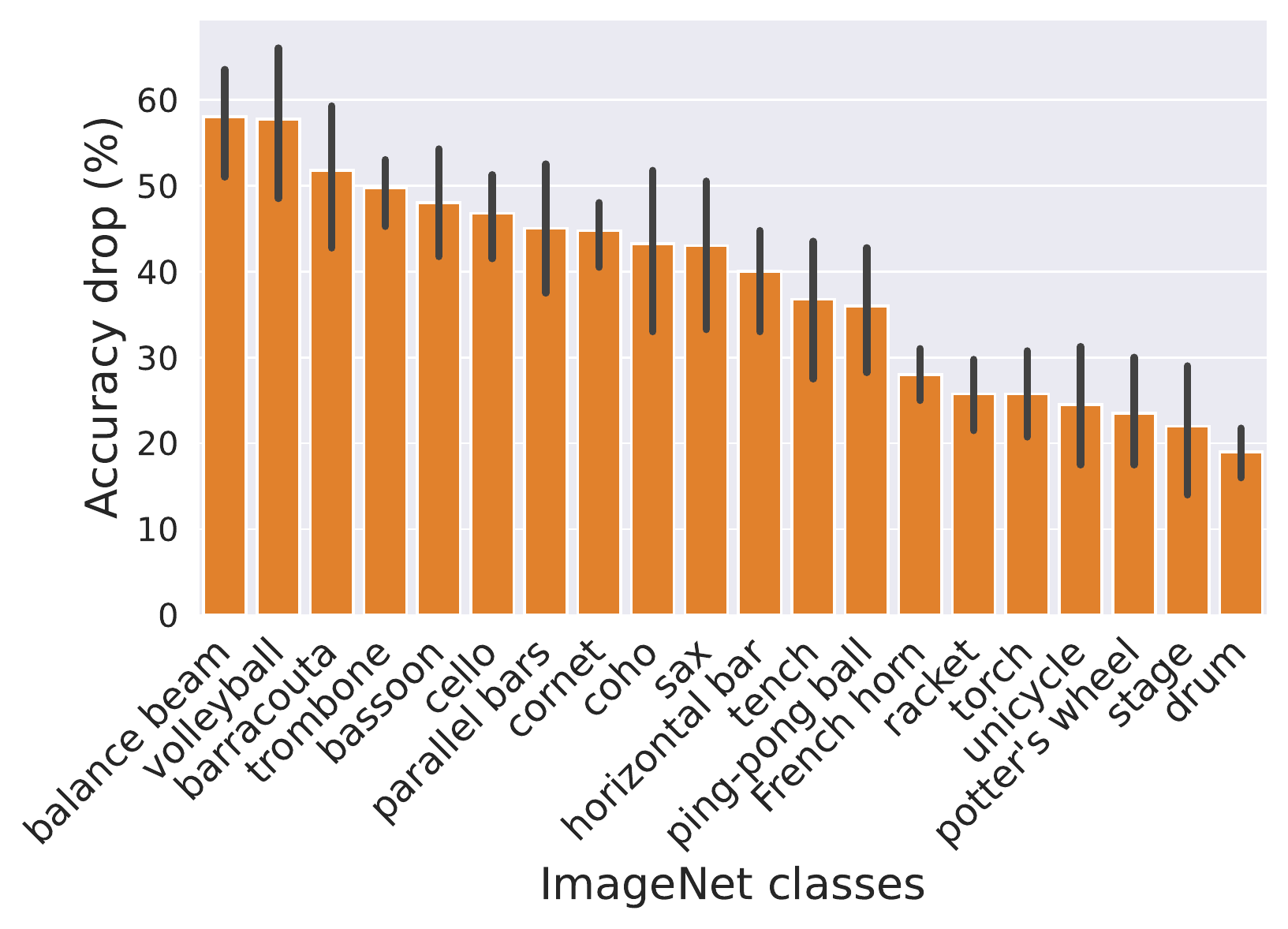}
		\caption{Concept: ``person''; Models: VGG16 
		(\emph{left}) and ResNet-50 (\emph{right}) trained on ImageNet-1k. }
	\end{subfigure}
	\begin{subfigure}[b]{0.9\textwidth}
		\centering
		\includegraphics[width=4.5cm]{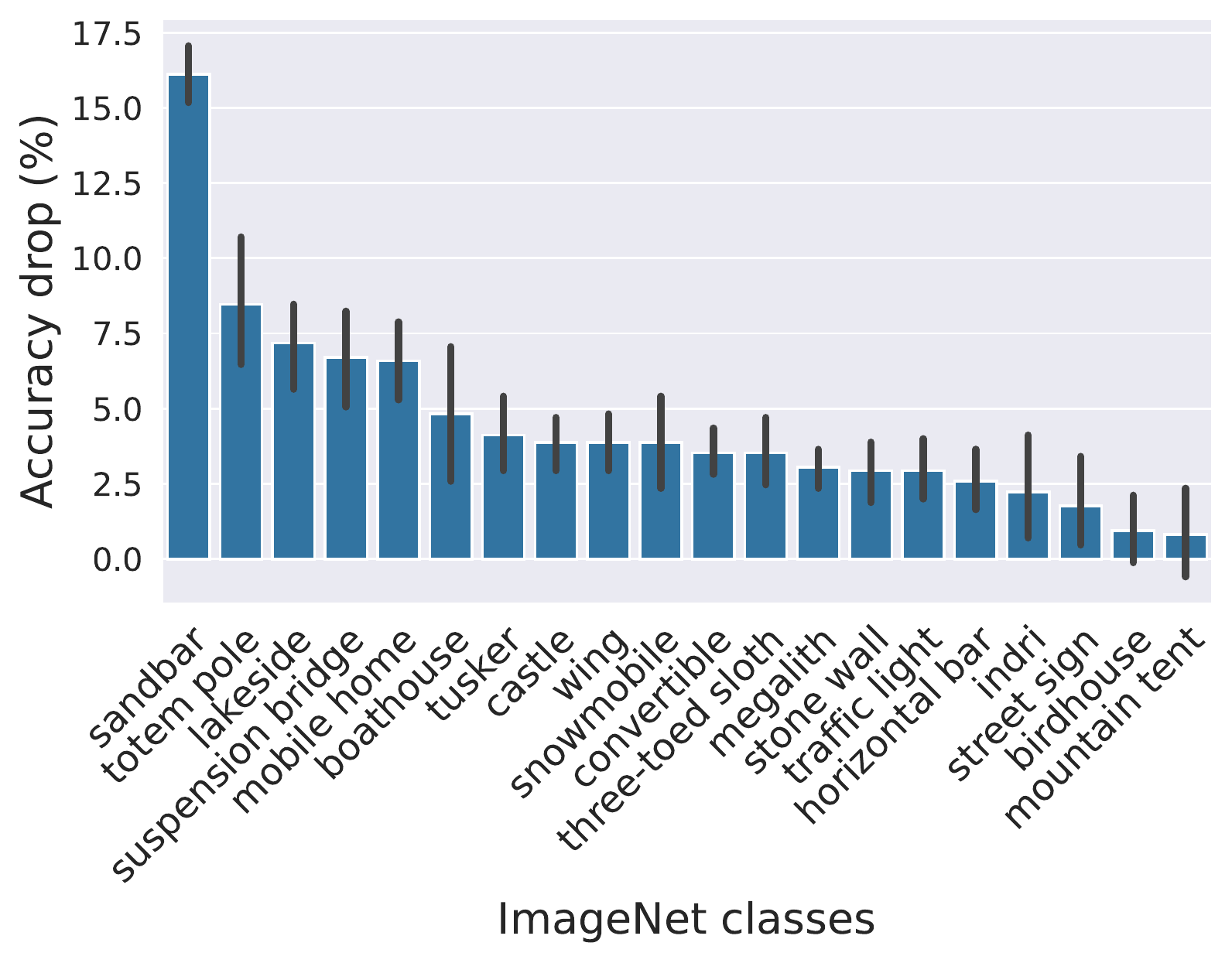}
		 \hfil
		\includegraphics[width=4.5cm]{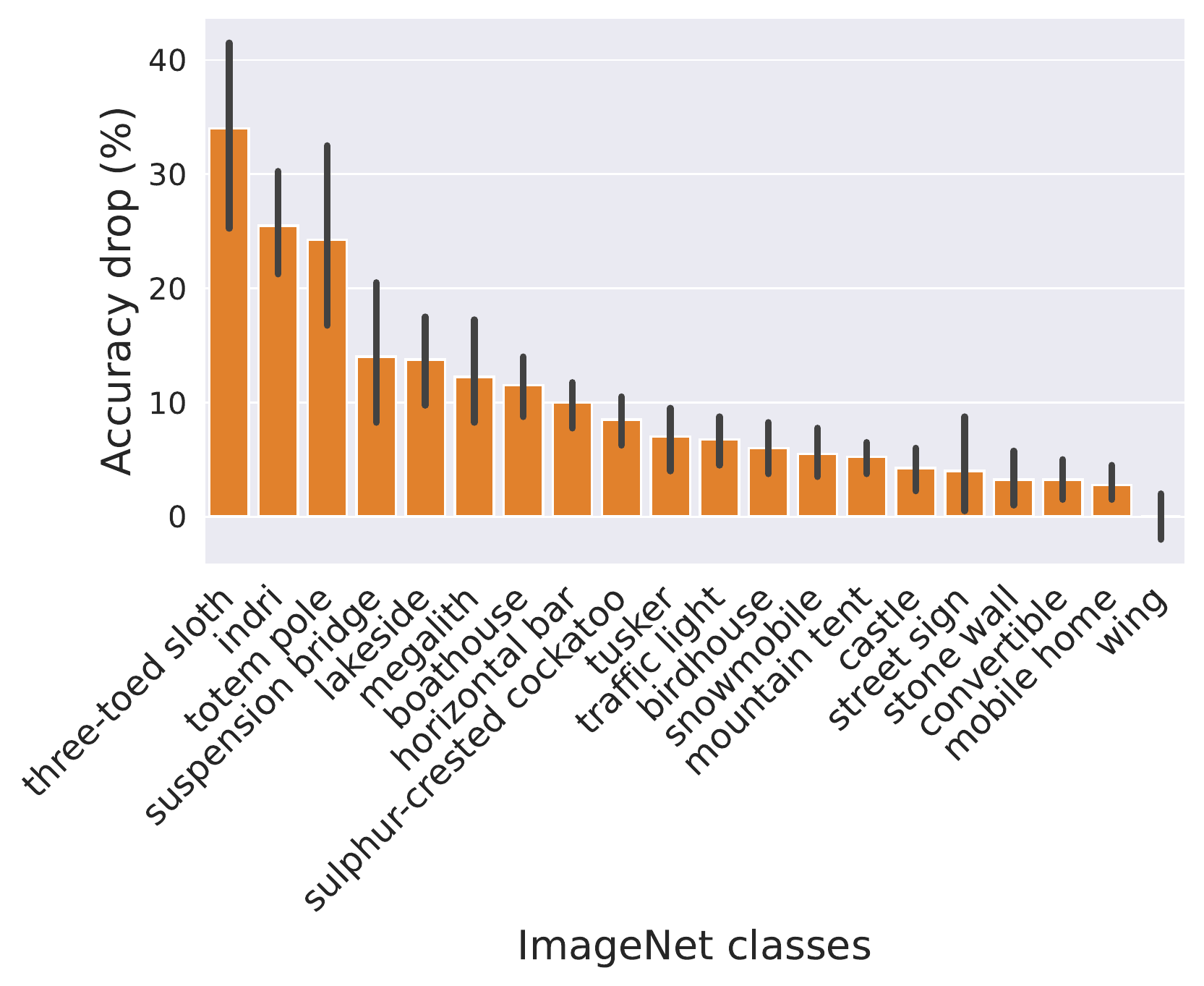}
		\caption{Concept: ``tree''; Models: VGG16 
			(\emph{left}) and ResNet-50 (\emph{right})  trained on ImageNet-1k.}
	\end{subfigure}
	\begin{subfigure}[b]{0.9\textwidth}
	\centering
	\includegraphics[width=4.5cm]{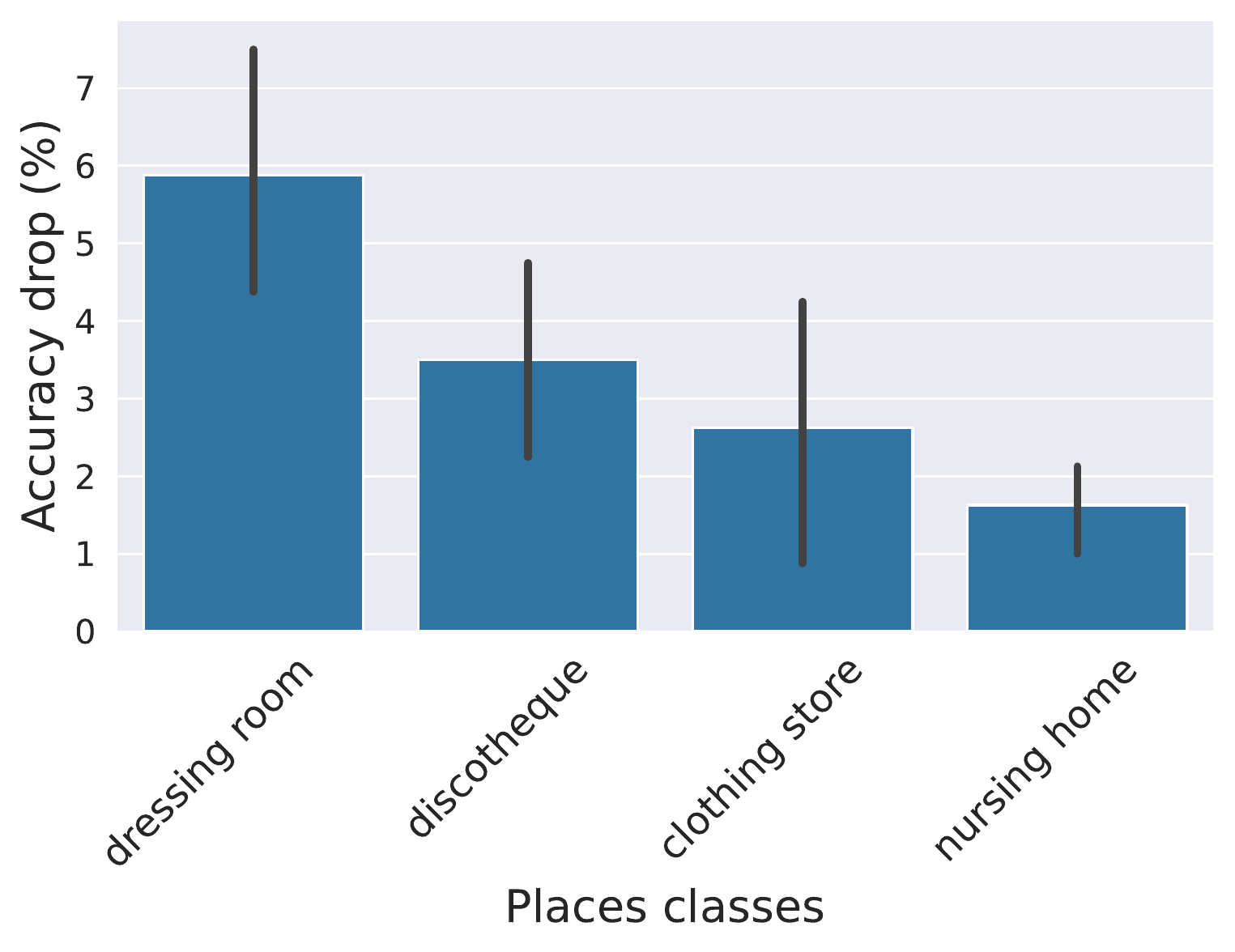}
	 \hfil
	\includegraphics[width=4.5cm]{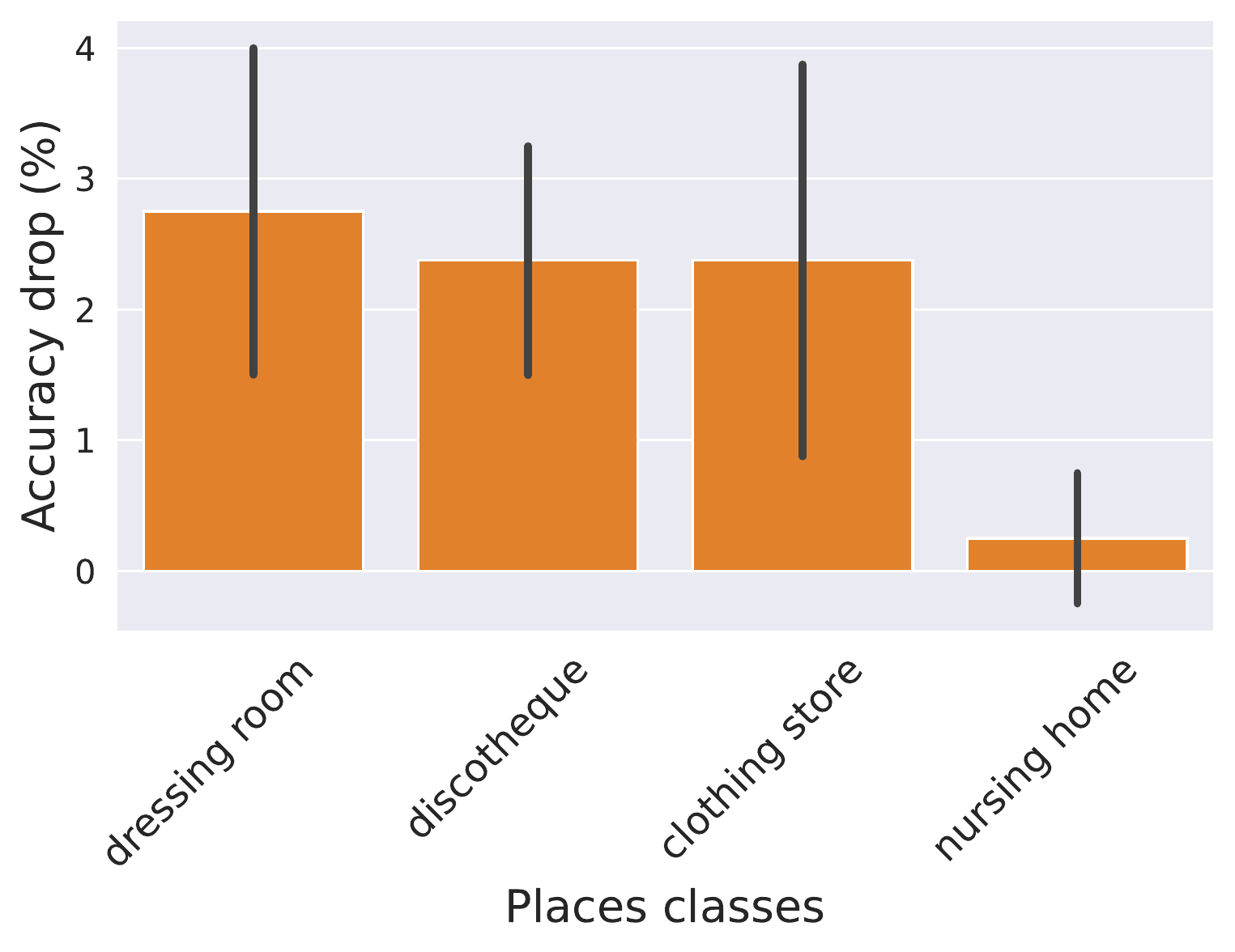}
	\caption{Concept: ``dress''; Models: VGG16 
		(\emph{left}) and ResNet-18 (\emph{right})  trained on Places-365. }
\end{subfigure}
\begin{subfigure}[b]{0.9\textwidth}
	\centering
	\includegraphics[width=4.5cm]{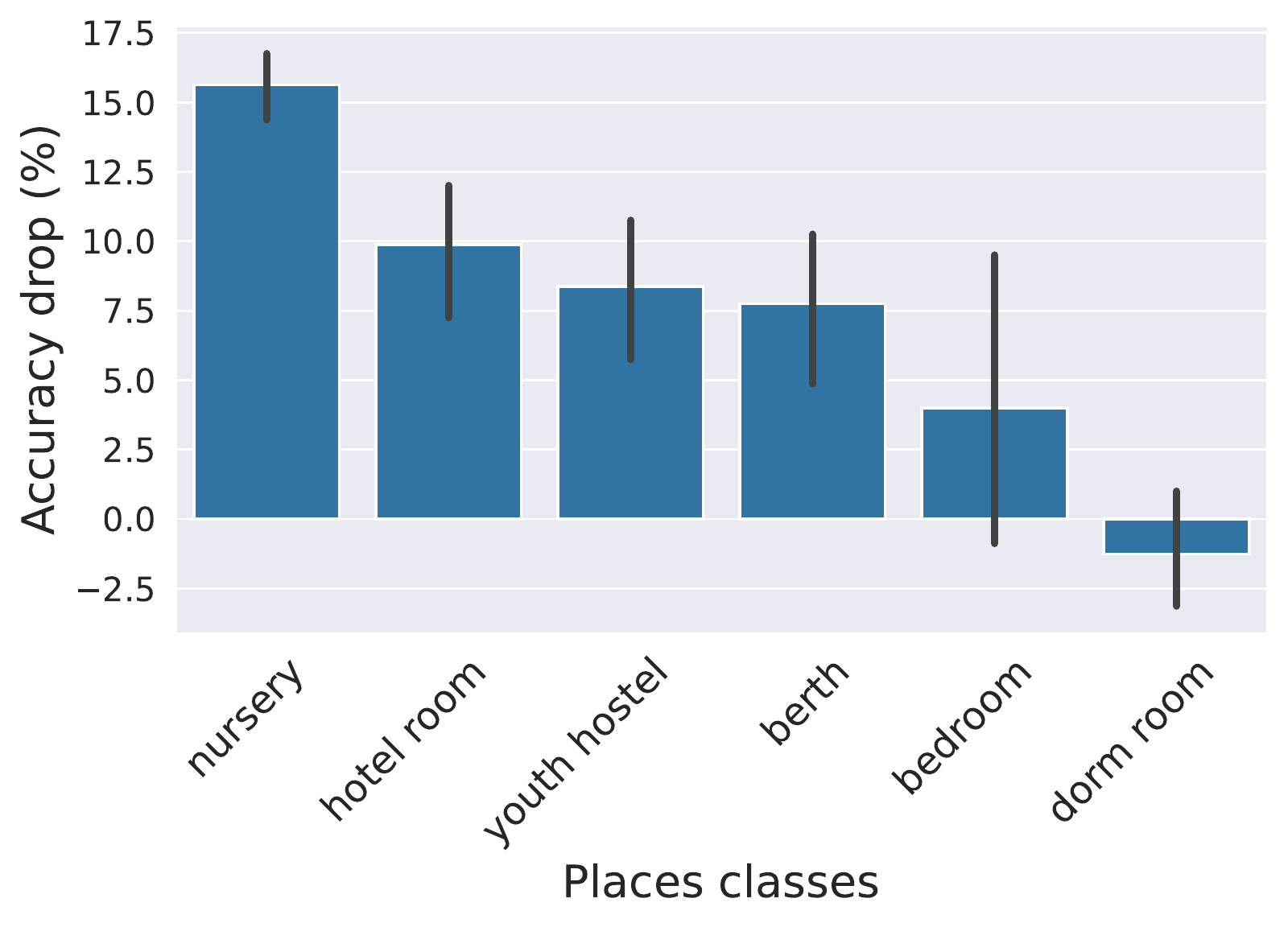}
	 \hfil
	\includegraphics[width=4.5cm]{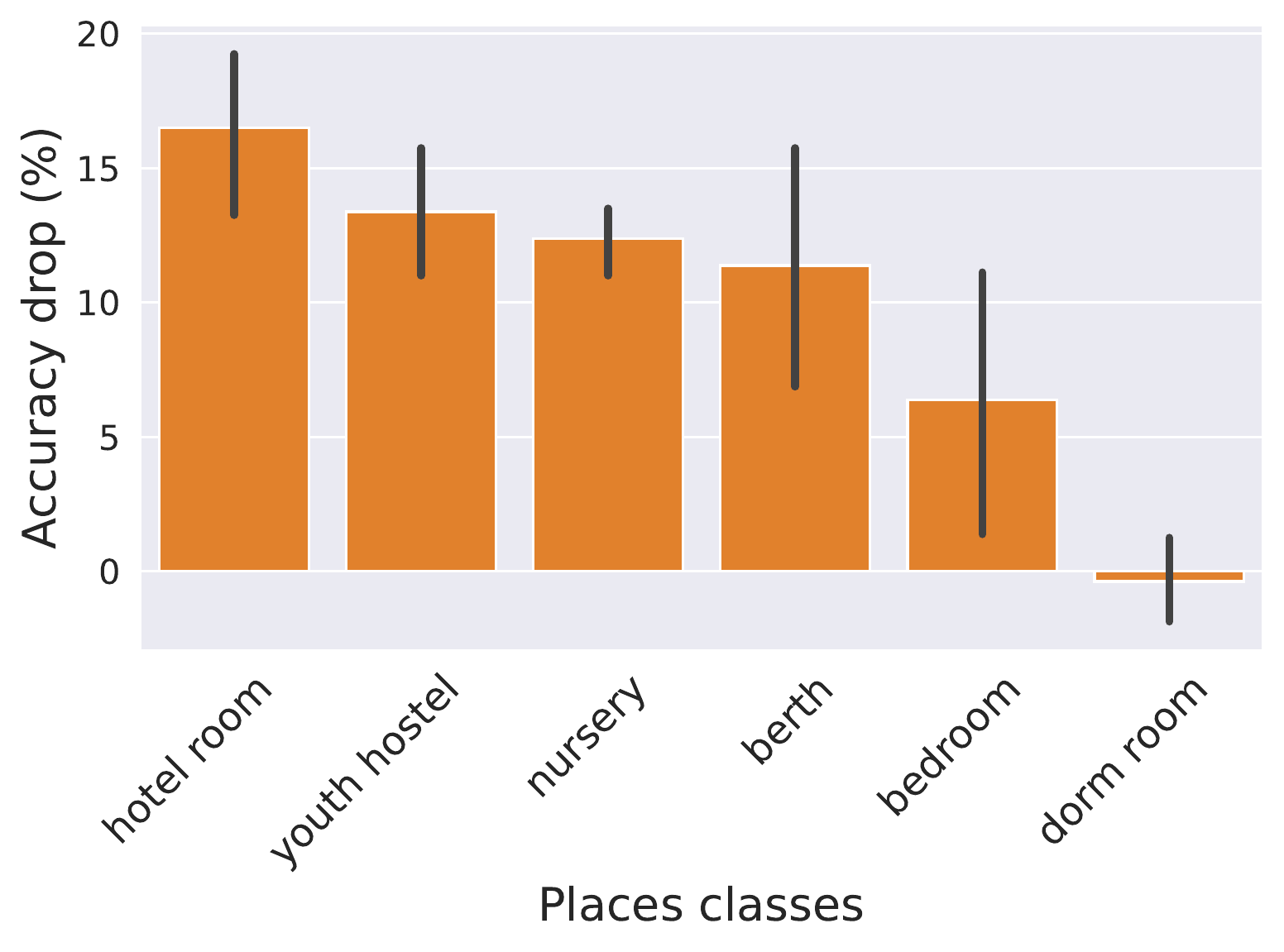}
	\caption{Concept: ``bed''; Models: VGG16 
		(\emph{left}) and ResNet-18 (\emph{right}) trained on Places-365. }
\end{subfigure}
\begin{subfigure}[b]{0.9\textwidth}
	\centering
	\includegraphics[width=4.5cm]{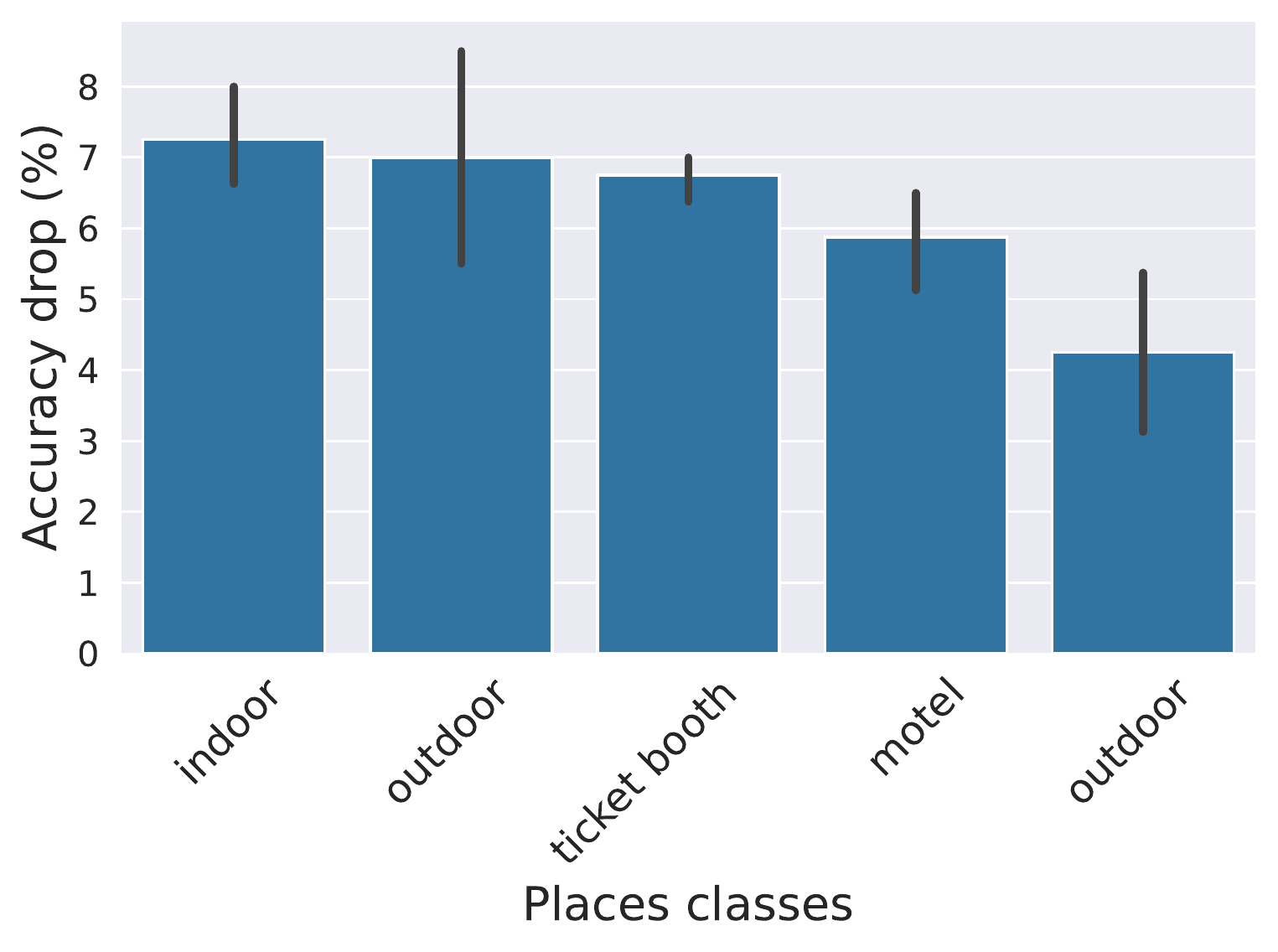}
	\hfil
	\includegraphics[width=4.5cm]{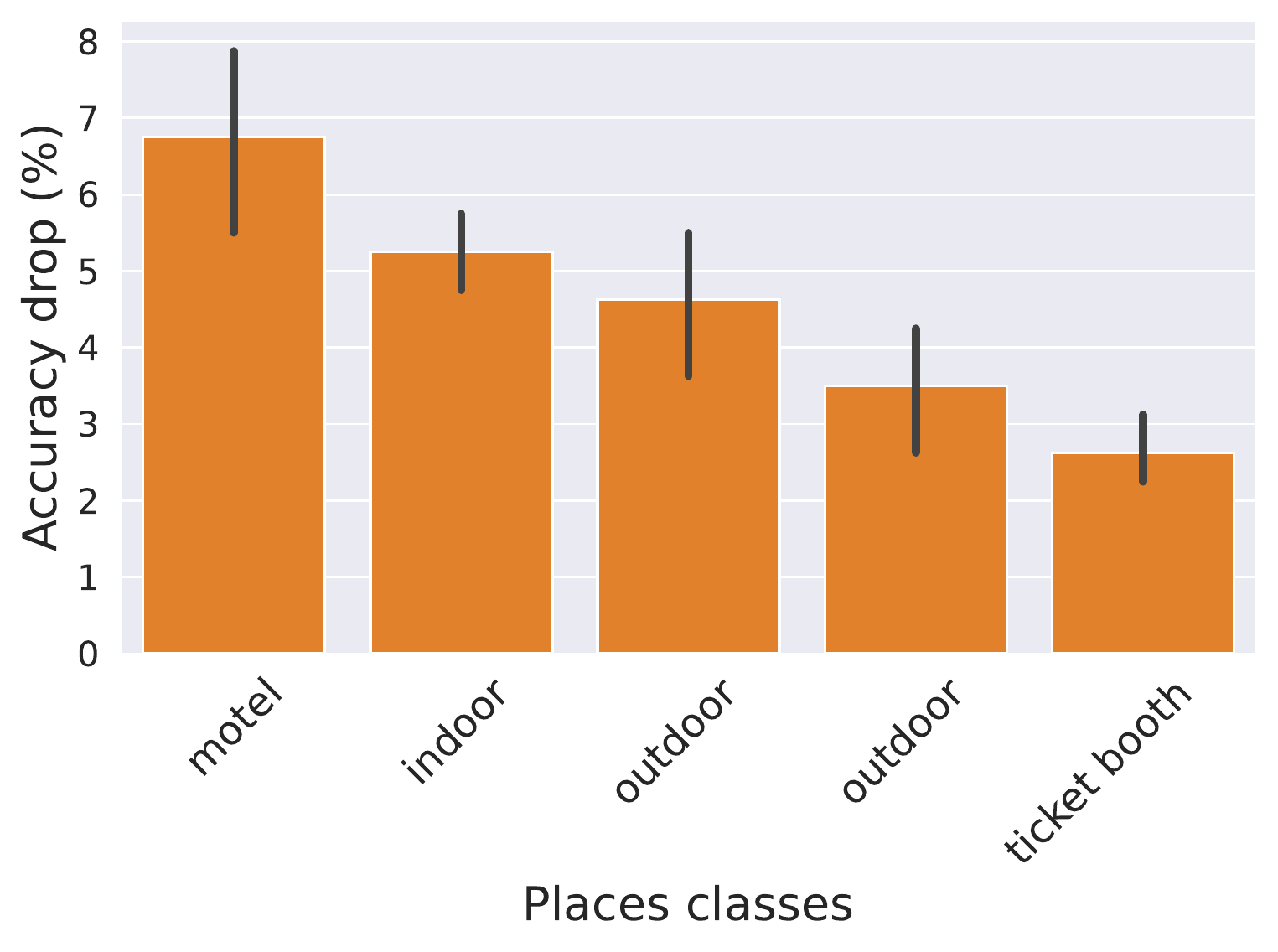}
	\caption{Concept: ``signboard''; Models: VGG16 
		(\emph{left}) and ResNet-18 (\emph{right}) trained on Places-365. }
\end{subfigure}
	\caption{Dependence of a classifier on high-level concepts: 
	average accuracy drop (along with 95\% confidence 
	intervals obtained via bootstrapping), 
	over various styles, induced by the transformation of said concept.
	The classes for which the concept is most often 
	present are shown. }
	\label{fig:app_variation_across_concepts}
\end{figure}

\paragraph{Per concept.} In 
Figure~\ref{fig:app_variation_across_concepts}, we visualize the accuracy 
drop induced by transformations of a specific concept for
classifiers trained on ImageNet and Places-365 (similar to 
Figure~\ref{fig:errors_examples}a). Here, the accuracy drop 
post-transformation is 
measured only on images that contain the concept of interest.
We then present the average drop across transformations, along with 95\% 
confidence intervals. 
We find that there is a large variance between: (i) a model's reliance on 
different concepts, and (ii) different model's reliance on a single 
concept.
For instance, the accuracy of a ResNet-50 ImageNet classifier drops by more 
than 30\% on the class \class{three-toed sloth} when ``tree''s in the image 
are modified, while the accuracy of a VGG16 model drops by less than 5\% 
under the same setup.

\begin{figure}[!h]
	\centering
	\begin{subfigure}[b]{0.9\textwidth}
		\centering
		\includegraphics[width=0.4\columnwidth]{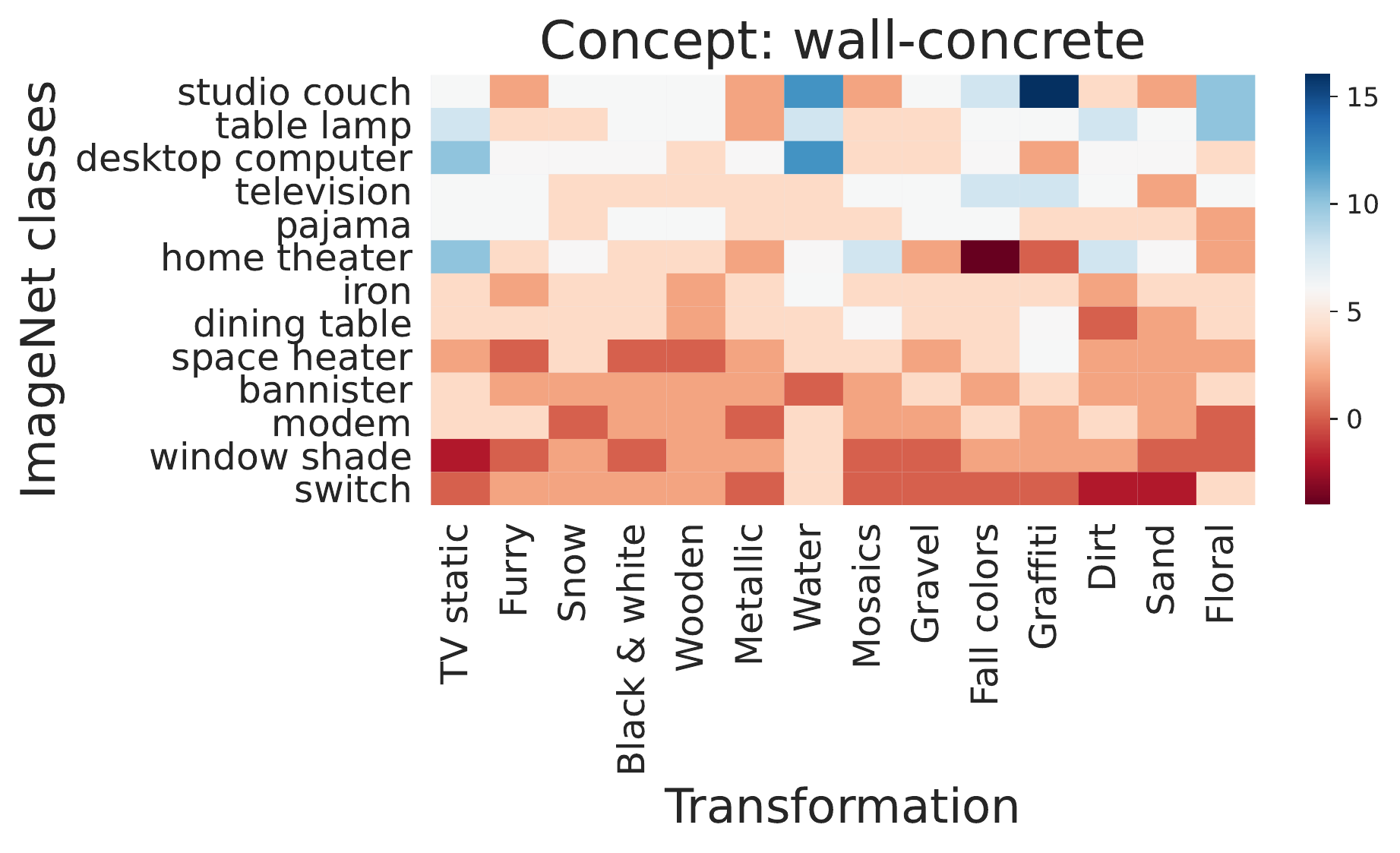}
		\includegraphics[width=0.4\columnwidth]{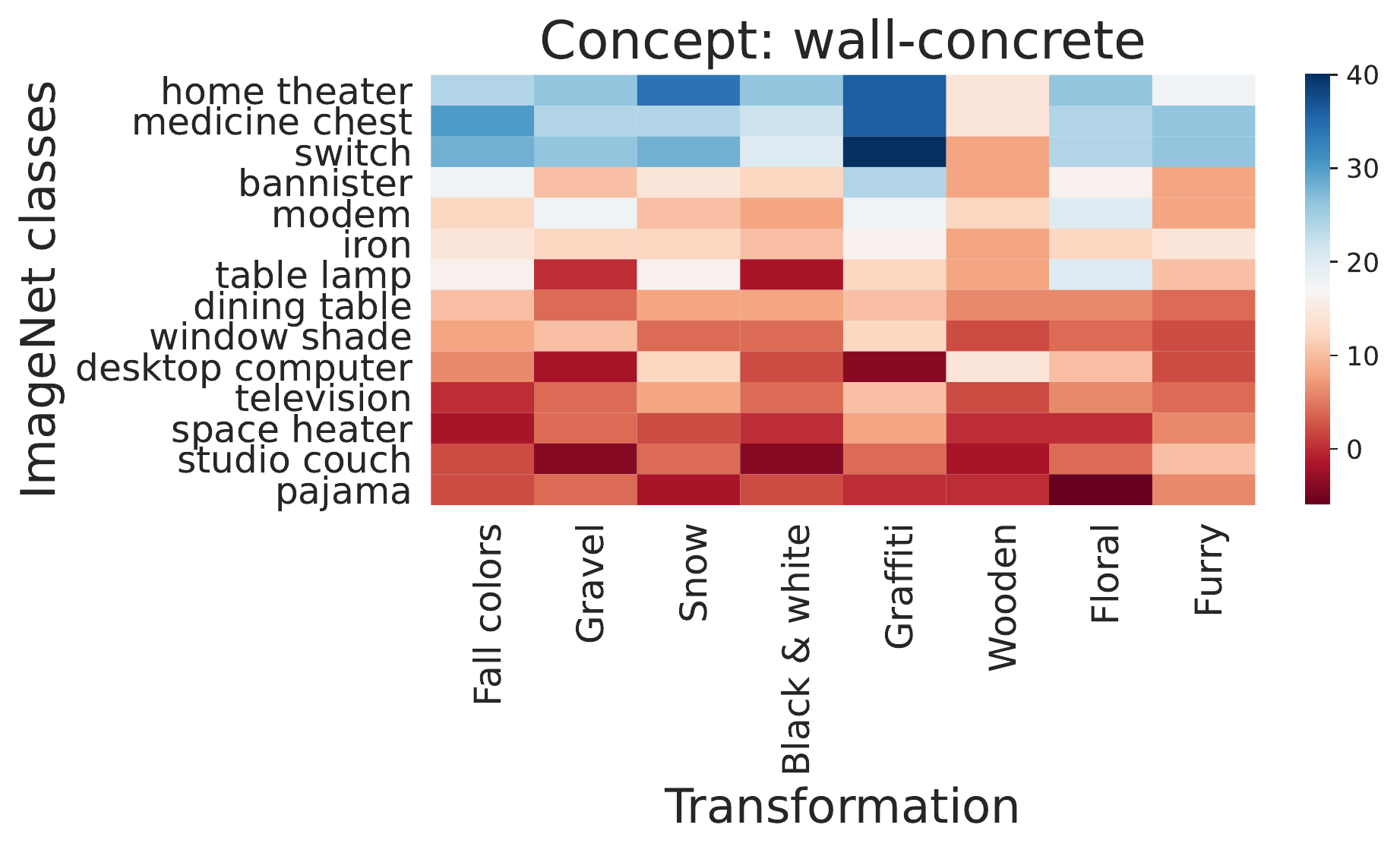}
	\end{subfigure}
\begin{subfigure}[b]{0.9\textwidth}
	\centering
	\includegraphics[width=0.4\columnwidth]{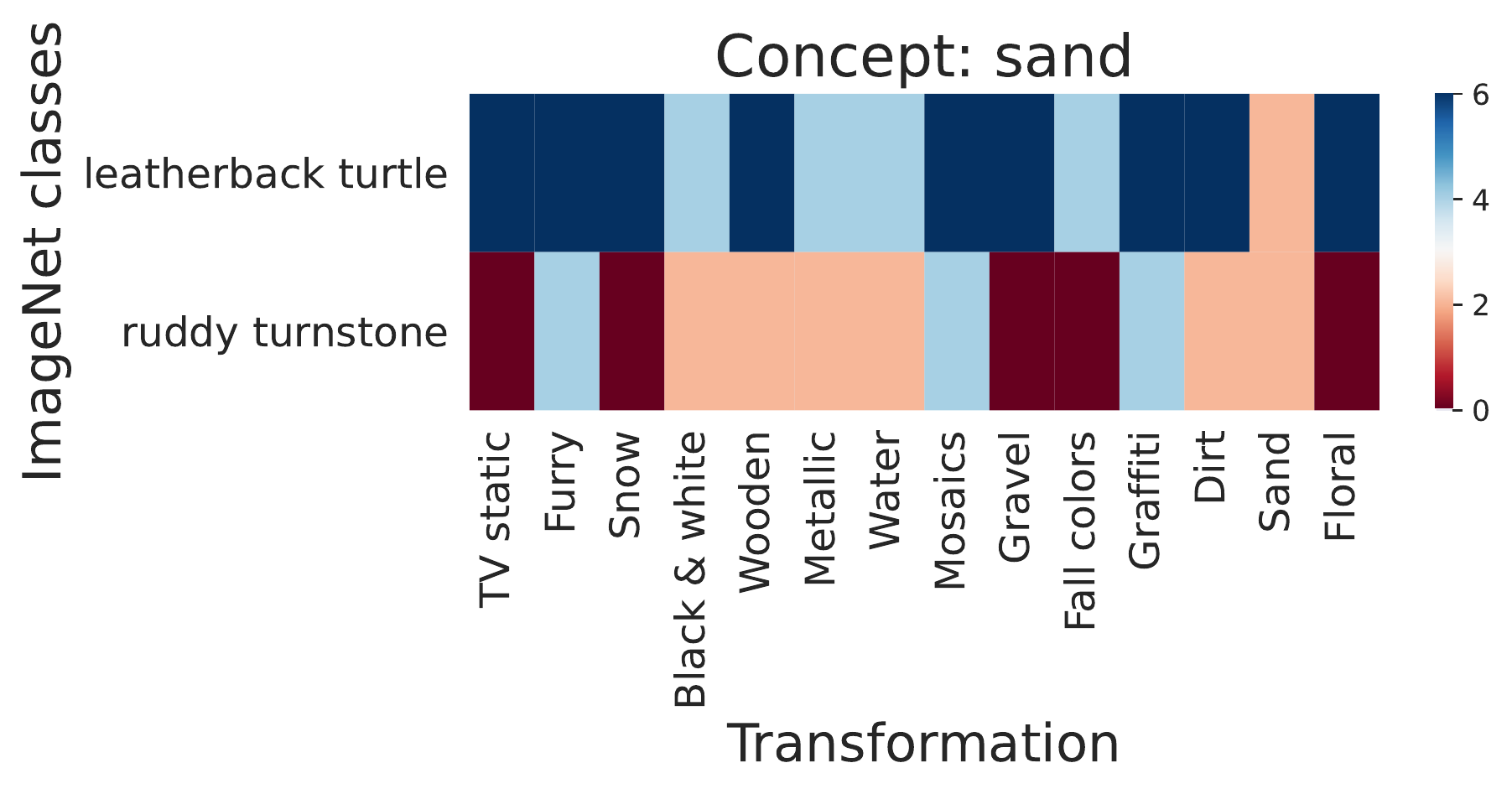}
	\includegraphics[width=0.4\columnwidth]{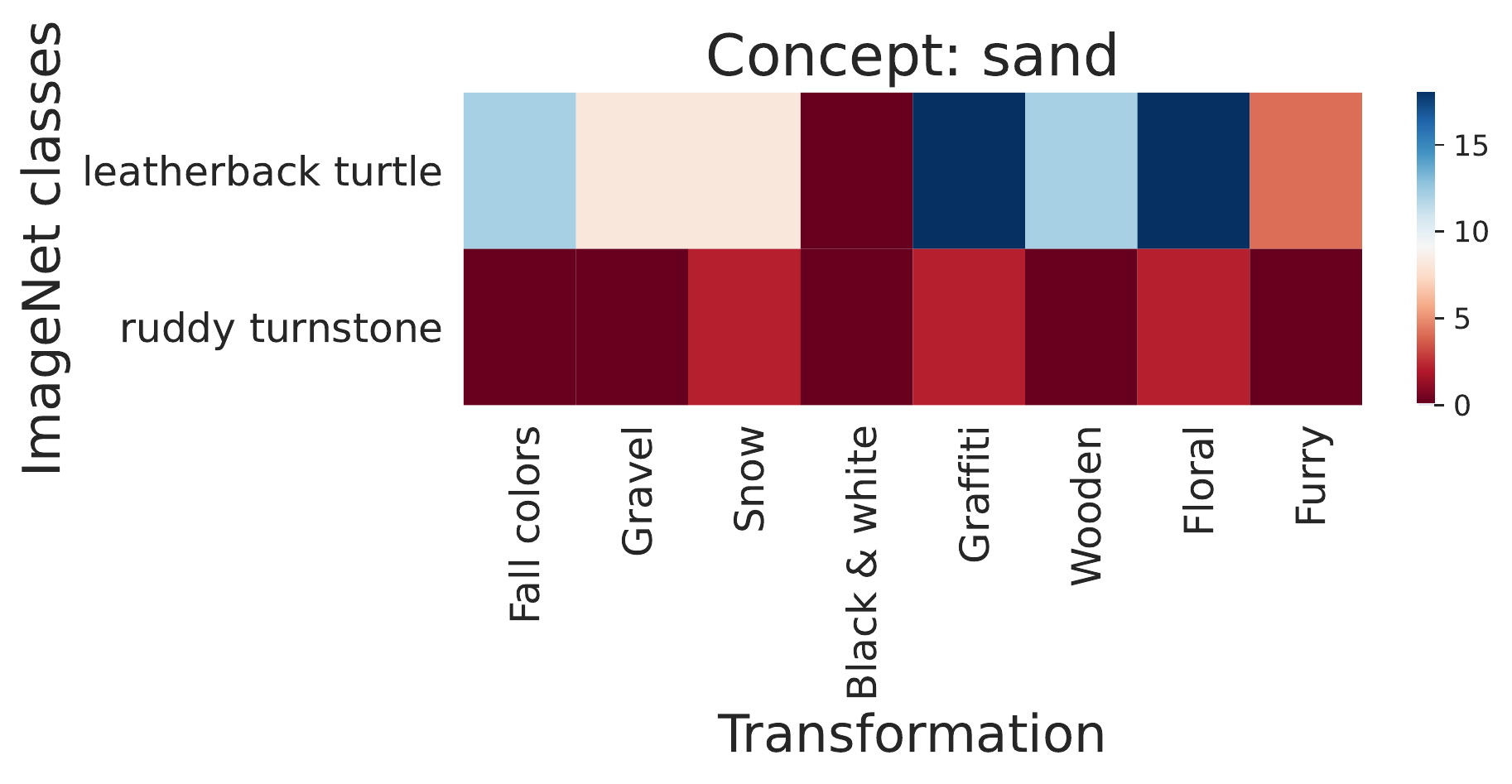}
\end{subfigure}
 	\begin{subfigure}[b]{0.9\textwidth}
 		\centering
 		\includegraphics[width=0.4\columnwidth]{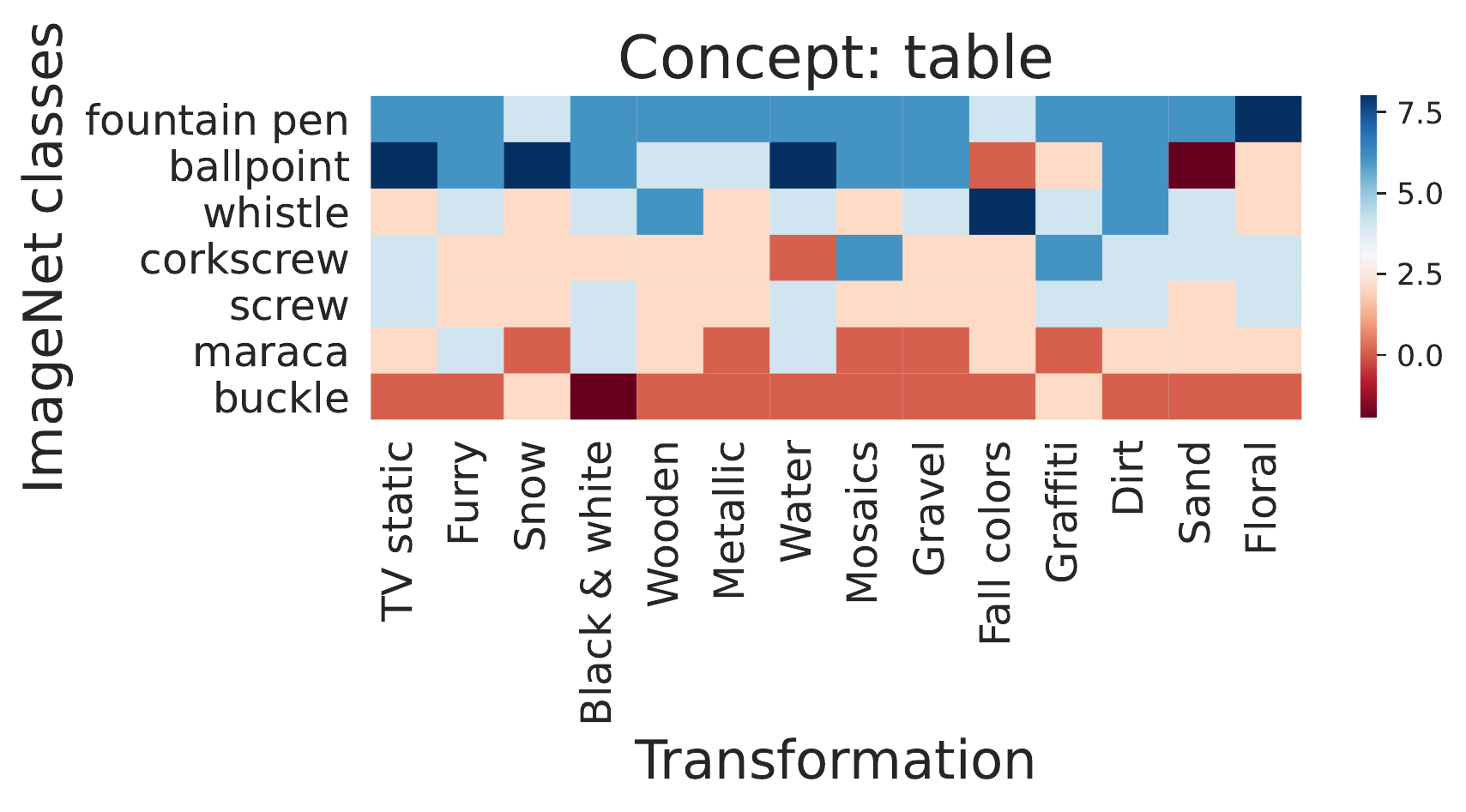}
 		\includegraphics[width=0.4\columnwidth]{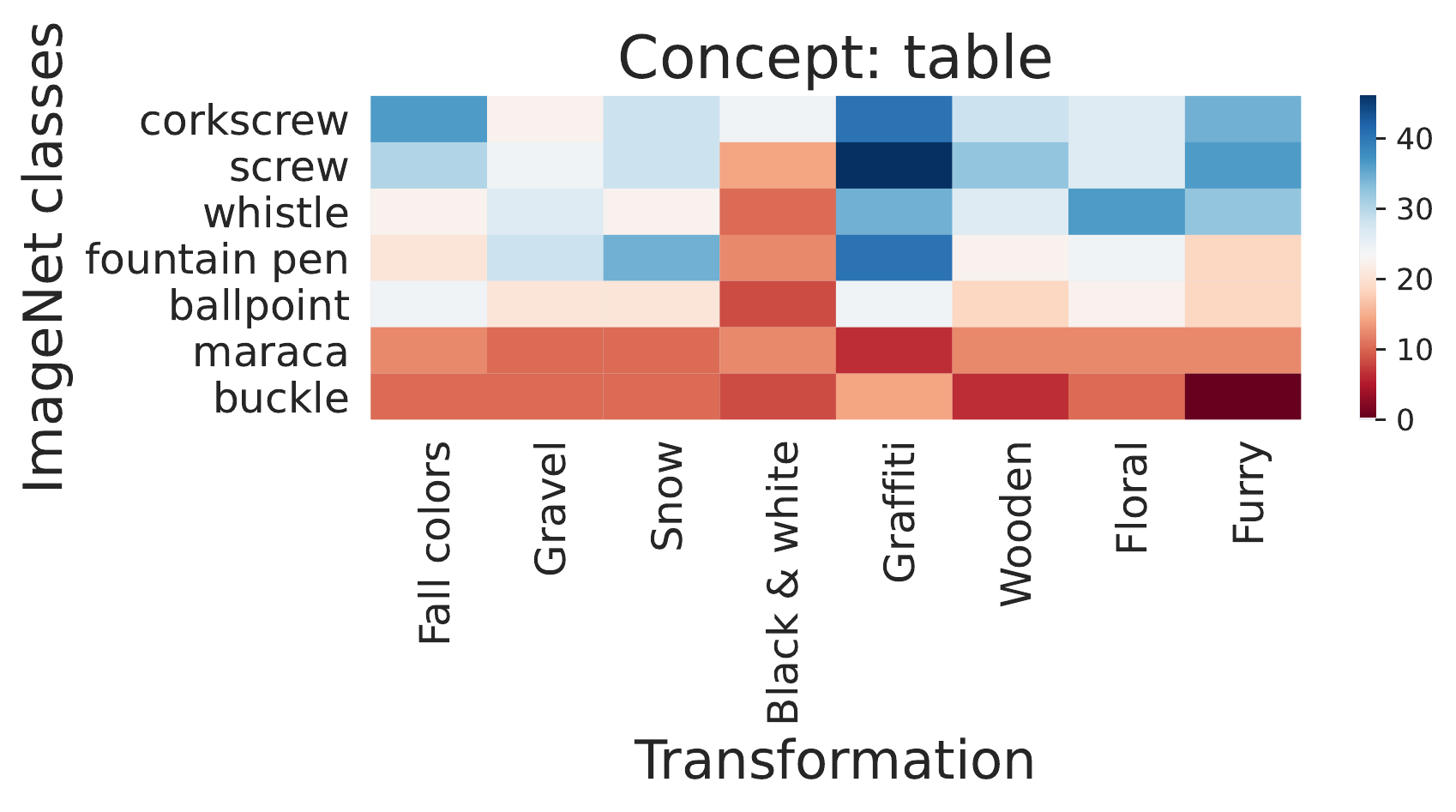}
 		\caption{VGG16 (\textit{left}) 
 			and ResNet-50 (\textit{right}) models trained on the ImageNet-1k 
 			dataset.}
 \end{subfigure} \\ 
	\begin{subfigure}[b]{0.9\textwidth}
	\centering
	\includegraphics[width=0.4\columnwidth]{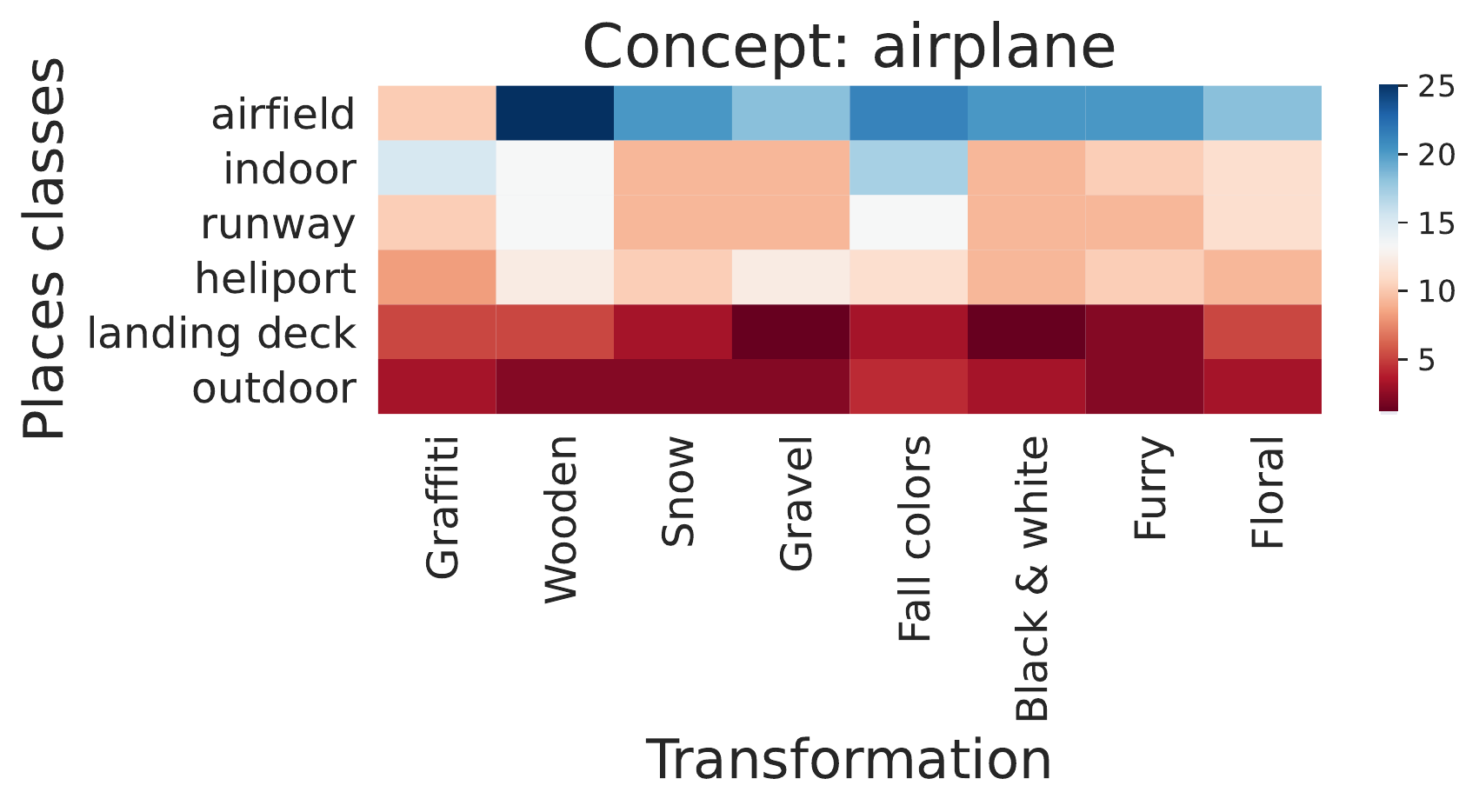}
	\includegraphics[width=0.4\columnwidth]{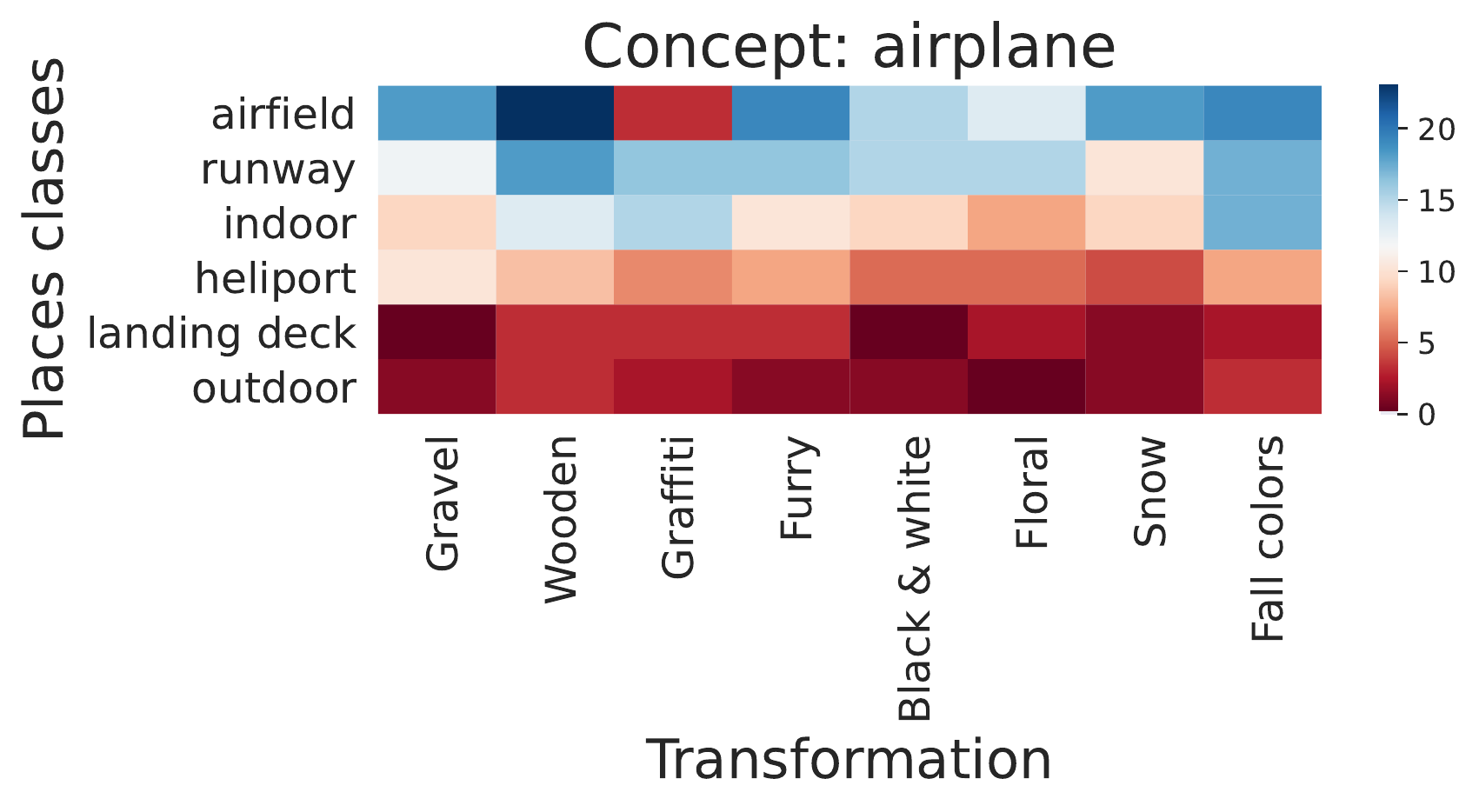}
\end{subfigure}
\begin{subfigure}[b]{0.9\textwidth}
	\centering
	\includegraphics[width=0.4\columnwidth]{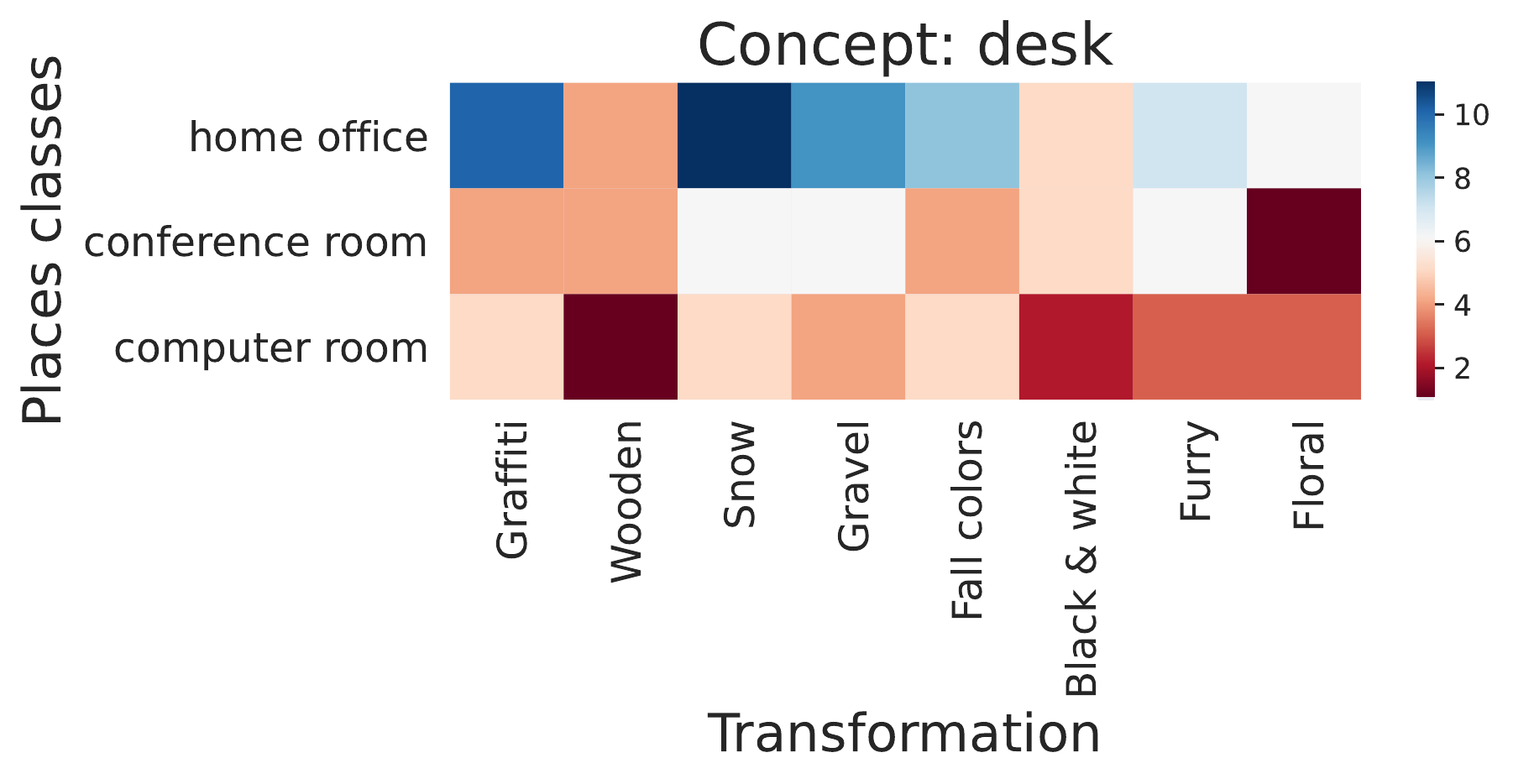}
	\includegraphics[width=0.4\columnwidth]{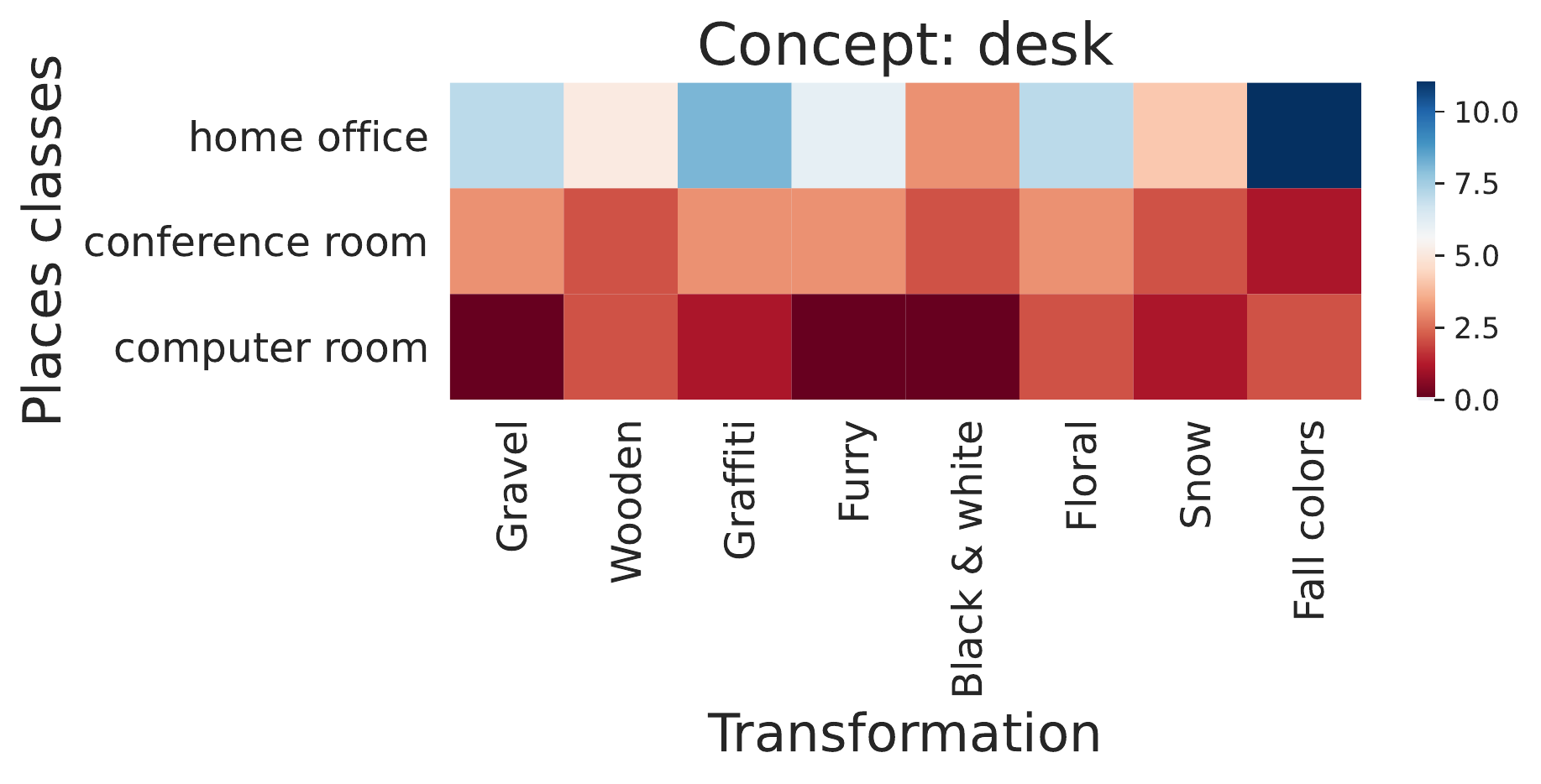}
\end{subfigure}
\begin{subfigure}[b]{0.9\textwidth}
	\centering
	\includegraphics[width=0.4\columnwidth]{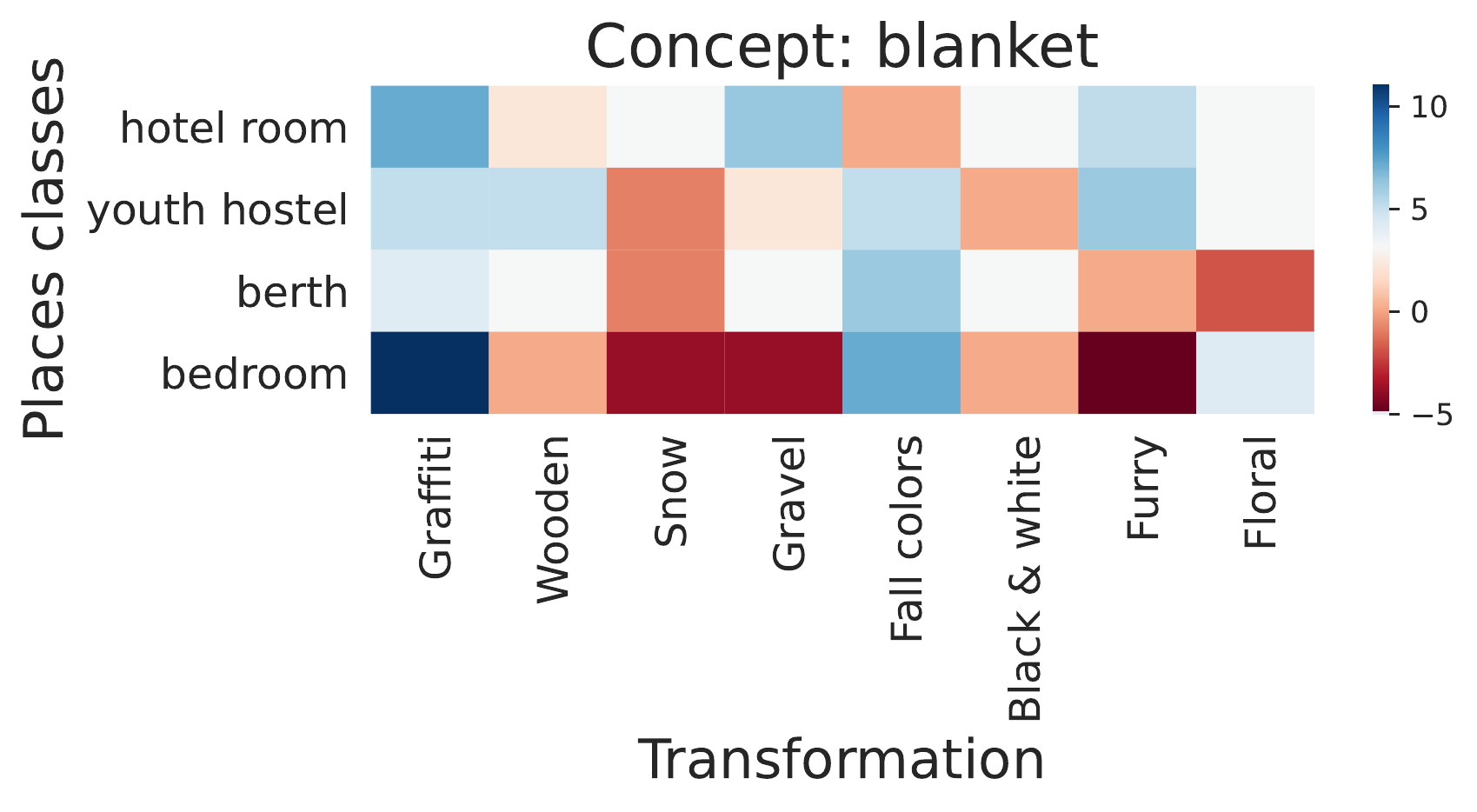}
	\includegraphics[width=0.4\columnwidth]{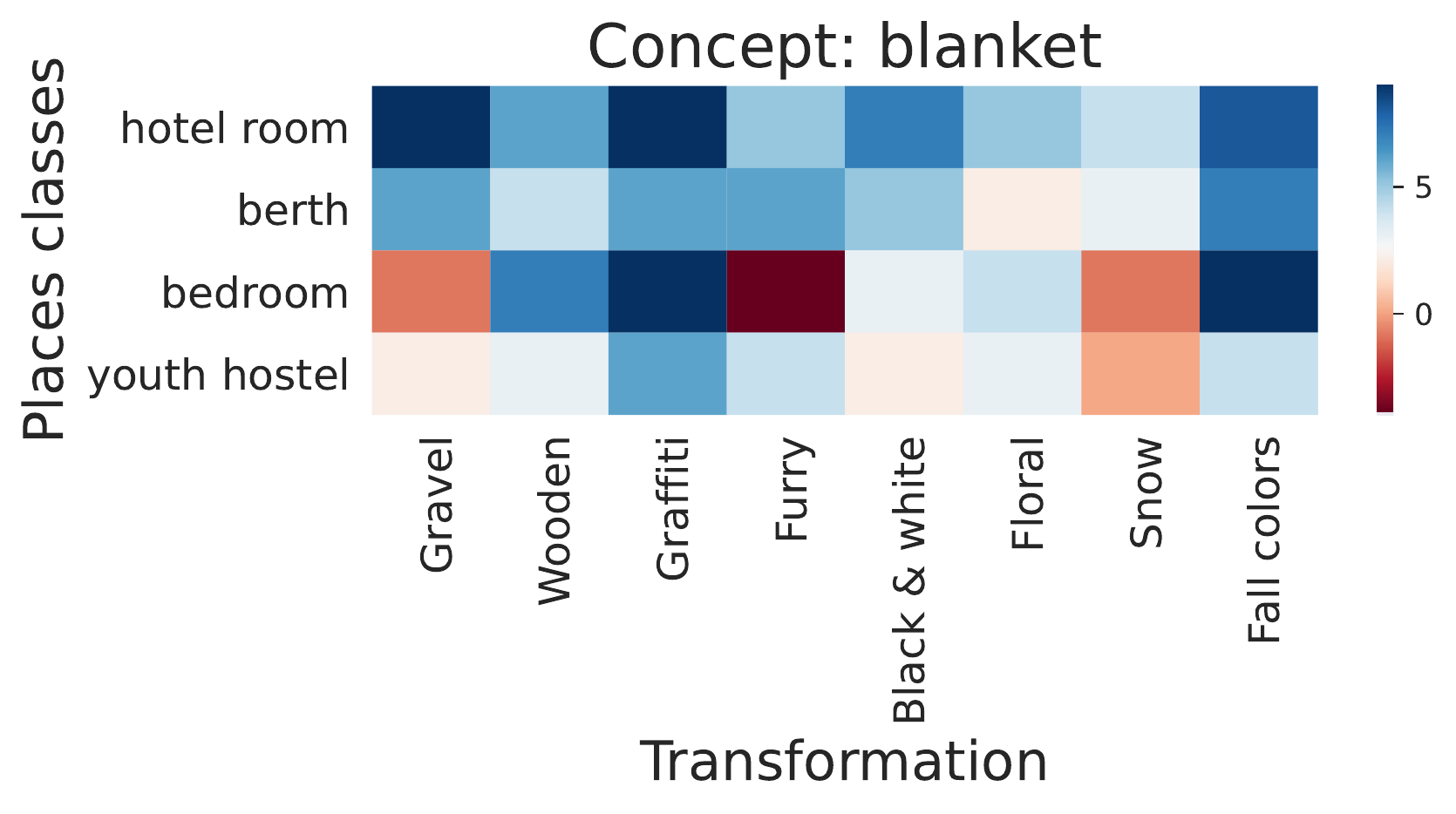}
	\caption{VGG16 (\textit{left}) 
		and ResNet-18 (\textit{right}) models trained on the Places-365 
		dataset.}
\end{subfigure}

	\caption{Heatmaps illustrating classifier sensitivity to various concept-level 
	transformations. Here, we measure model sensitivity in terms of the per 
	class drop 
	in model accuracy induced by the transformation on images of that 
	class which contain the concept of interest.
	}
	\label{fig:app_variation_across_styles}
\end{figure}

\paragraph{Per transformation.} In 
Figure~\ref{fig:app_variation_across_styles}, we illustrate how the model's 
sensitivity to specific concepts varies depending on the applied 
transformation. Across concepts, we find that models are more sensitive to 
transformations to textures such as ``grafitti'' and ``fall colors'' than they 
are to ``wooden'' or ``metallic''.

\begin{figure}[!h]
	\centering
	\begin{subfigure}[b]{0.95\textwidth}
		\centering
    \hfill
		\includegraphics[height=4.5cm]{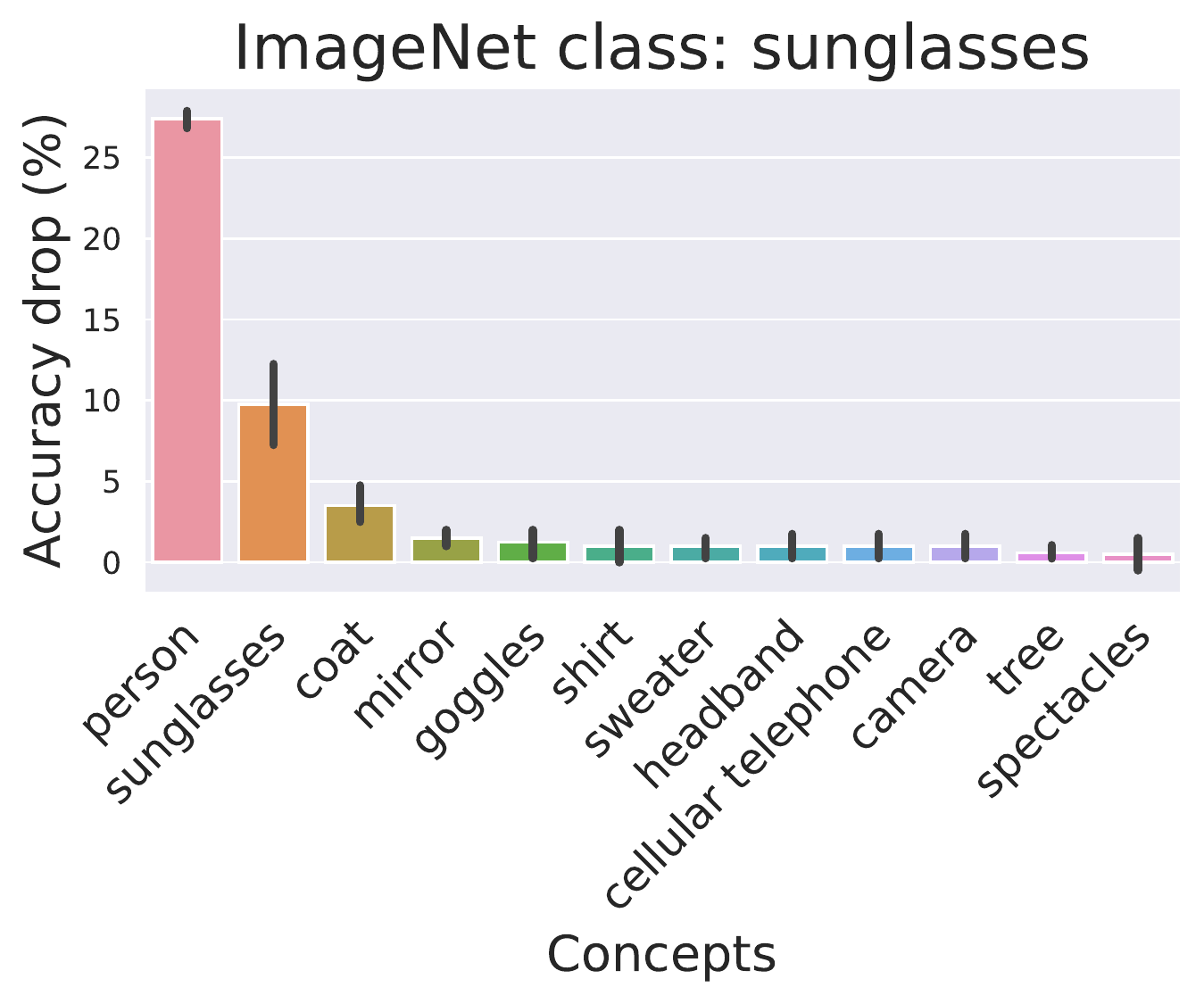}
	\hfill
		\includegraphics[height=4.5cm]{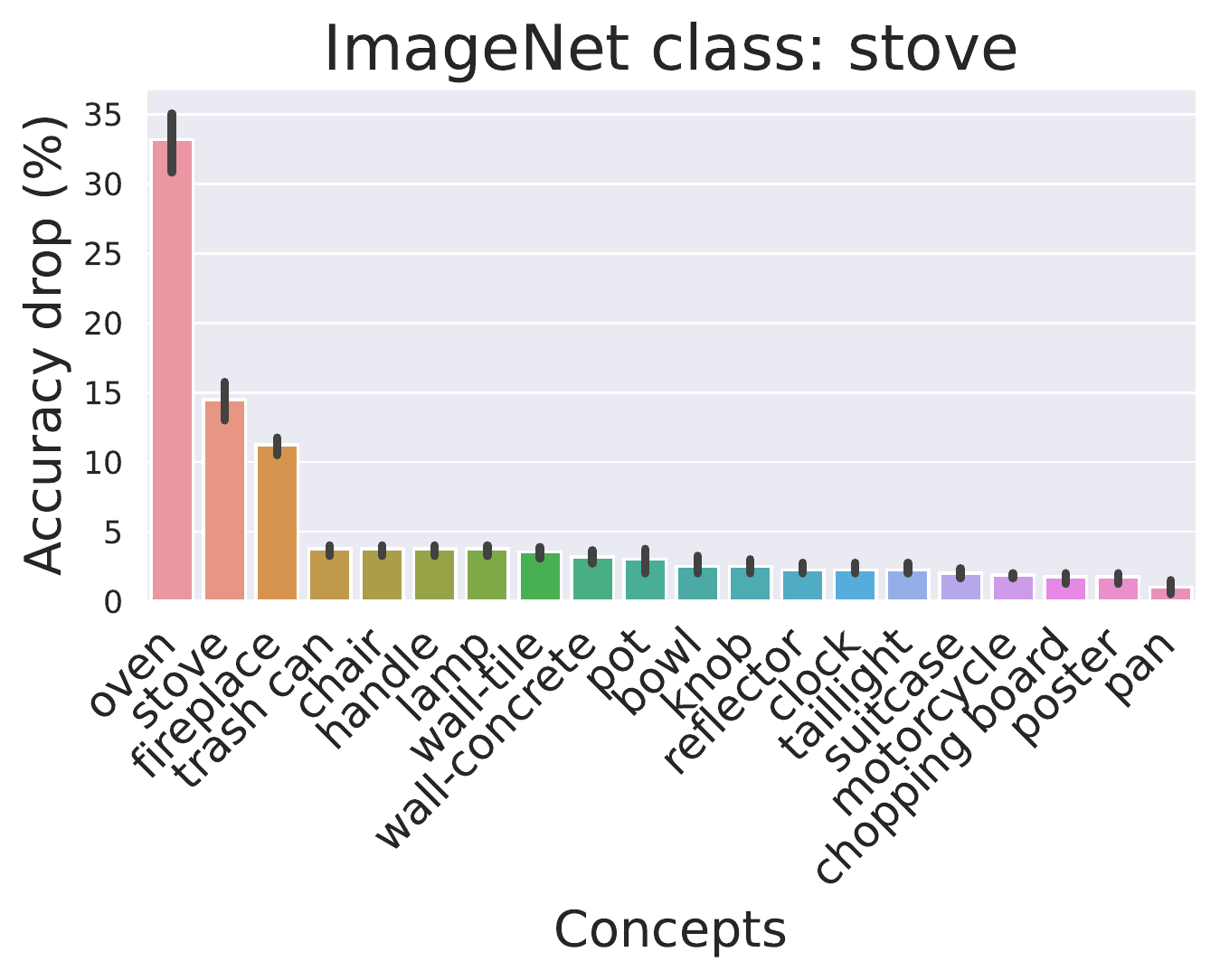}
	\hfill\phantom{}
	\end{subfigure}
	\begin{subfigure}[b]{0.95\textwidth}
	\centering
    \hfill
	\includegraphics[height=4.5cm]{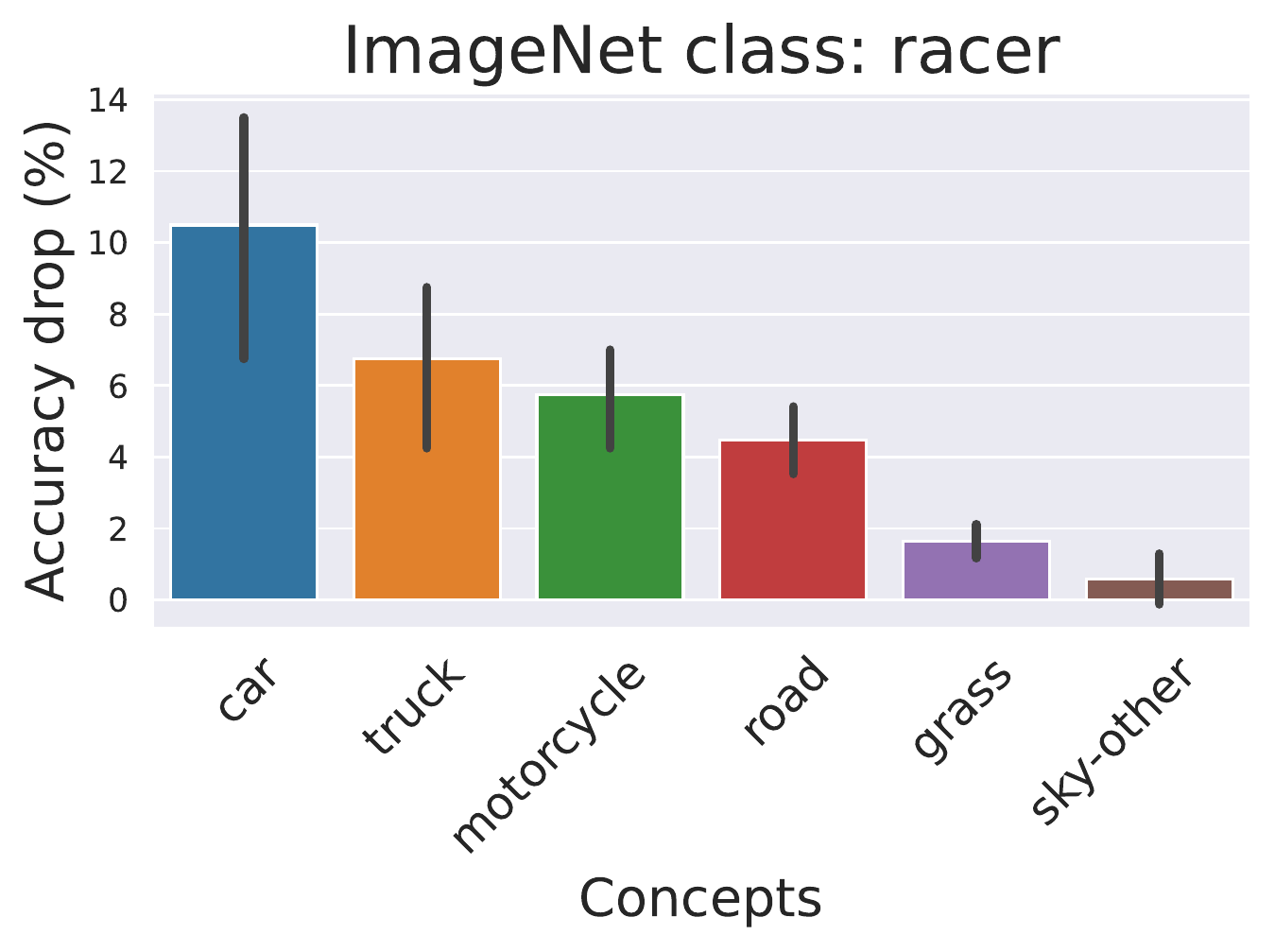}
	\hfill
	\includegraphics[height=4.5cm]{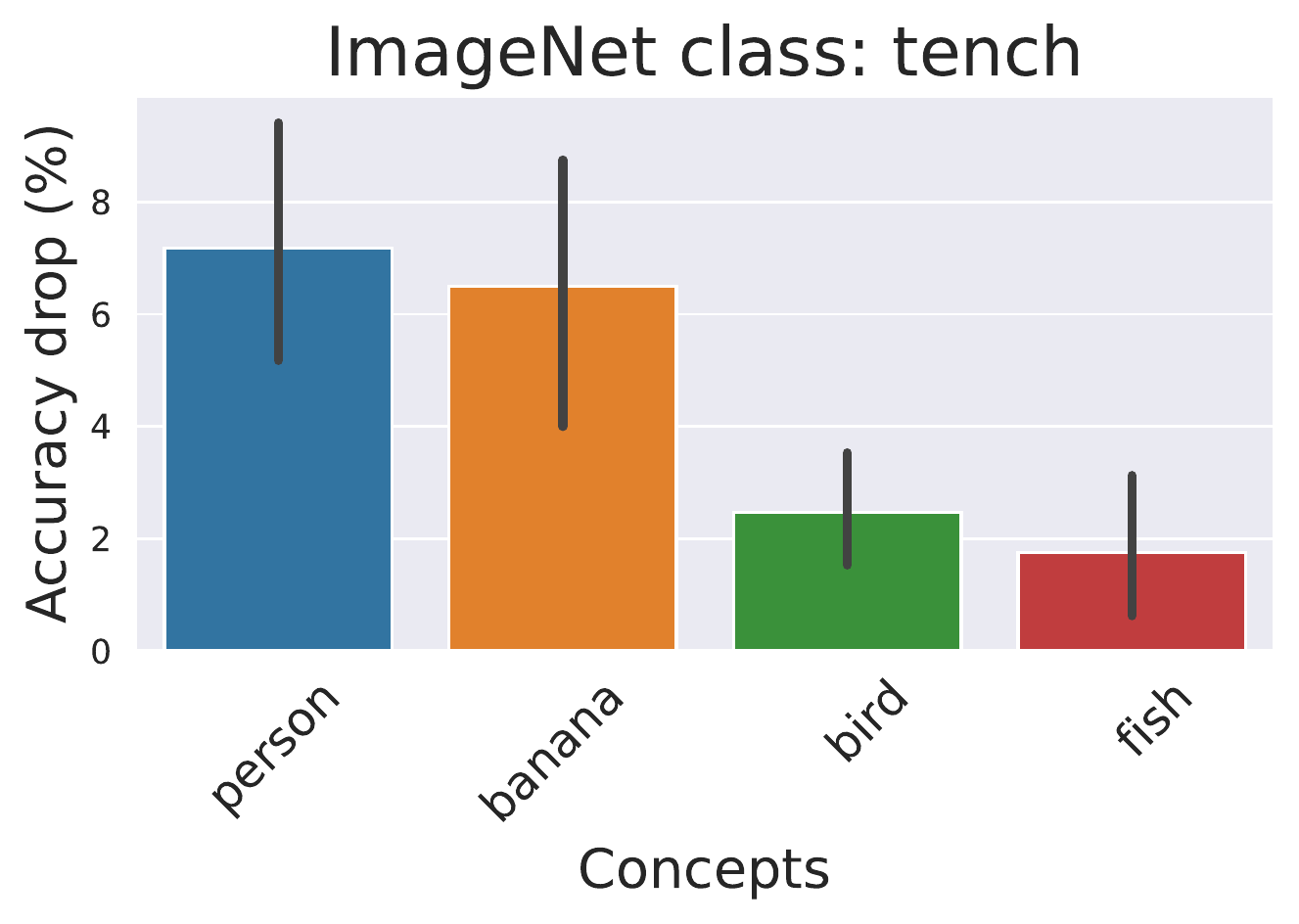}
	\hfill\phantom{}
	\caption{VGG16 classifier trained on the ImageNet
		dataset. }
\end{subfigure} 
	\begin{subfigure}[b]{0.95\textwidth}
	\centering
	\hfill
	\includegraphics[height=4.5cm]{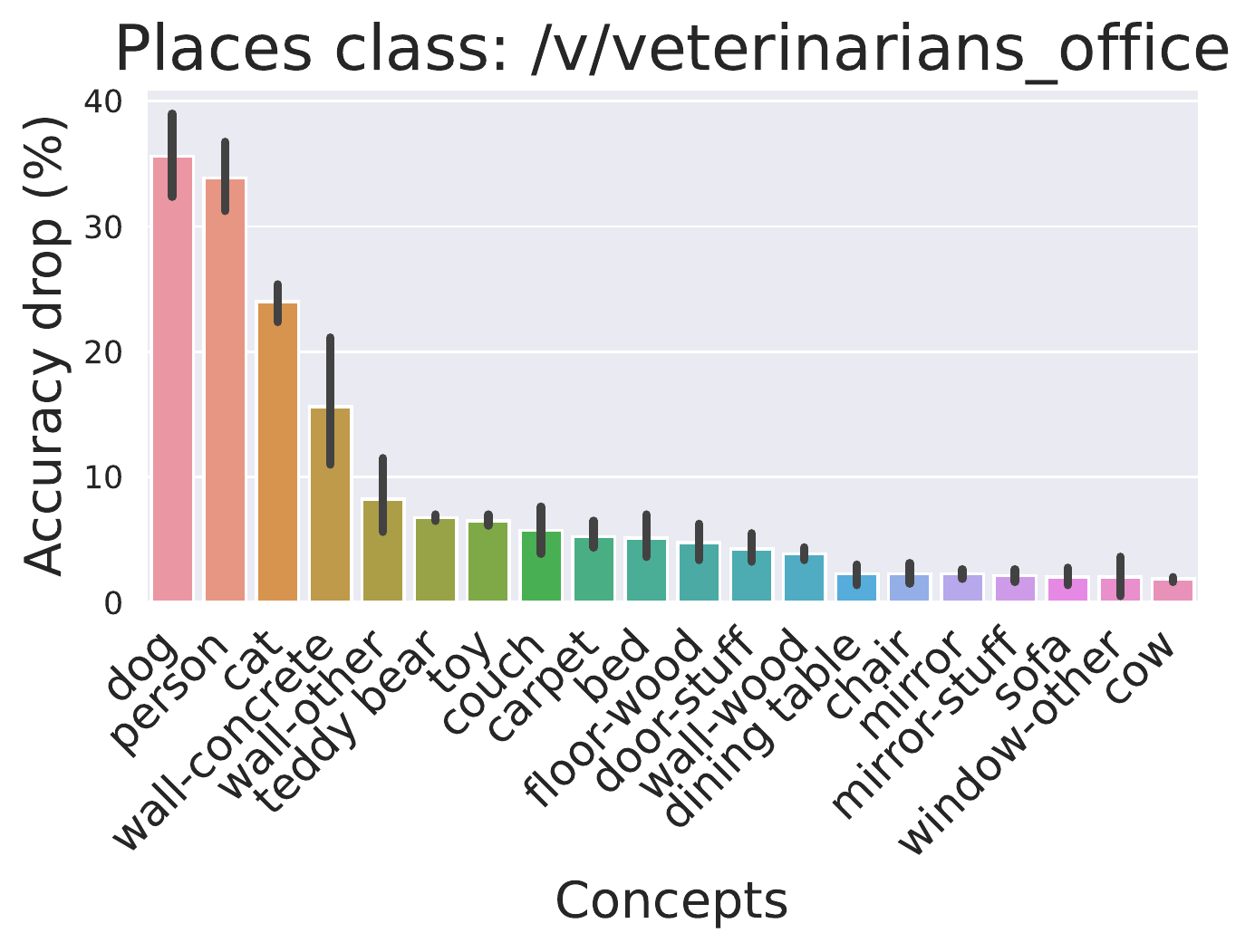}
	\hfill
	\includegraphics[height=4.5cm]{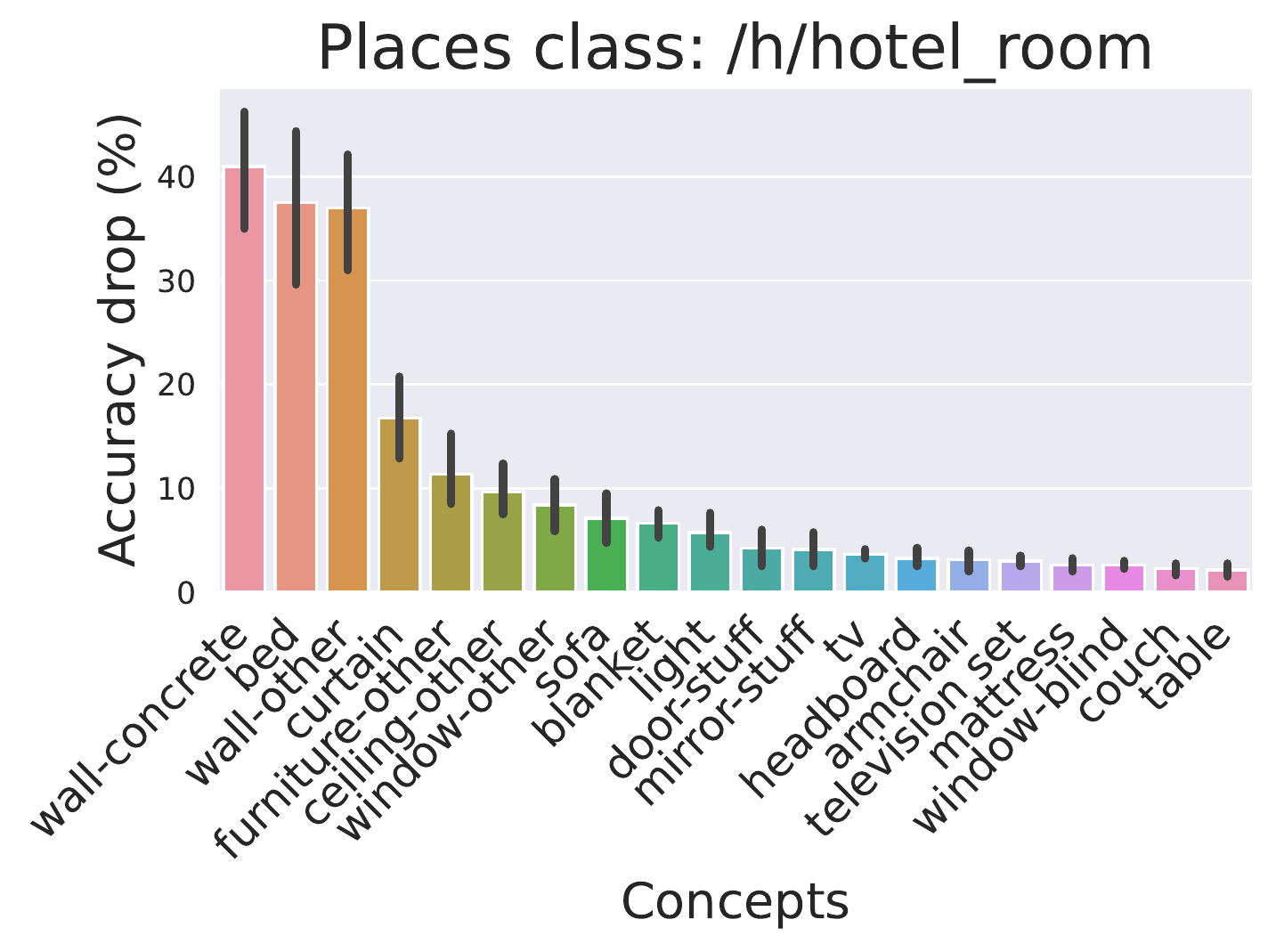}
	\hfill\phantom{}
\end{subfigure}
\begin{subfigure}[b]{0.95\textwidth}
	\centering
	\hfill
	\includegraphics[height=4.5cm]{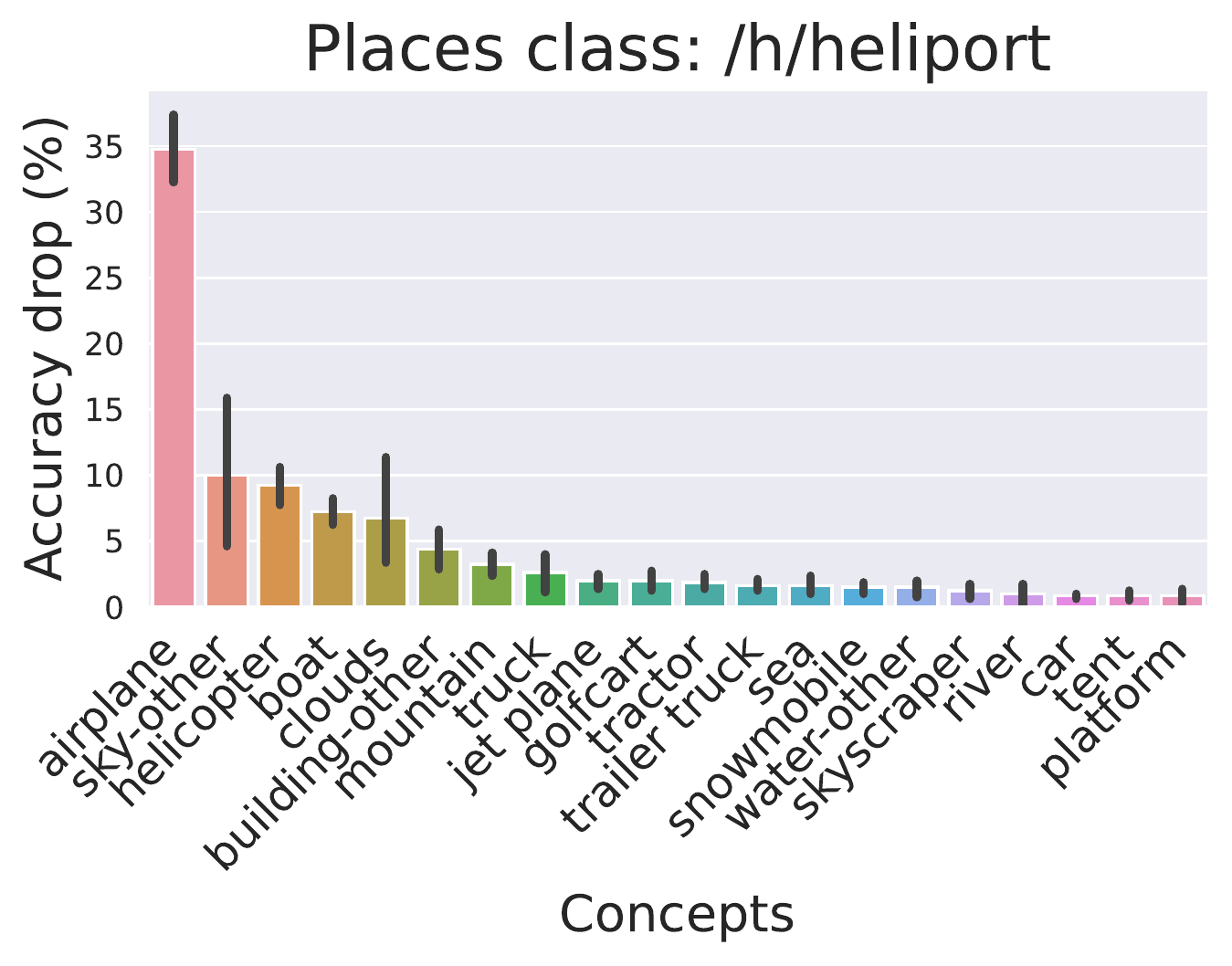}
	\hfill
	\includegraphics[height=4.5cm]{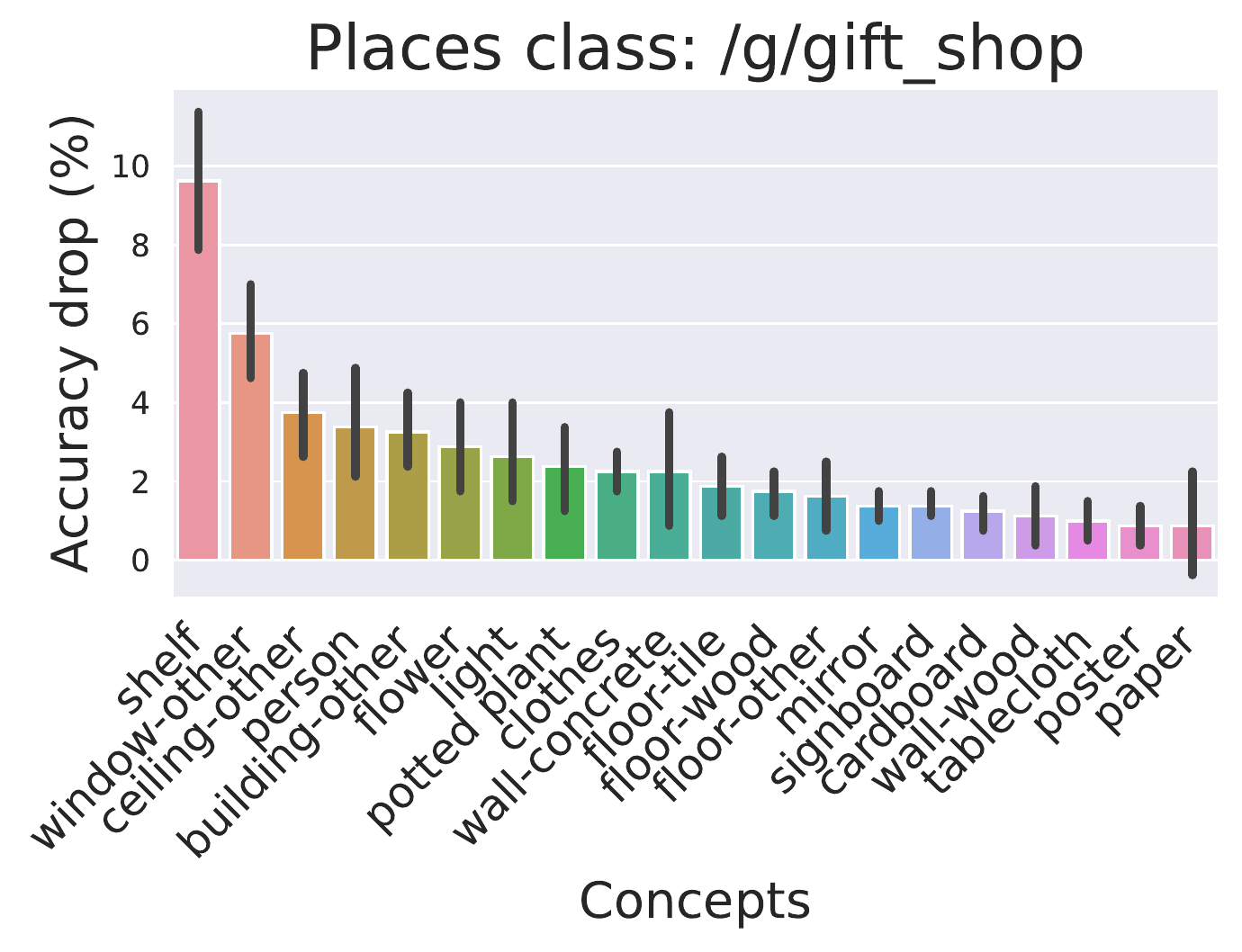}
	\hfill\phantom{}
	\caption{ResNet-18 classifier trained on the Places-365
		dataset.}
\end{subfigure}
	\caption{Per-class prediction rules: high-level concepts, which when 
	transformed, significantly hurt model performance on that class. Here, we 
	visualize average accuracy drop (along with 95\% confidence 
	intervals obtained via bootstrapping) for a specific concept, over various 
	styles.}
	\label{fig:app_debugging_classes}
\end{figure}

\paragraph{Per class (prediction-rules).} In 
Figure~\ref{fig:app_debugging_classes}, we provide additional examples of 
class-level prediction rules identified using our methodology. Specifically, for 
each class,  the highlighted concepts are those that hurt model 
accuracy when transformed.

%% file: sections/app_editing.tex
\subsection{Fine-grained model behavior on typographic attacks}
\begin{figure}[!h]
	\begin{subfigure}[b]{\textwidth}
		\centering
        \includegraphics[width=.8\textwidth]{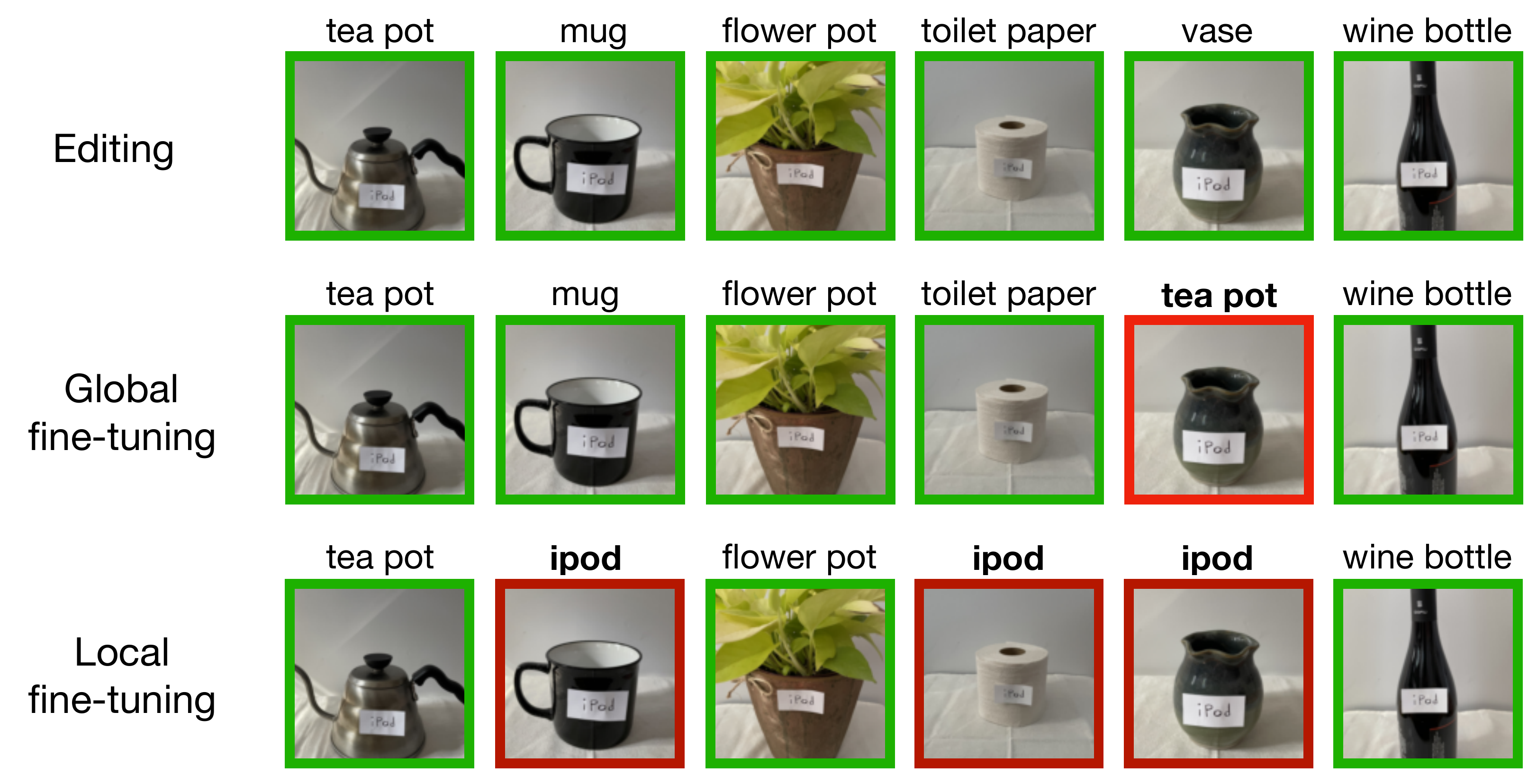}
        \caption{}
	\end{subfigure}
	\begin{subfigure}[b]{\textwidth}
		\centering
        \includegraphics[width=.6\textwidth]{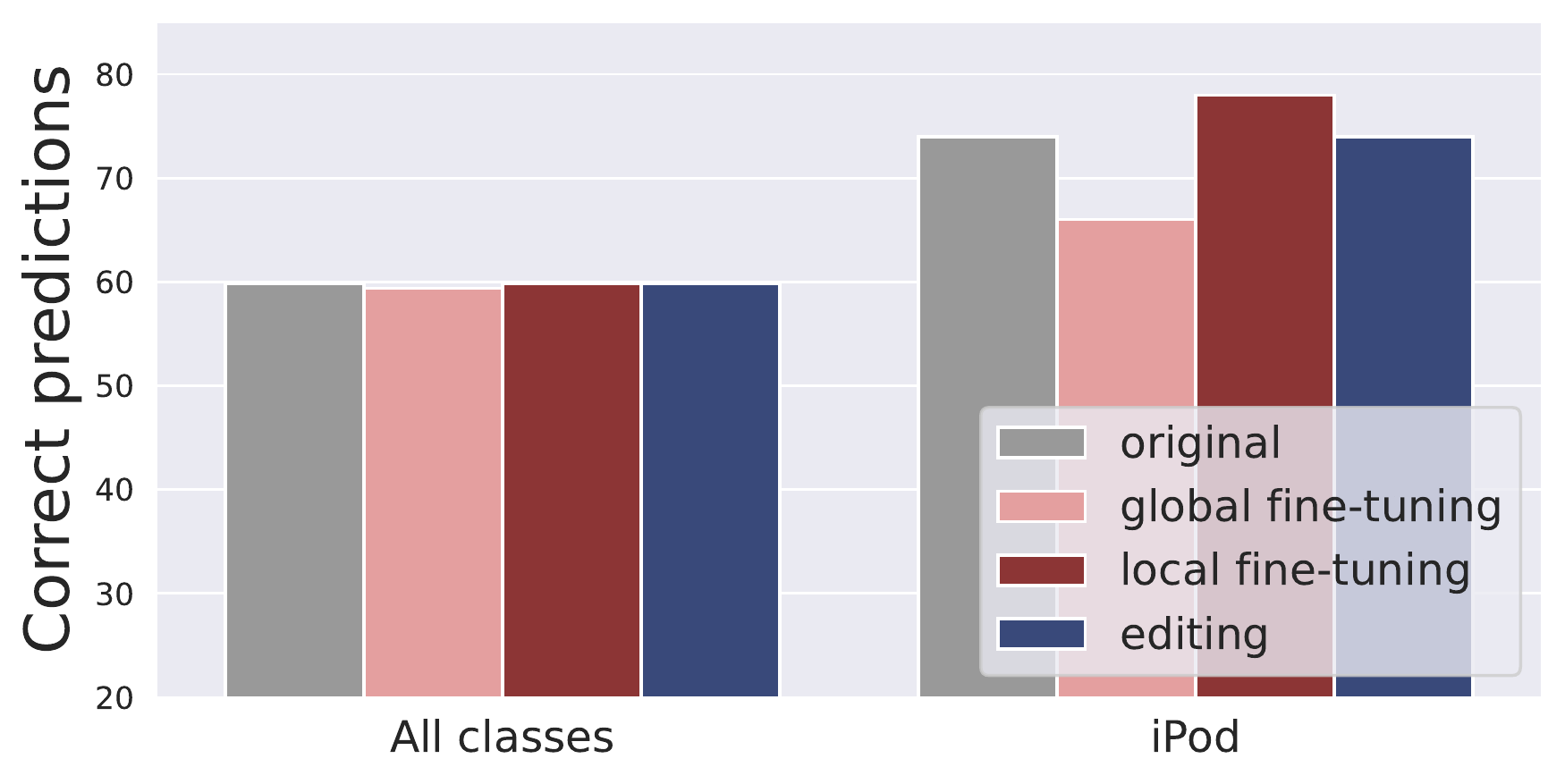}
        \caption{}
	\end{subfigure}
    \caption{Effectiveness of different modification procedures in preventing  
    typographic attacks. (a) Model predictions after the rewrite---local 
    fine-tuning often fails to prevent such attacks, while global fine-tuning 
    results in the model 
    associating
        ``iPod'' with the target class used for fine-tuning (\class{teapot}). (b)
        Accuracy on the original test set and specifically on clean samples from 
        class
        \class{ipod} before and after the rewrite. While global fine-tuning is 
        fairly effective at mitigating typographic attacks, it disproportionately
     reduces model accuracy on clean images from the \class{iPod} class.}
	\label{fig:app_ft_clip_failure}
\end{figure}

    In Figure~\ref{fig:app_ft_clip_failure}, we take a closer look at how effective
    different rewriting methods (with one train exemplar)
     are in mitigating typographic 
    attacks. 
    We find that:
    \begin{itemize}
    	\item \emph{Local-finetuning:} Corrects only a subset of the errors.
    	\item \emph{Global-finetuning:} Corrects most errors on the 
    	attacked images. However, on the flip side it: (i) causes 
    	the model to spuriously associate other images with the target class 
    	(\class{teapot}) used to perform fine-tuning and (ii) significantly 
    	reduces 
    	model accuracy on clean images of the class \class{iPod}.
    	\item \emph{Editing:} Corrects all errors without substantially hurting 
    	model accuracy on clean images.
    \end{itemize}

\clearpage
\subsubsection{The effectiveness of editing}
\begin{figure}[!h]
	\begin{subfigure}[b]{1\textwidth}
		\centering
		\includegraphics[width=0.45\columnwidth]{figures/editing/ImageNet_COCO_vgg16_0_3_25.pdf}
		 \hfil
		\includegraphics[width=0.45\columnwidth]{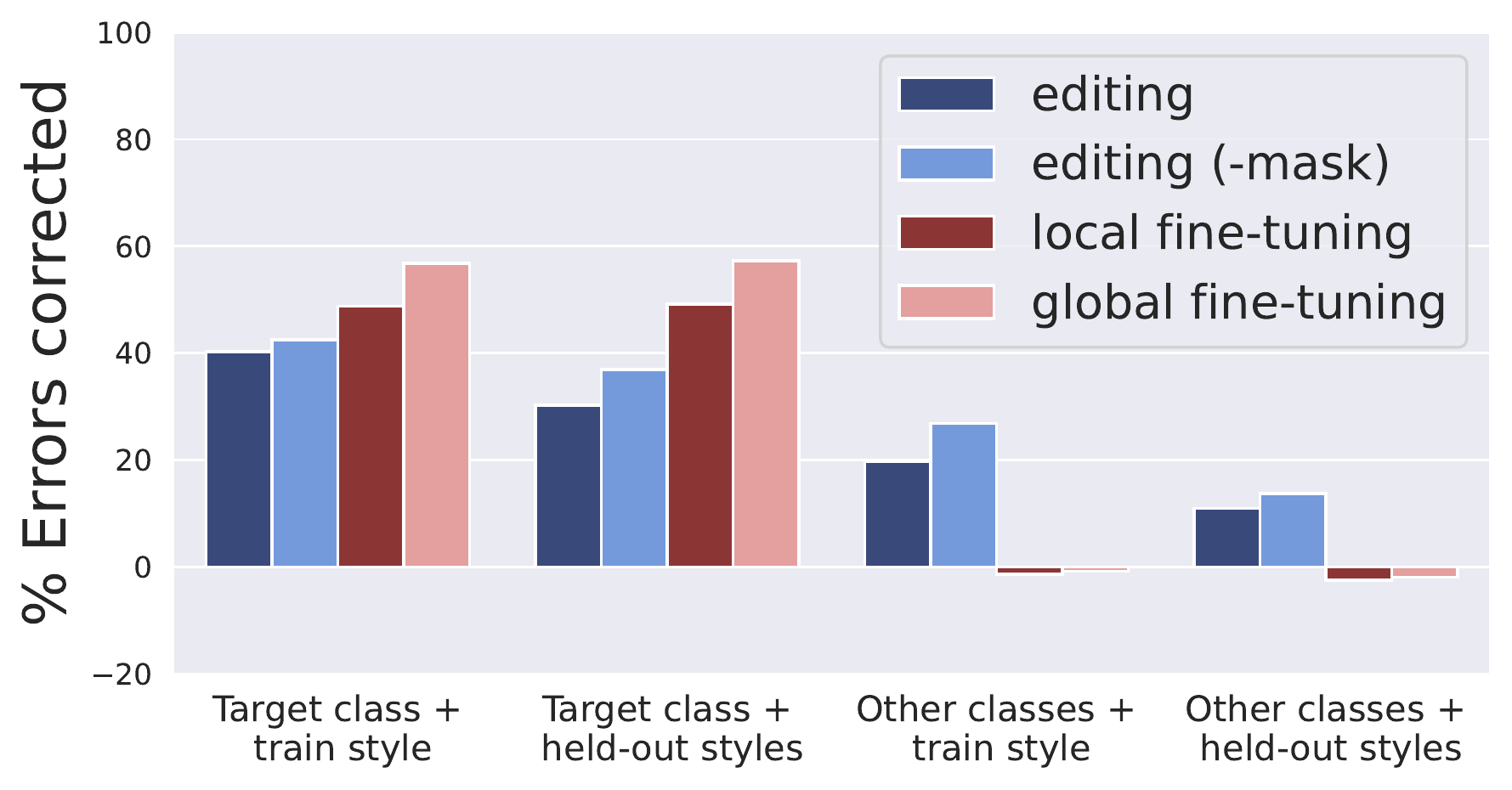}
		\caption{$3$ training exemplars}
	\end{subfigure}
\begin{subfigure}[b]{1\textwidth}
	\centering
	\includegraphics[width=0.45\columnwidth]{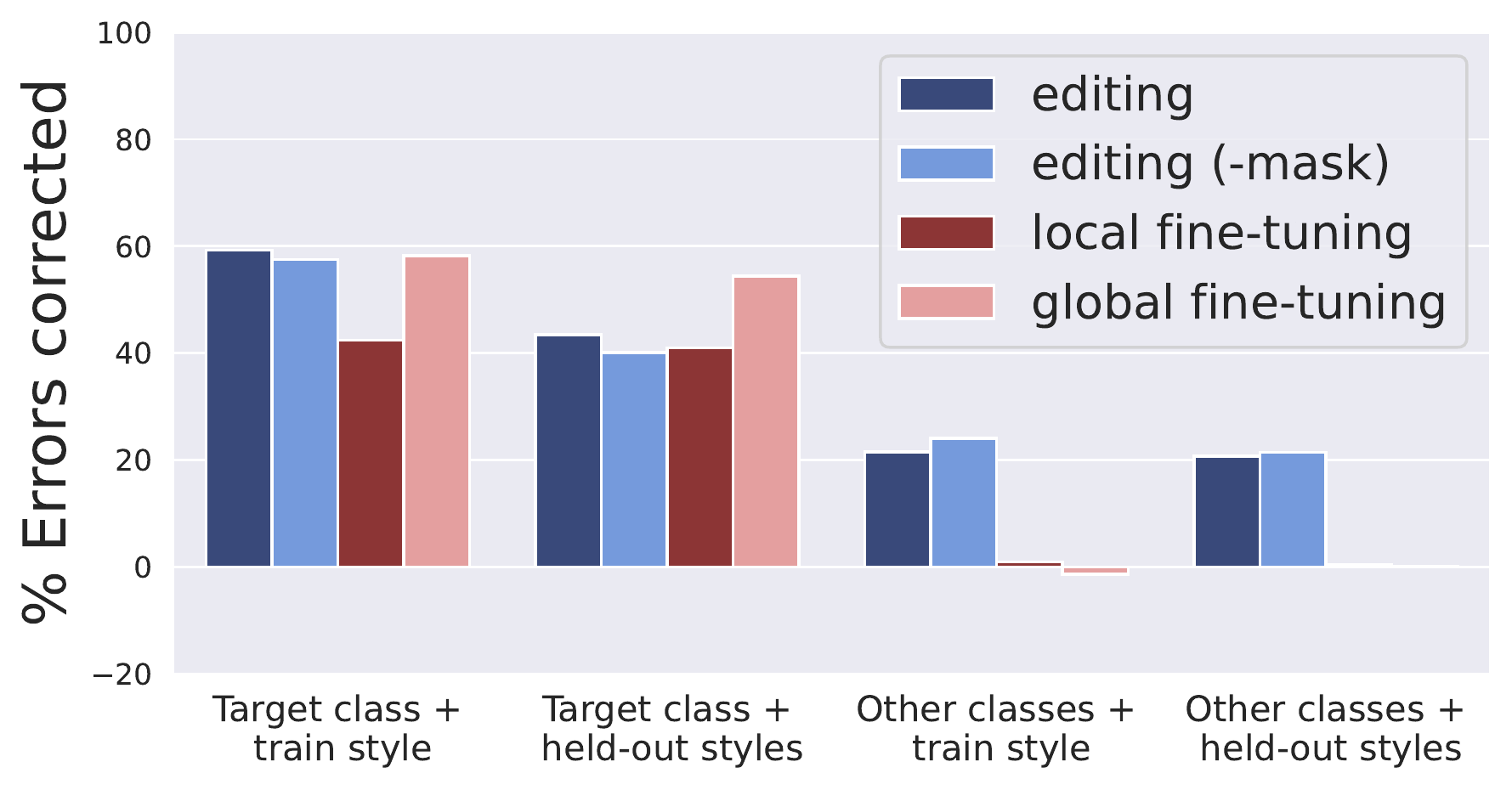}
	\hfil
	\includegraphics[width=0.45\columnwidth]{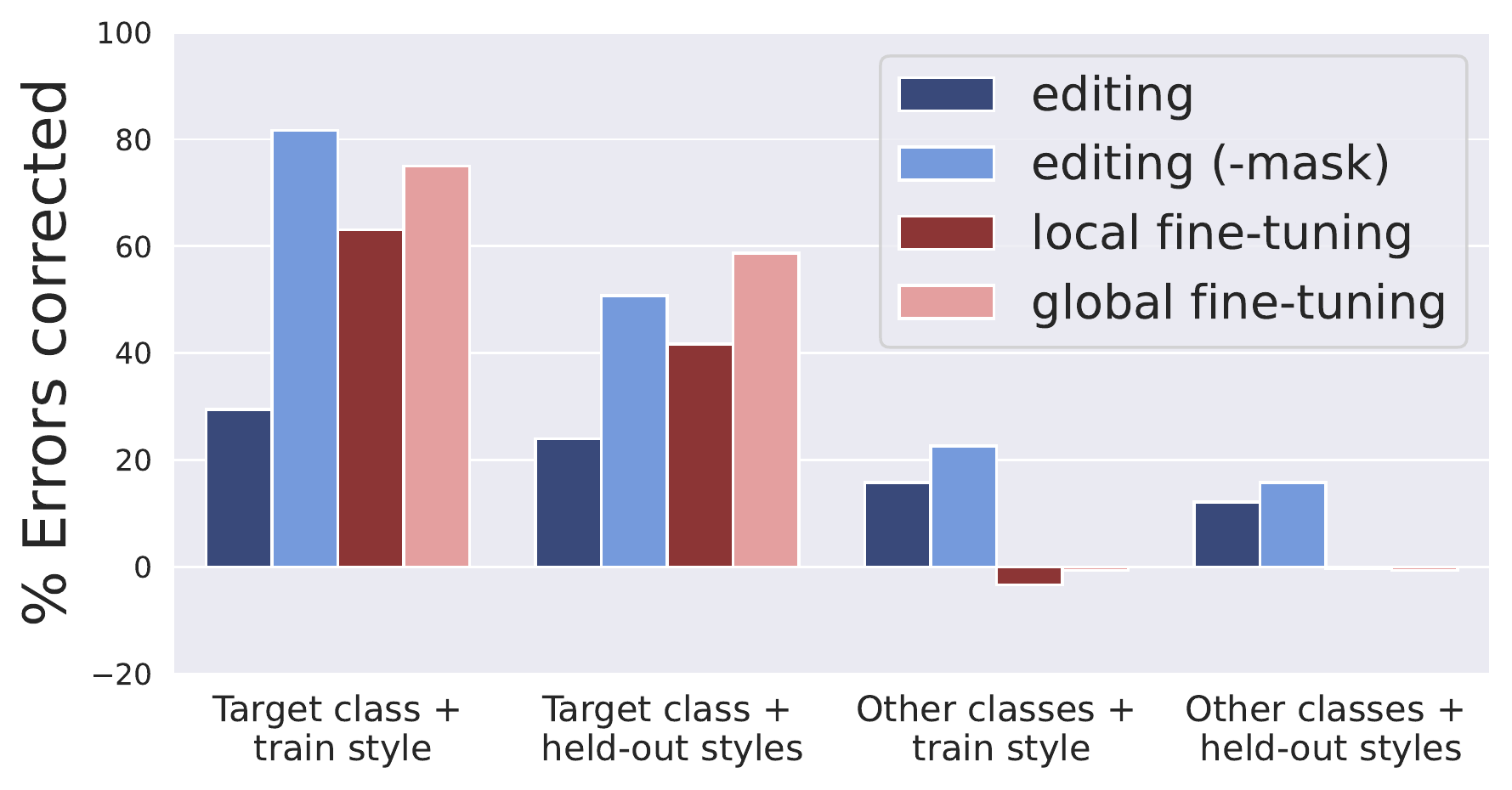}
	\caption{$10$ training exemplars}
\end{subfigure}
	\caption{Editing vs. fine-tuning: average number of 
	misclassifications corrected by the method when applied to an 
	ImageNet-trained VGG16 classifier. 
	Here, the average is computed over different concept-transformation 
	pairs---with concepts derived from instance 
	segmentation modules 
	trained on MS-COCO (\emph{left}) and 
	LVIS (\emph{right}); and transformations described in 
	Appendix~\ref{app:editing_setup}.
	For both editing and fine-tuning, the overall drop in model accuracy is less 
	than 
	$0.25\%$.}
	\label{fig:app_performance_imagenet_3}
\end{figure}

\begin{figure}
	\begin{subfigure}[b]{1\textwidth}
	\centering
	\includegraphics[width=0.45\columnwidth]{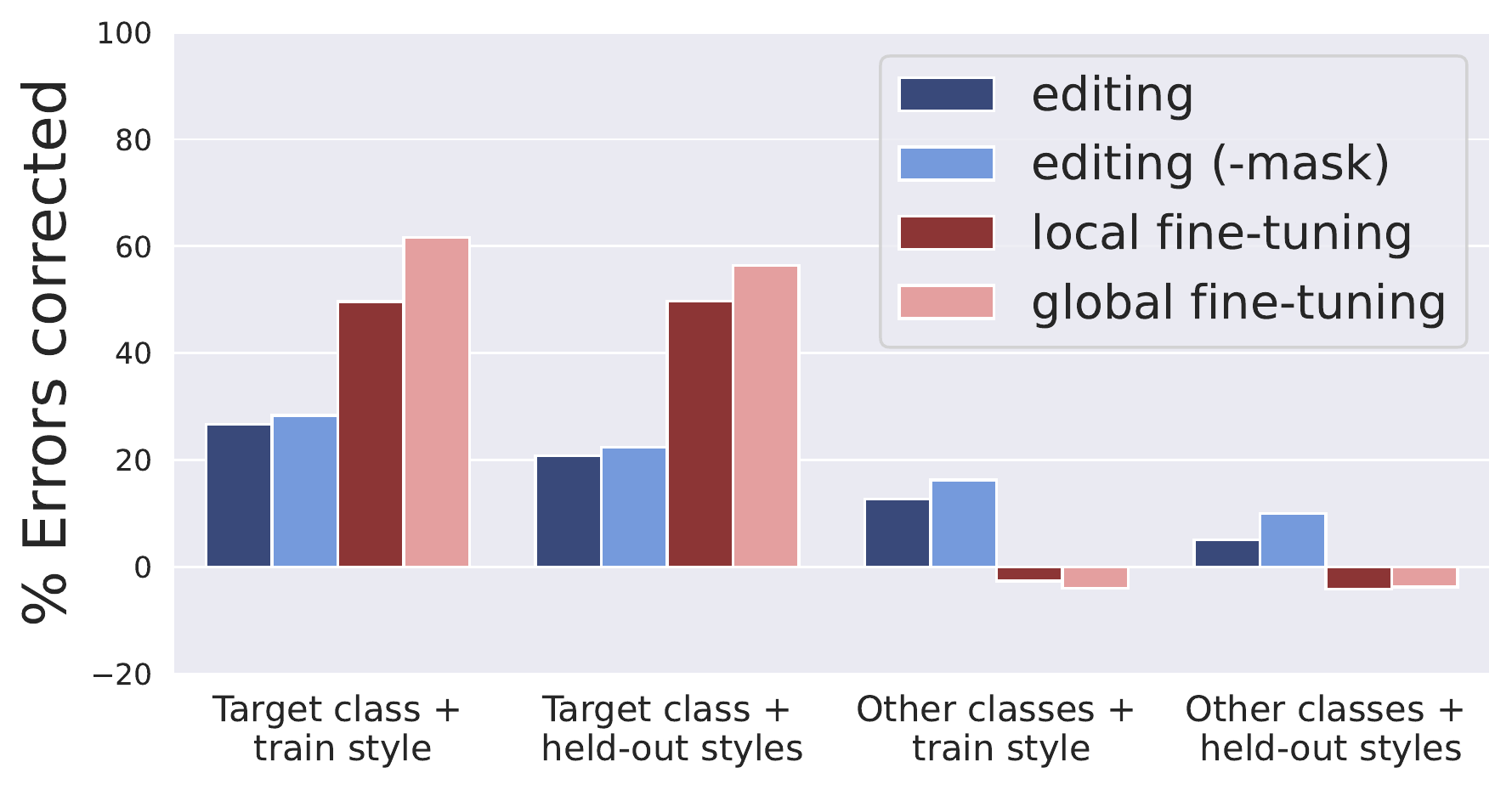}
	\hfil
	\includegraphics[width=0.45\columnwidth]{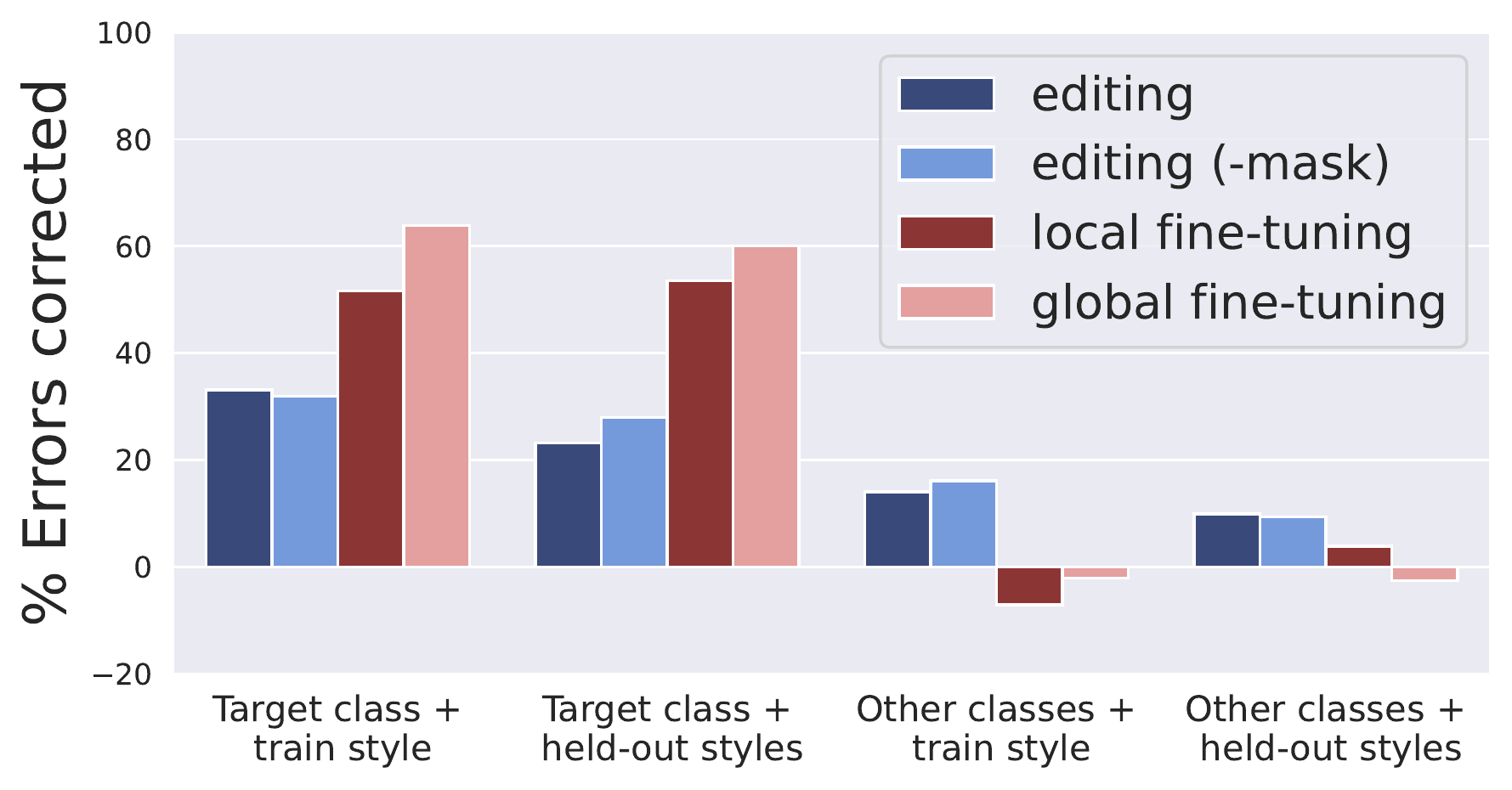}
	\caption{$3$ training exemplars}
\end{subfigure}
\begin{subfigure}[b]{1\textwidth}
	\centering
	\includegraphics[width=0.45\columnwidth]{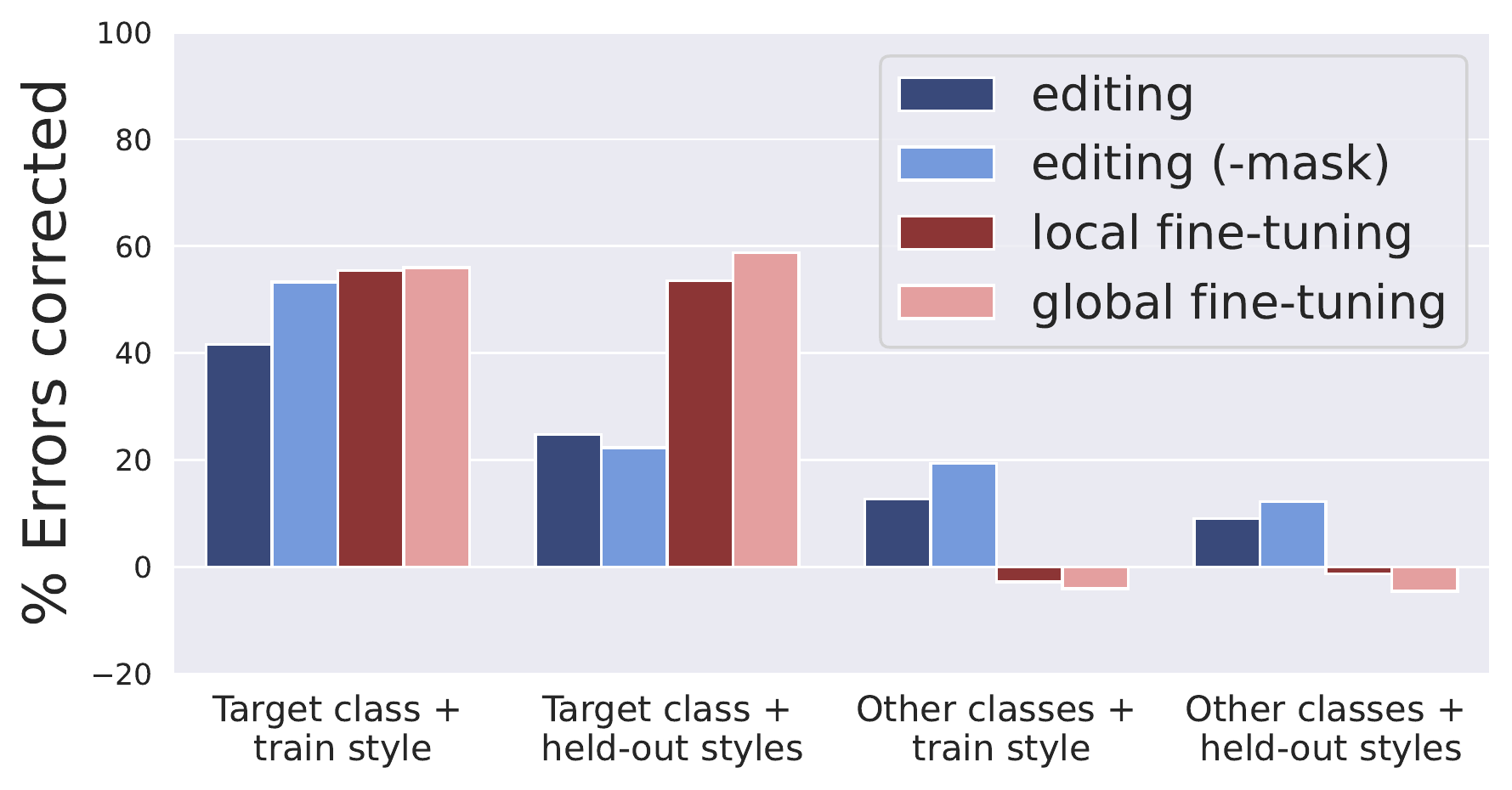}
	\hfil
	\includegraphics[width=0.45\columnwidth]{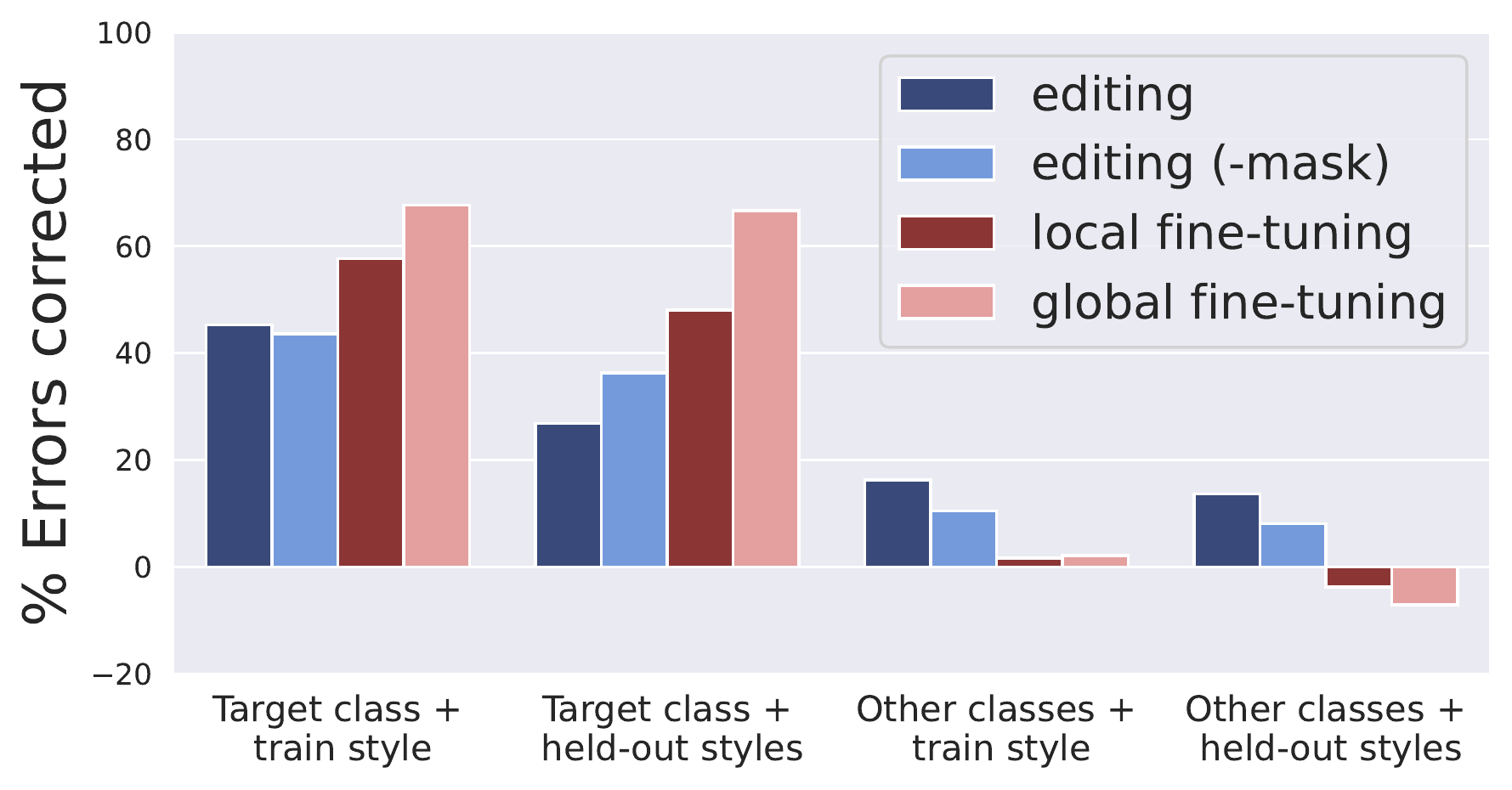}
	\caption{$10$ training exemplars}
\end{subfigure}
\caption{Repeating the analysis in Appendix 
Fig.~\ref{fig:app_performance_imagenet_3} on an 
	ImageNet-trained ResNet-50 classifier.}
	\label{fig:app_performance_imagenet_10}
\end{figure}

\begin{figure}
	\begin{subfigure}[b]{1\textwidth}
	\centering
	\includegraphics[width=0.45\columnwidth]{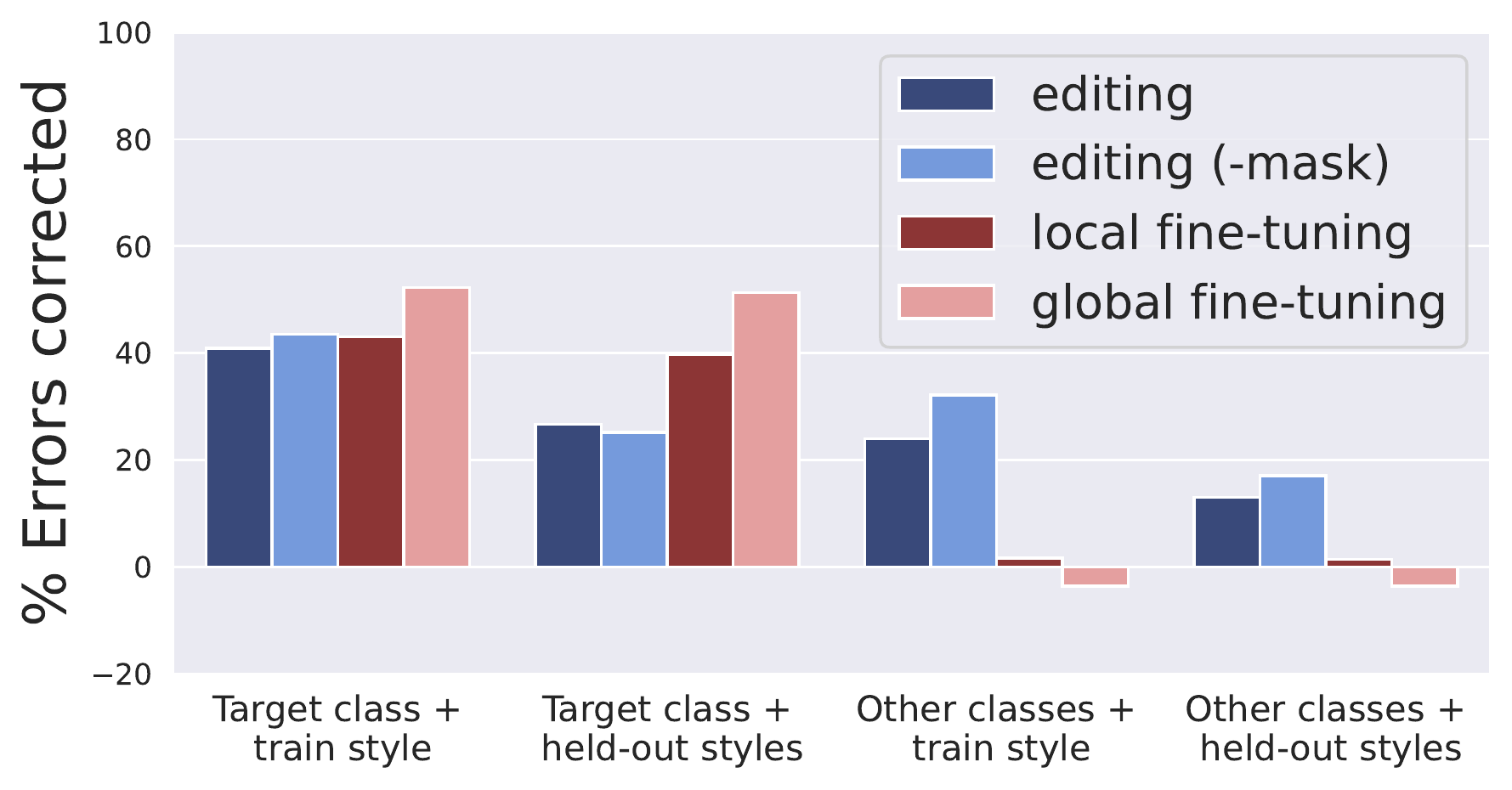}
	\hfil
	\includegraphics[width=0.45\columnwidth]{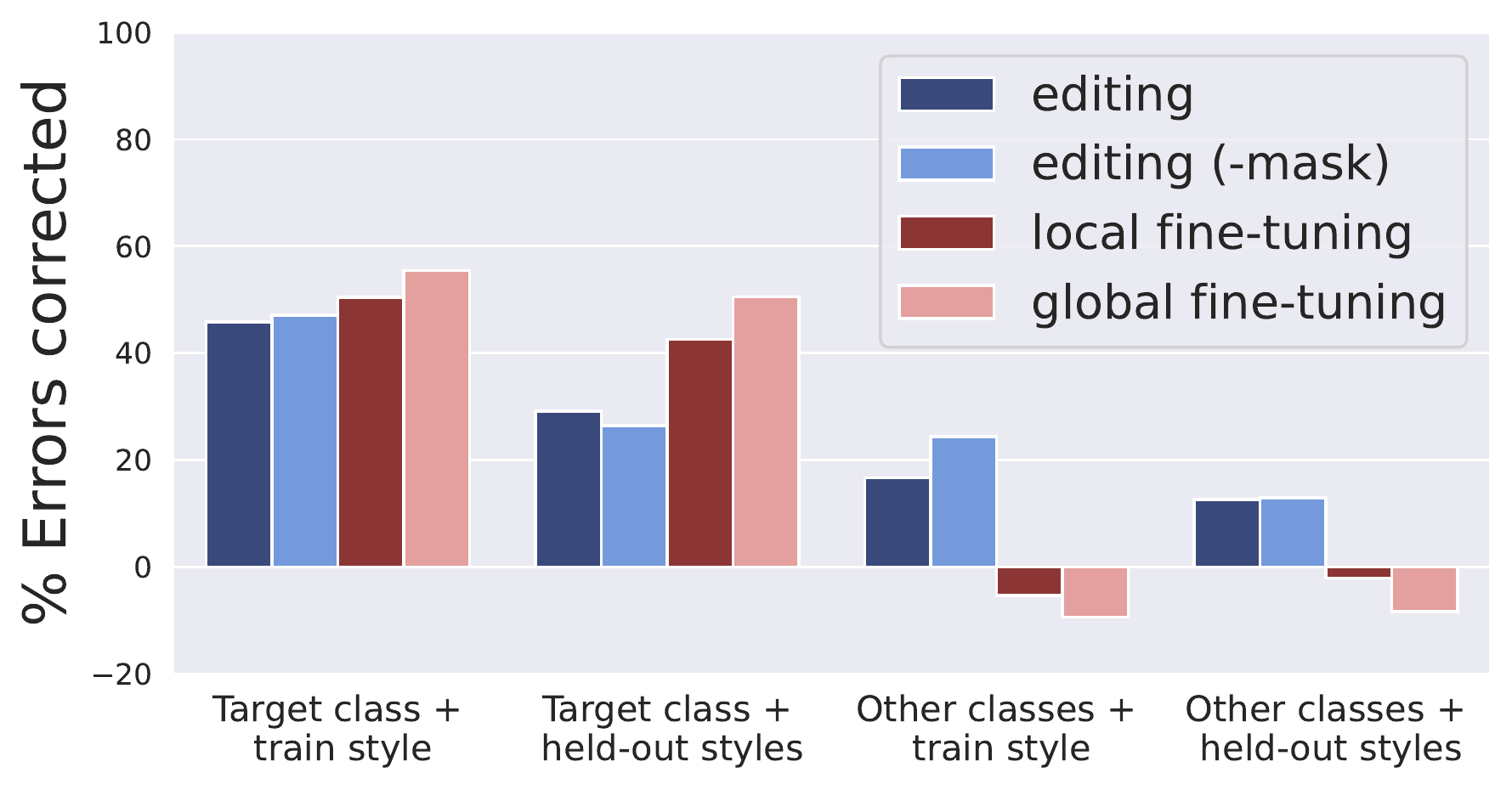}
	\caption{$3$ training exemplars}
\end{subfigure}
\begin{subfigure}[b]{1\textwidth}
	\centering
	\includegraphics[width=0.45\columnwidth]{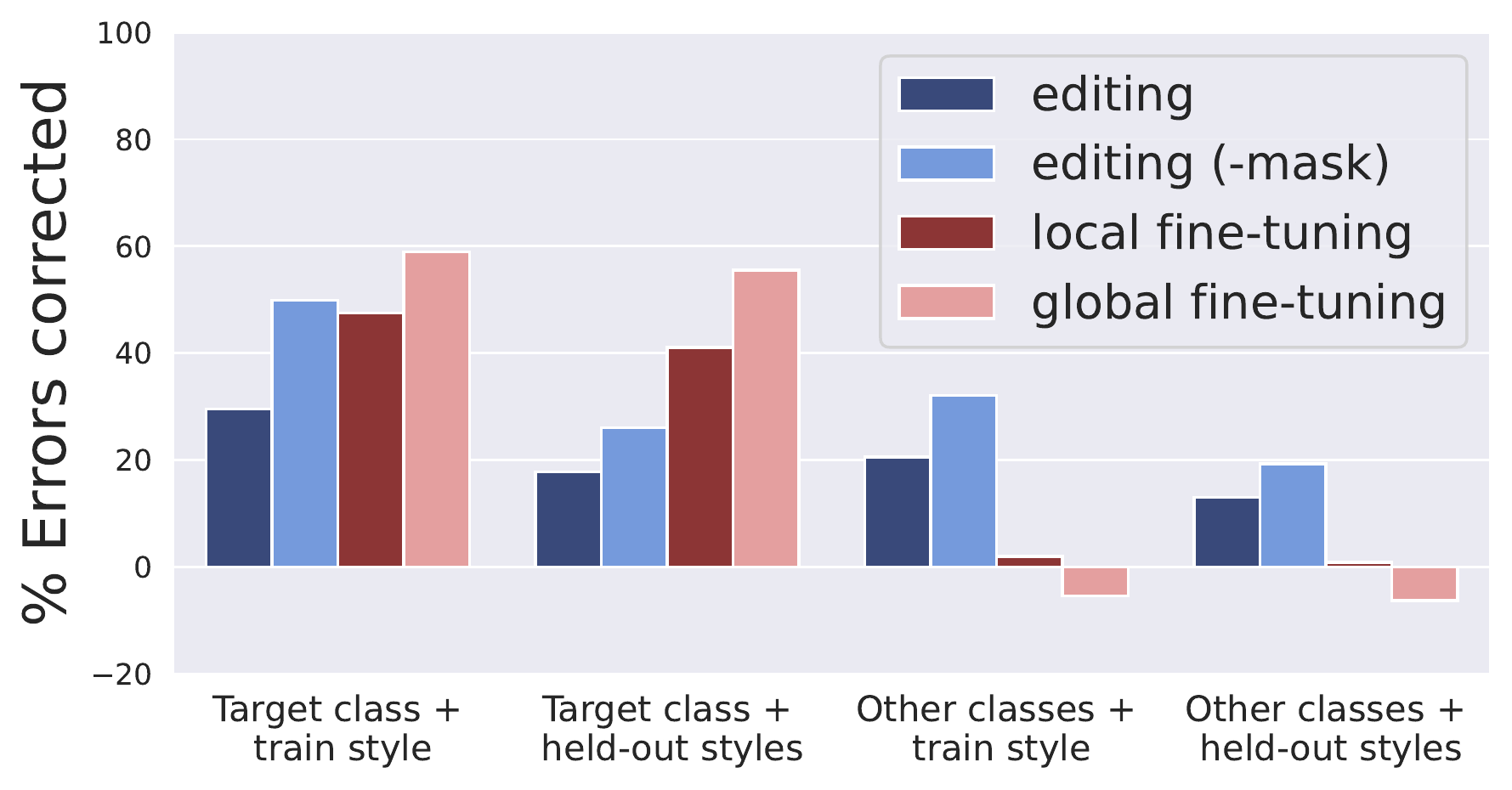}
	\hfil
	\includegraphics[width=0.45\columnwidth]{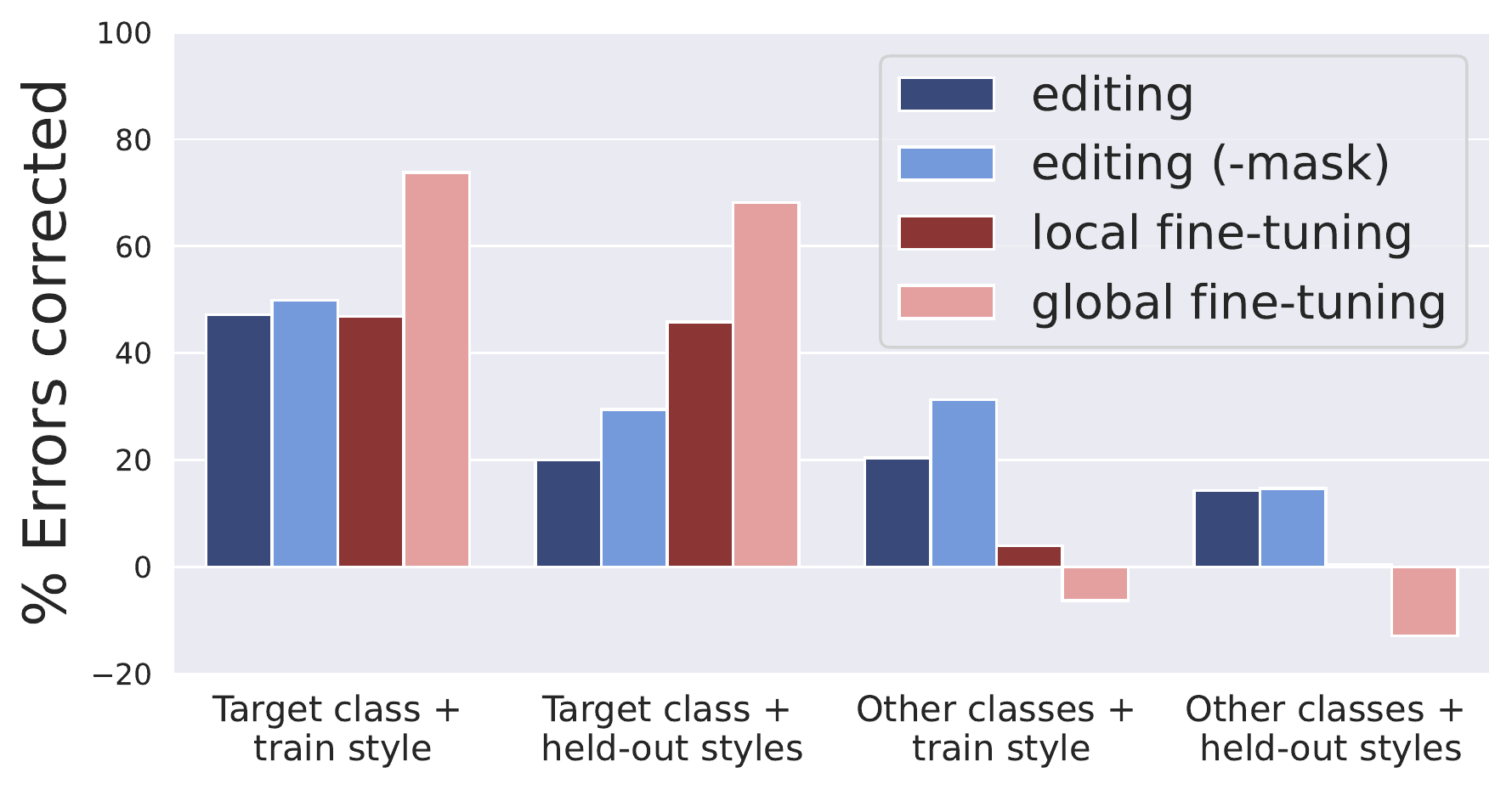}
	\caption{$10$ training exemplars}
\end{subfigure}
	\caption{Repeating the analysis in Appendix 
		Fig.~\ref{fig:app_performance_imagenet_3} on an  Places365-trained 
		VGG16 classifier. }
	\label{fig:app_performance_places_3}
\end{figure}

\begin{figure}
	\begin{subfigure}[b]{1\textwidth}
	\centering
	\includegraphics[width=0.45\columnwidth]{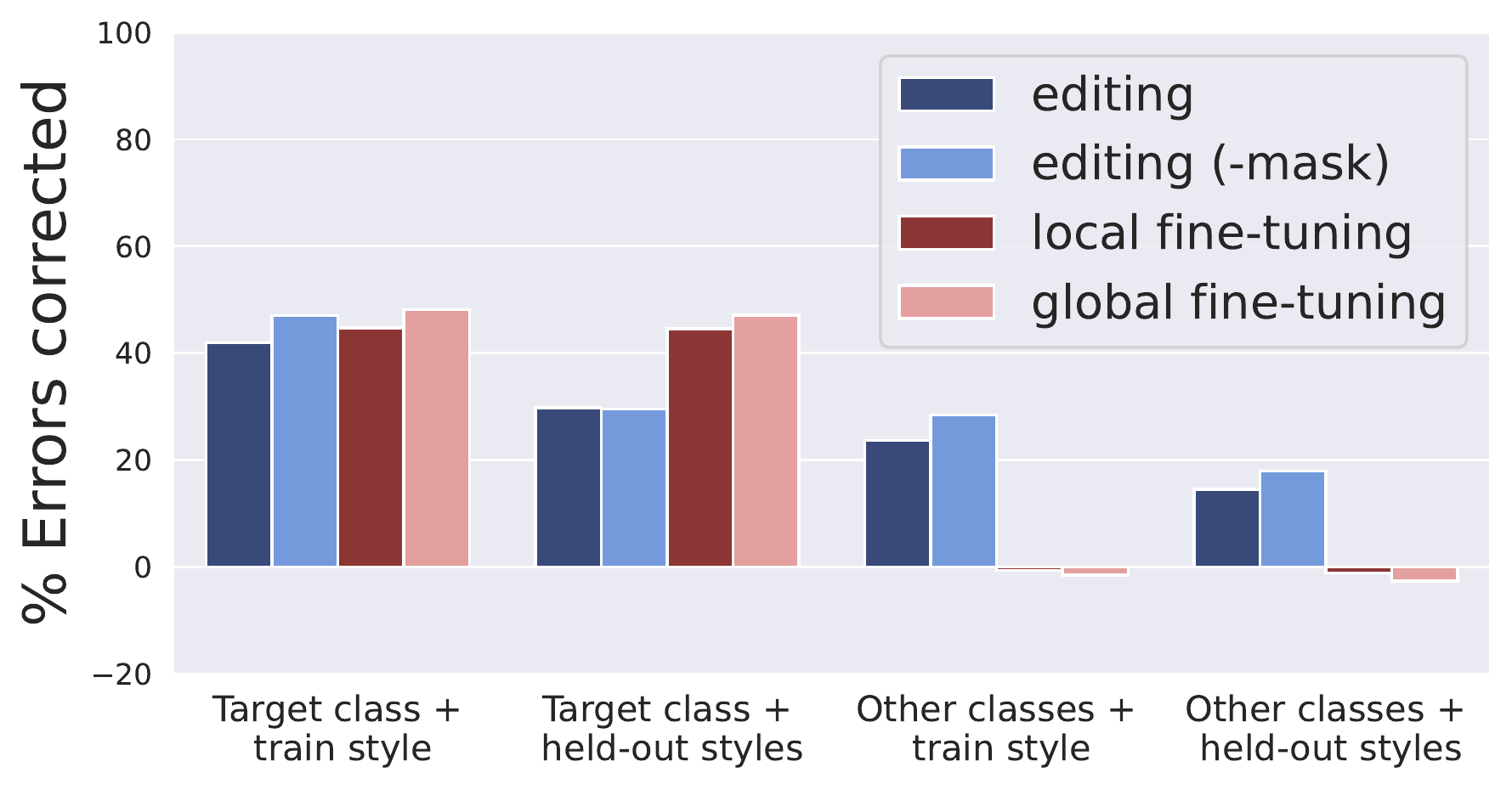}
	\hfil
	\includegraphics[width=0.45\columnwidth]{figures/editing/Places_LVIS_resnet18_0_3_25.pdf}
	\caption{$3$ training exemplars}
\end{subfigure}
\begin{subfigure}[b]{1\textwidth}
	\centering
	\includegraphics[width=0.45\columnwidth]{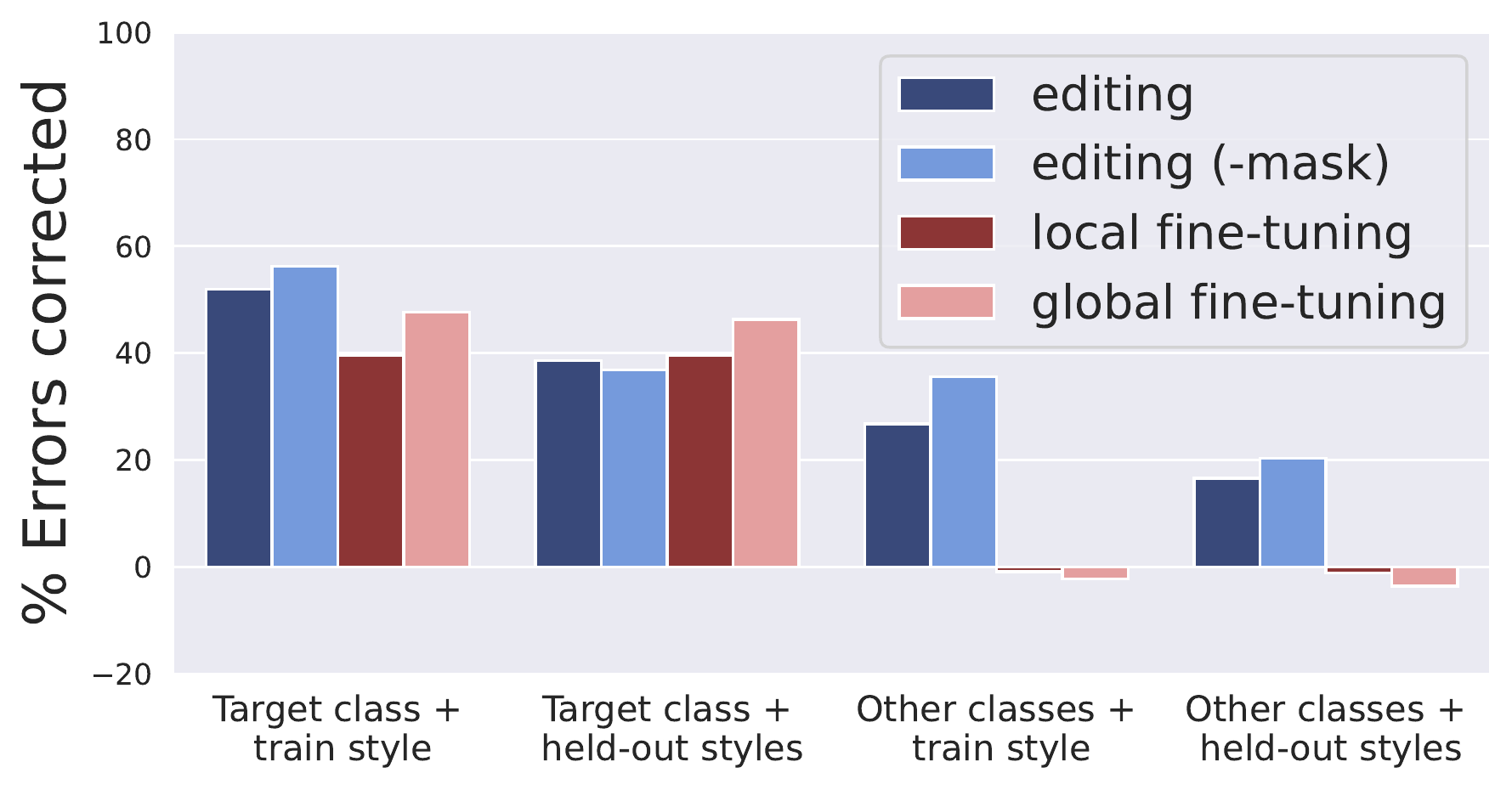}
	\hfil
	\includegraphics[width=0.45\columnwidth]{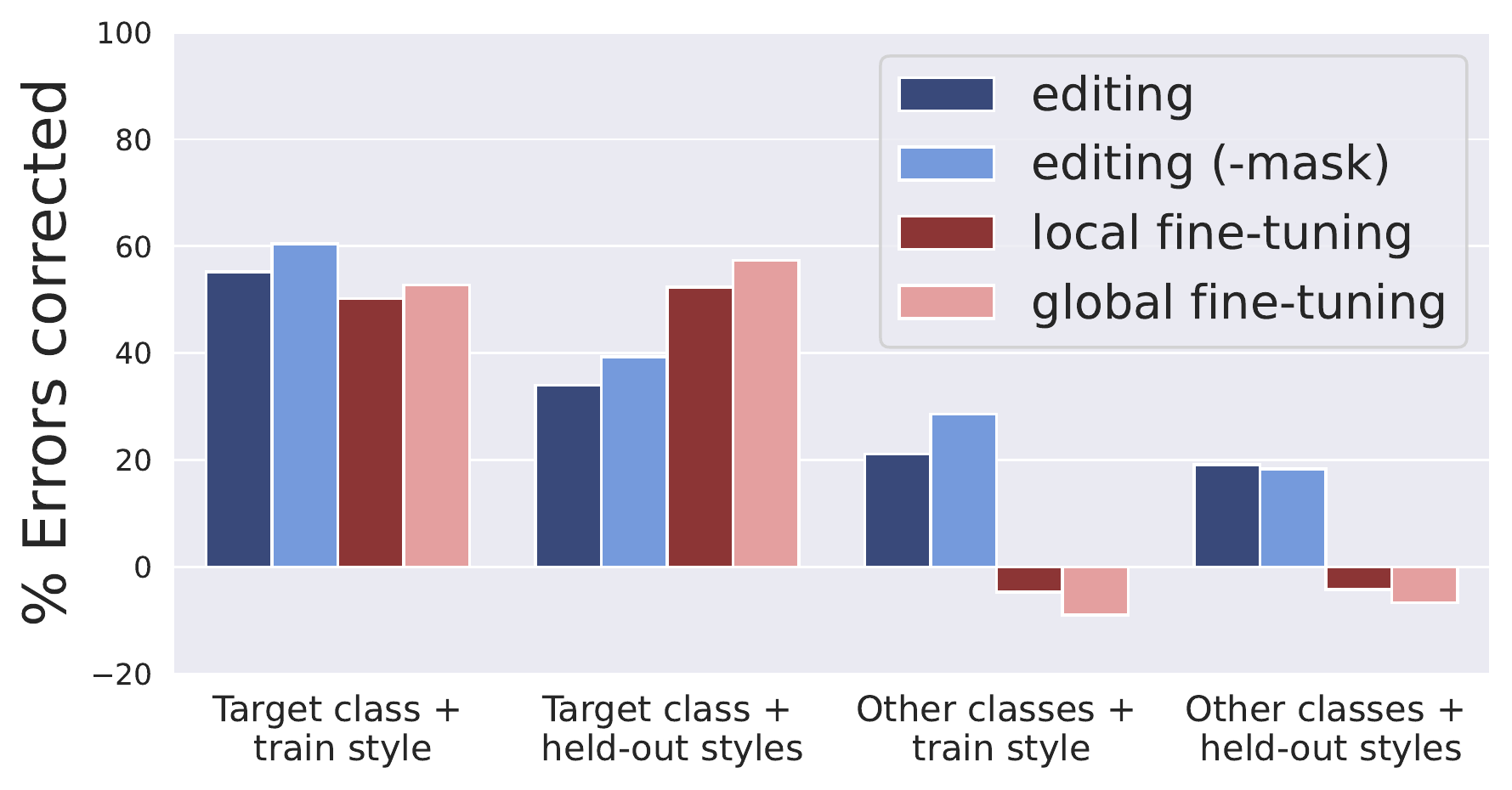}
	\caption{$10$ training exemplars}
\end{subfigure}
\caption{Repeating the analysis in Appendix 
	Fig.~\ref{fig:app_performance_imagenet_3} on an 
	Places365-trained ResNet-18 classifier. }
	\label{fig:app_performance_places_10}
\end{figure}

In 
Figures~\ref{fig:app_performance_imagenet_3}-~\ref{fig:app_performance_places_10},
 we compare the generalization performance of editing and fine-tuning (and 
their variants)---for different datasets (ImageNet and Places), 
architectures 
(VGG16 and ResNets) and number of exemplars (3 and 10).
In performing these evaluations, we only consider hyperparameters (for each 
concept-style pair) that do not drop the overall (test set) accuracy of the 
model by over 0.25\%.
The complete accuracy-performance trade-offs of editing and fine-tuning 
(and their variants) are illustrated in Appendix 
Figures~\ref{fig:app_tradeoff_editing_imagenet_vgg}-\ref{fig:app_tradeoff_editing_places_resnet}.

We observe that both methods successfully generalize to 
held-out samples from the target class (used to perform the 
modification)---even when the transformation is performed using held-out 
styles.
However, while the performance improvements of editing also extend to 
other classes containing the same concept, this does not seem to be the 
case for fine-tuning.
These trends hold even when we use more 
exemplars to perform the modification.
In Appendix Figure~\ref{fig:app_errors_corrected}, we illustrate sample error 
corrections (and failures to do so) due to editing and fine-tuning. 


\begin{figure}[!h]
	\centering
		\begin{subfigure}[b]{1\textwidth}
		\centering
		\includegraphics[width=0.7\columnwidth]{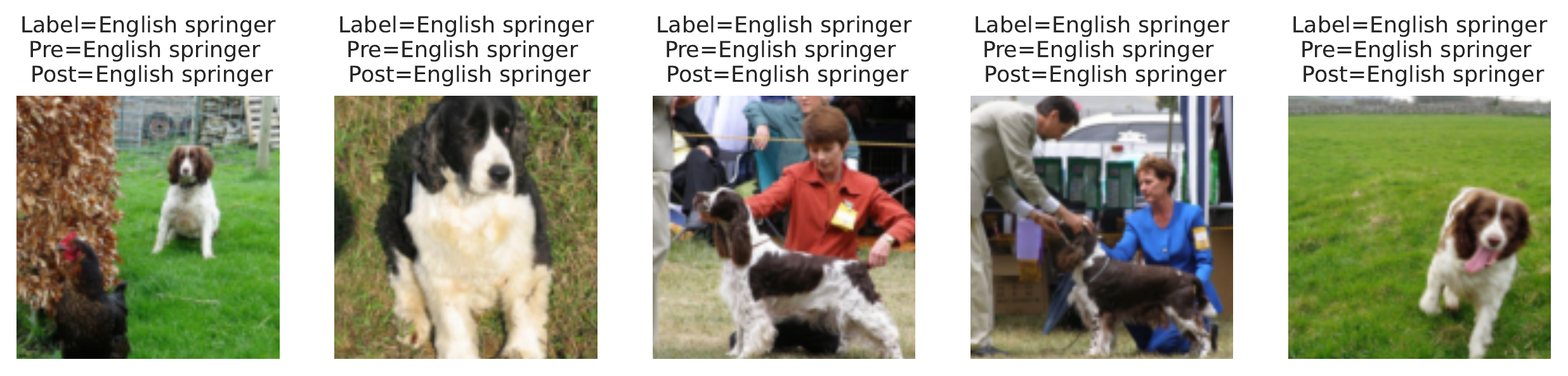}
		\includegraphics[width=0.7\columnwidth]{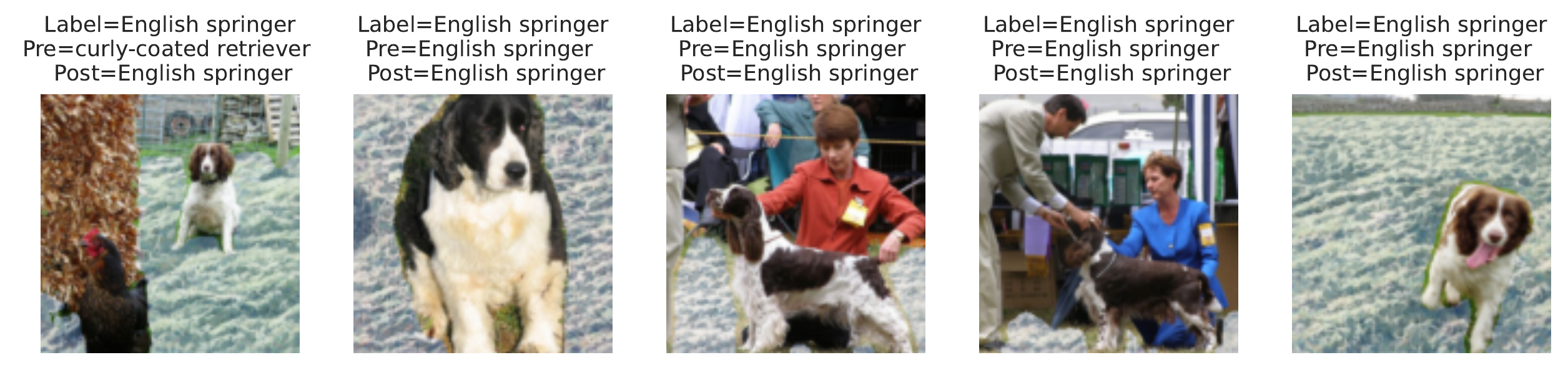}
		\caption{Train exemplars: (\emph{top}) Original and (\emph{bottom}) 
		transformed images.}
	\end{subfigure} \\
    \vspace{-.5cm}
\caption*{\\ \large{Method: Fine-tuning}}
		\begin{subfigure}[b]{0.28\textwidth}
	\centering
	\includegraphics[width=.9\columnwidth]{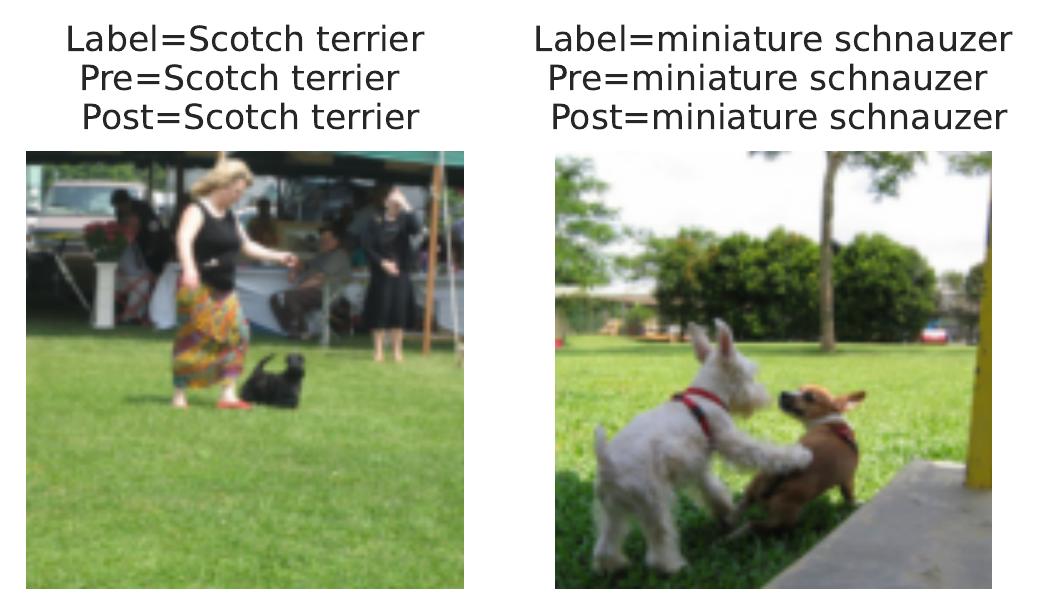}
	\includegraphics[width=.9\columnwidth]{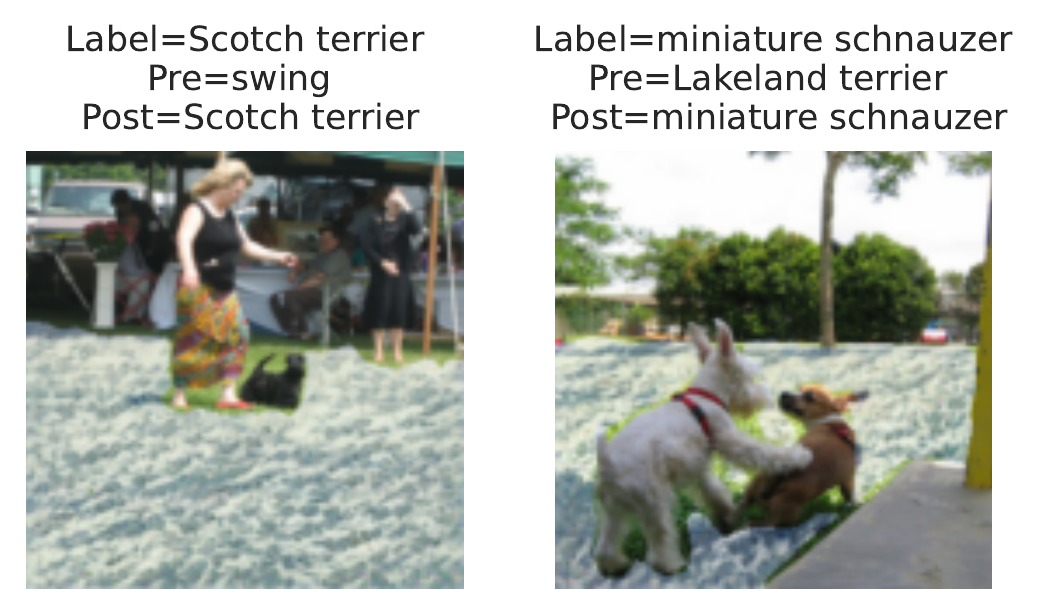}
	\caption{Corrections (3.12\%)}
\end{subfigure} \hfil
		\begin{subfigure}[b]{0.28\textwidth}
	\centering
	\includegraphics[width=.9\columnwidth]{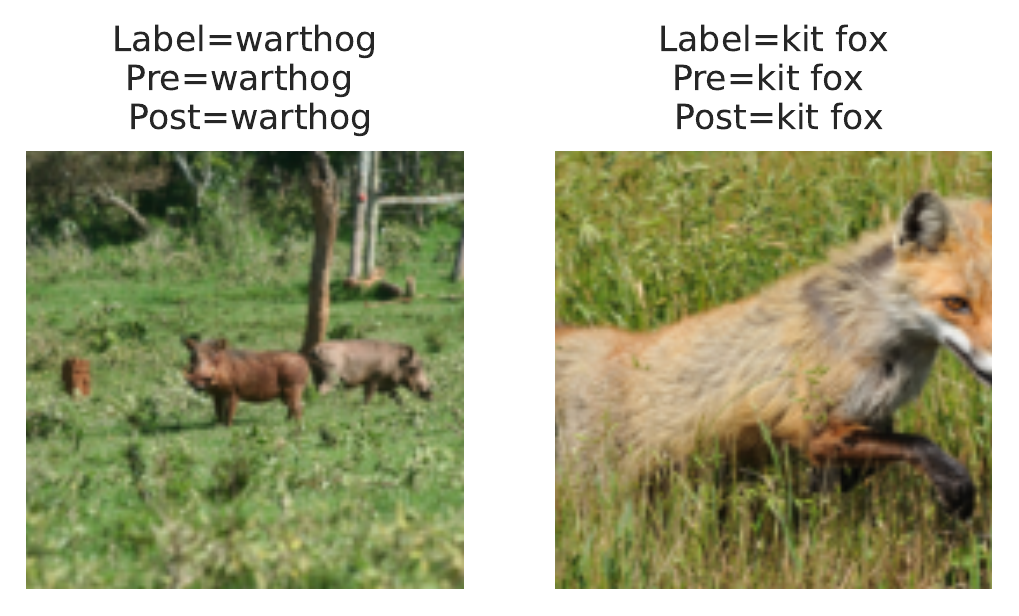}
	\includegraphics[width=.9\columnwidth]{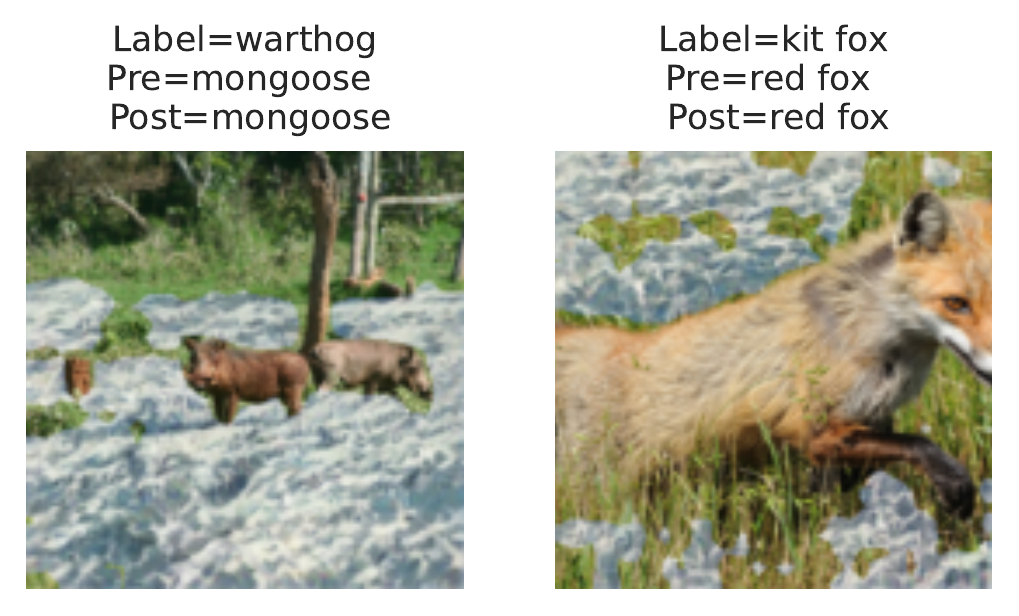}
	\caption{No change (85.63\%)}
\end{subfigure} \hfil
		\begin{subfigure}[b]{0.28\textwidth}
	\centering
	\includegraphics[width=.9\columnwidth]{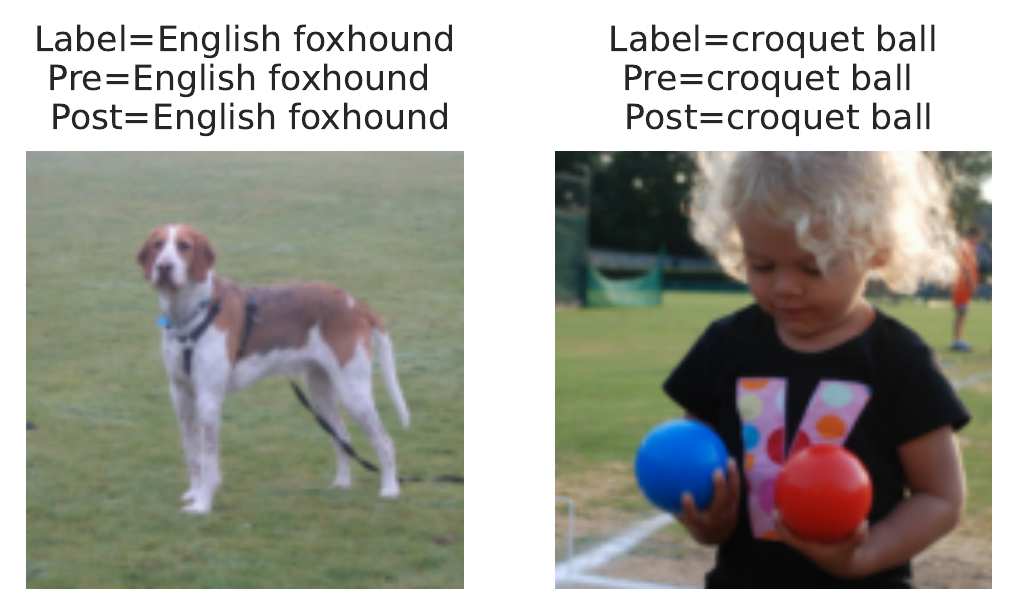}
	\includegraphics[width=.9\columnwidth]{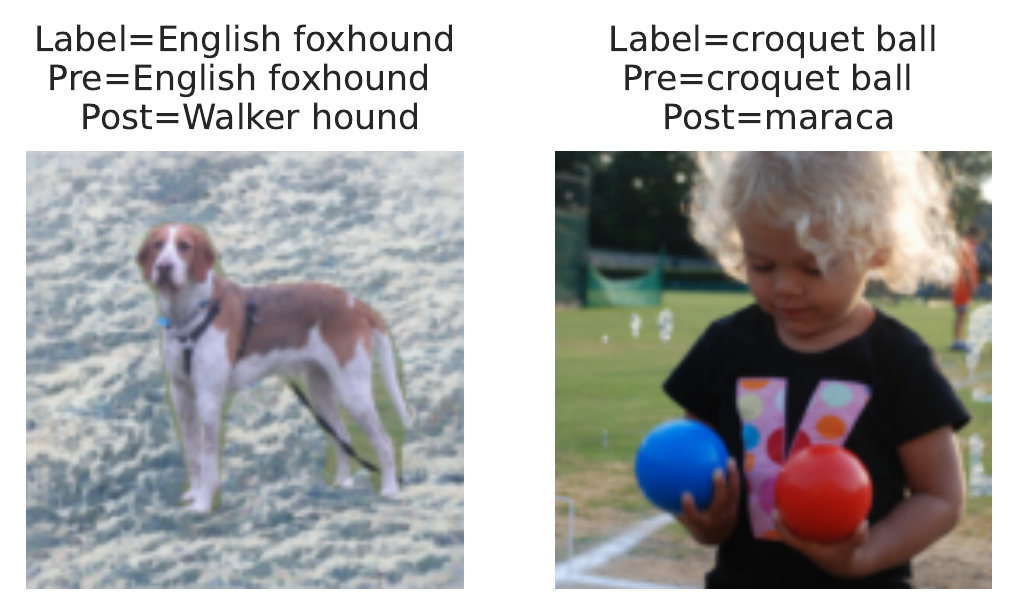}
	\caption{New errors (11.25\%)}
\end{subfigure}
    \vspace{-.5cm}
\caption*{\\ \large{Method: Editing}}
		\begin{subfigure}[b]{0.28\textwidth}
	\centering
	\includegraphics[width=.9\columnwidth]{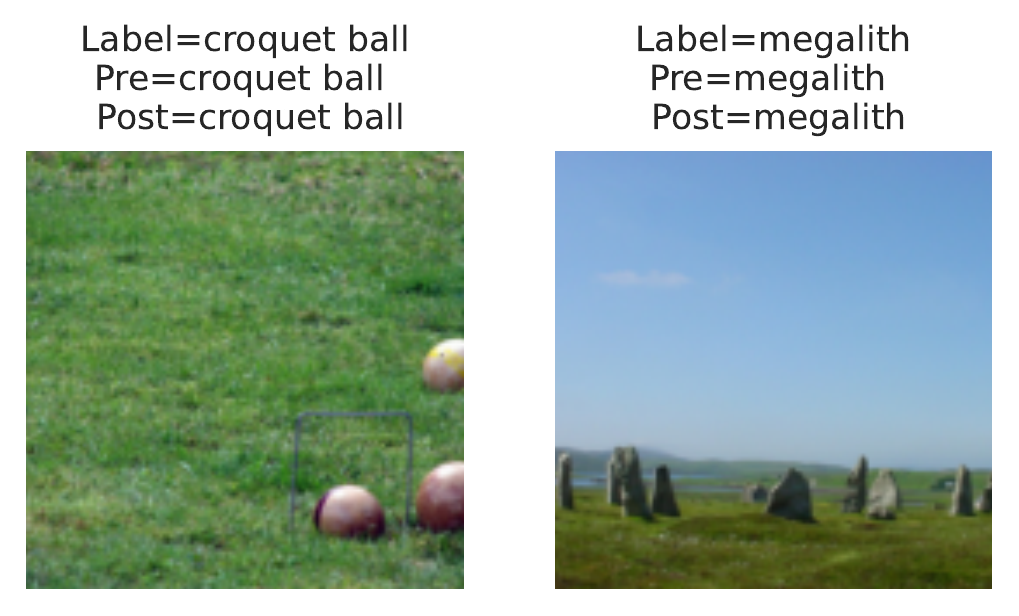}
	\includegraphics[width=.9\columnwidth]{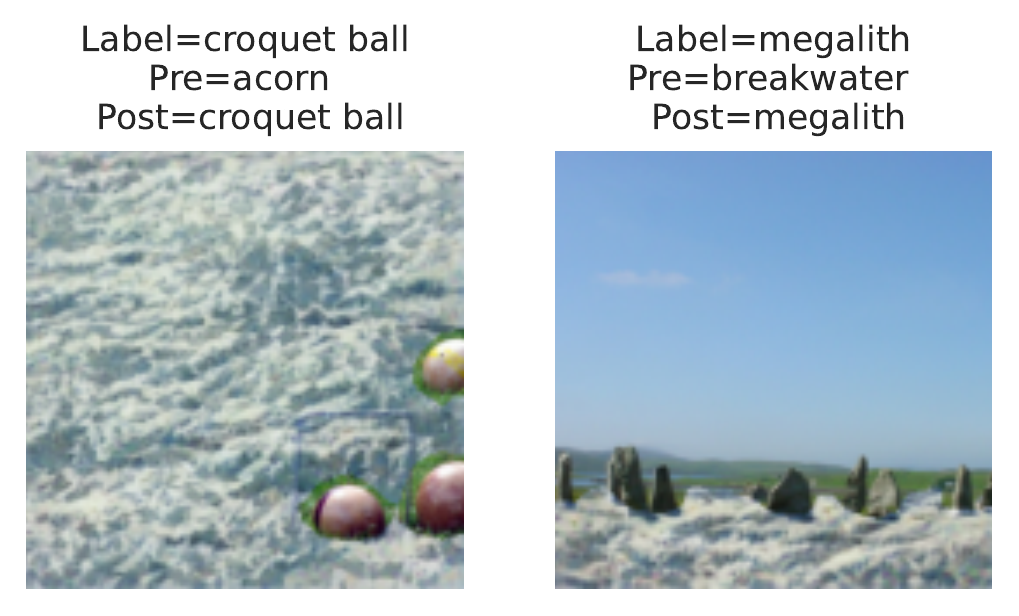}
	\caption{Corrections (32.89\%)}
\end{subfigure} \hfil
\begin{subfigure}[b]{0.28\textwidth}
	\centering
	\includegraphics[width=.9\columnwidth]{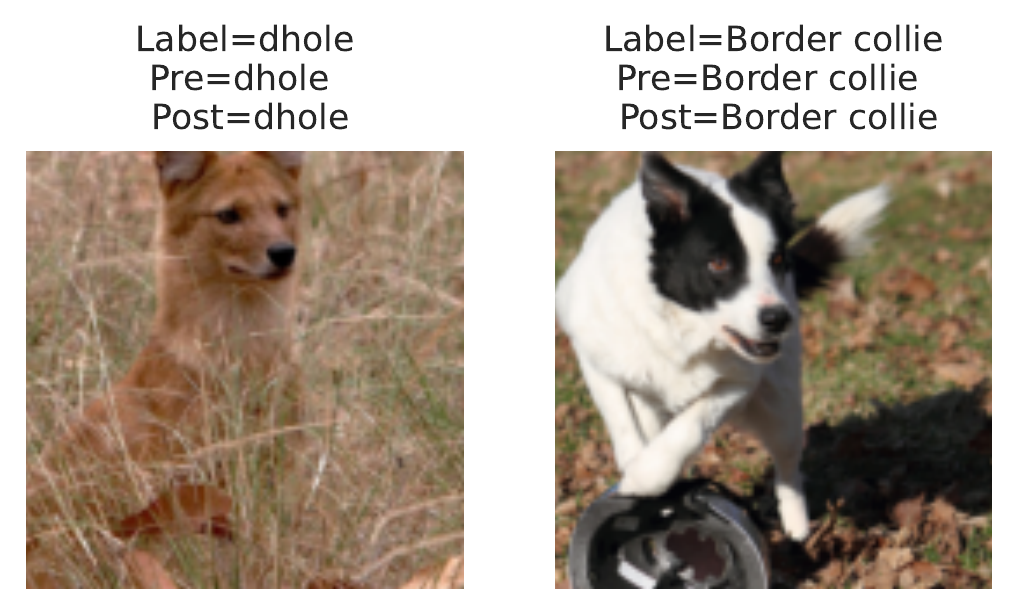}
	\includegraphics[width=.9\columnwidth]{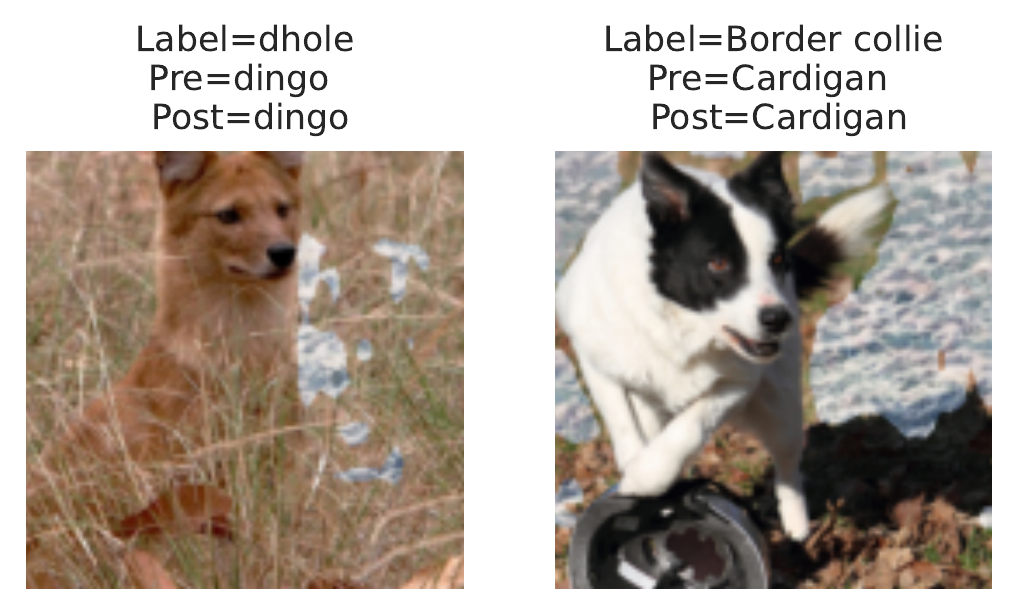}
	\caption{No change  (59.87\%)}
\end{subfigure} \hfil
\begin{subfigure}[b]{0.28\textwidth}
	\centering
	\includegraphics[width=.9\columnwidth]{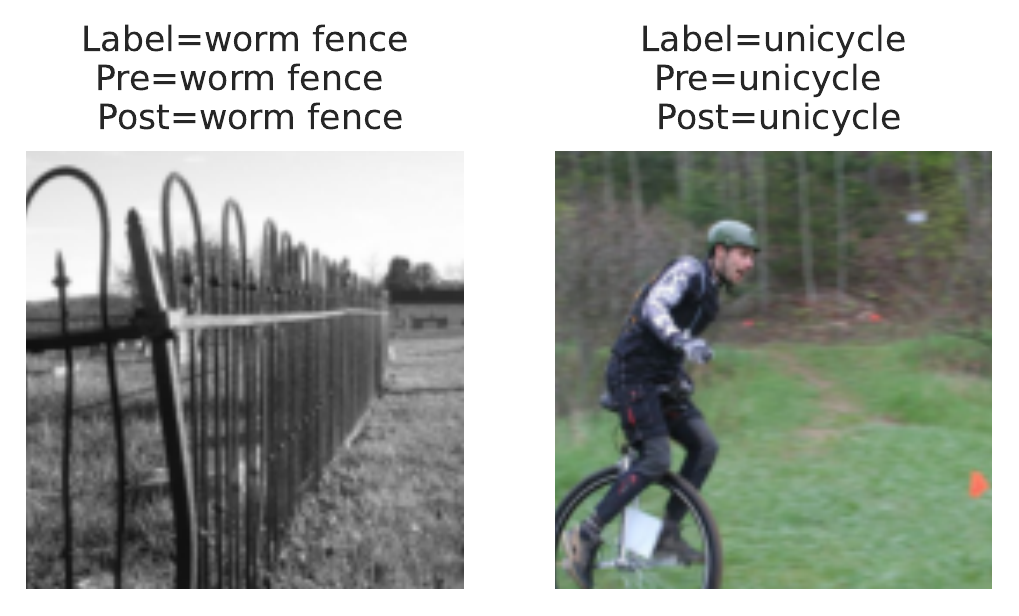}
	\includegraphics[width=.9\columnwidth]{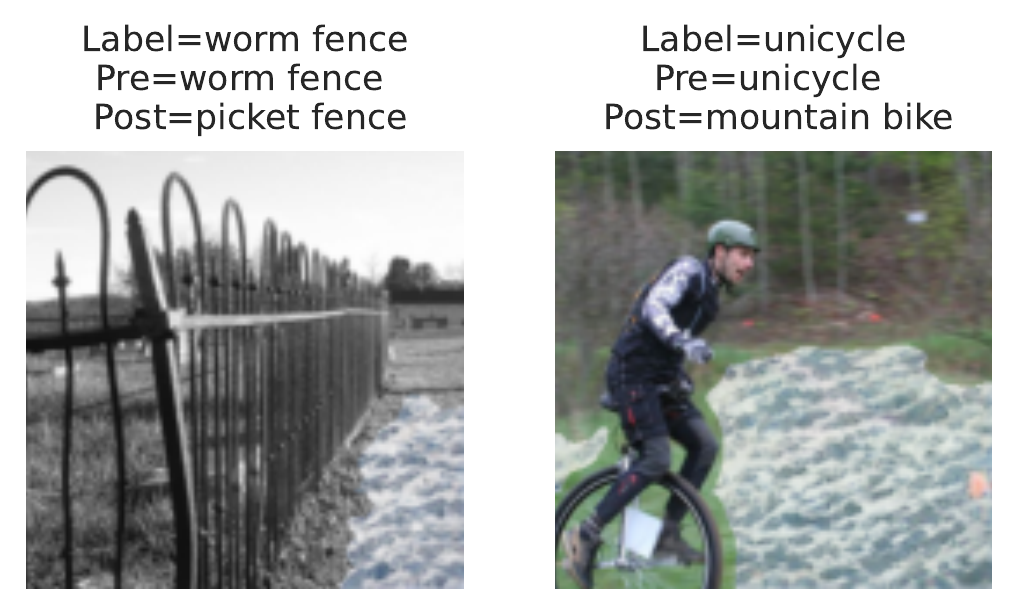}
	\caption{New errors (7.24\%)}
\end{subfigure}
	\caption{Examples of errors (not) corrected by the rewrite:  Here, the goal 
	is to improve the accuracy of a VGG16 model 
	when ``grass'' in ImageNet images is replaced with ``snow''. The true 
	label and the pre/post-edit predictions
	for each image are in the title. (a) Train exemplars 
	for editing and fine-tuning. Test set examples where fine-tuning and 
	editing
	correct the model error on the transformed example (b/e), do not cause 
	any 
	change (c/f) and induce an new error (d/g). The number in parenthesis 
	indicates the fraction of the test set that falls into each of these subsets.}
	\label{fig:app_errors_corrected}
\end{figure}

\begin{figure}[!h]
	\centering
	\begin{subfigure}[b]{0.9\textwidth}
		\centering
		\includegraphics[trim=0 0 0 30, clip, 
		width=1\columnwidth]{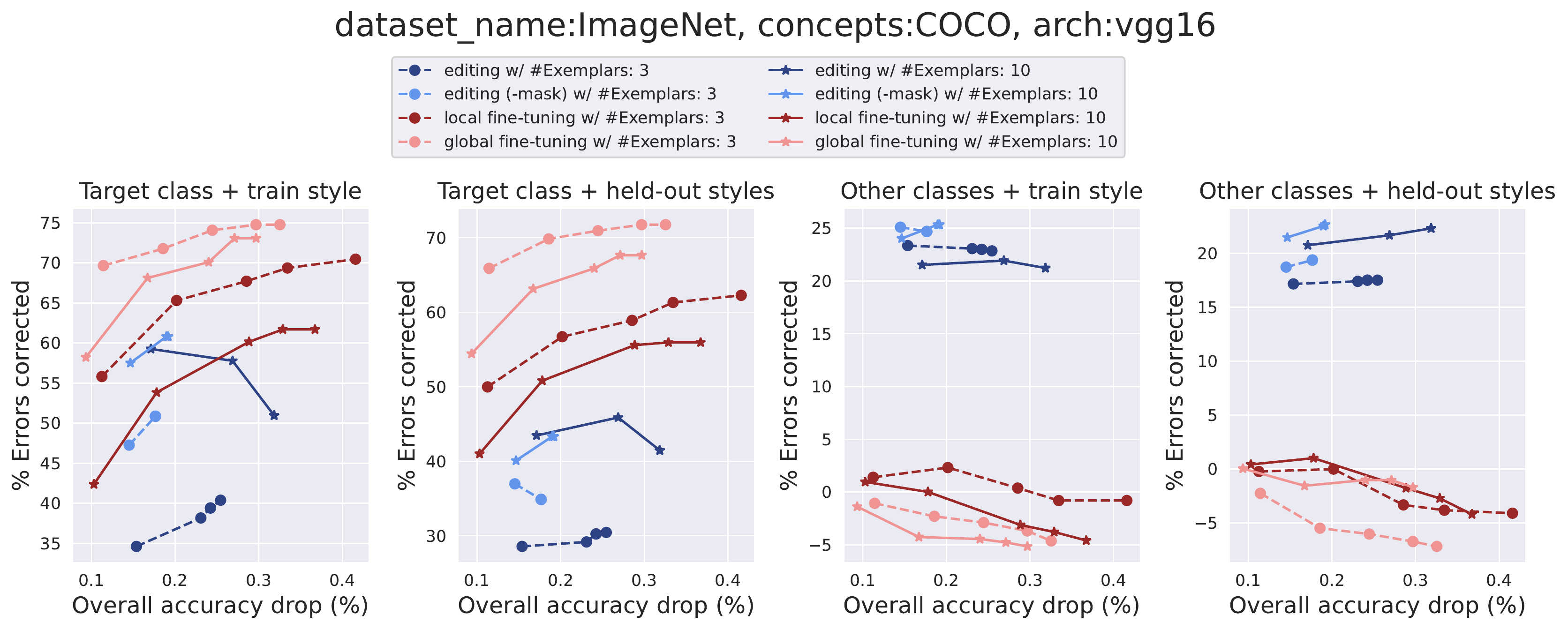}
		\caption{Concepts derived from an instance segmentation model trained 
		on MS-COCO.}
	\end{subfigure} 
	\begin{subfigure}[b]{0.9\textwidth}
		\vspace{0.25cm}
		\centering
		\includegraphics[trim=0 0 0 110, clip, 
		width=1\columnwidth]{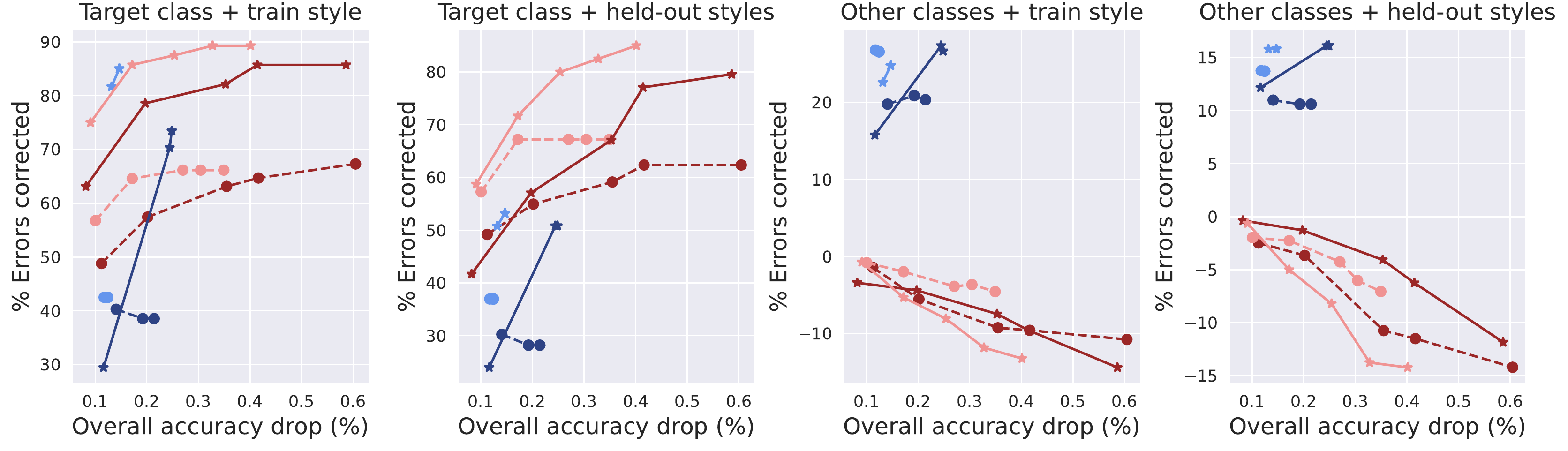}
		\caption{Concepts derived from an instance segmentation model trained 
			on LVIS.}
	\end{subfigure}
	\caption{Performance vs. drop in 
	overall test set accuracy: Here, we visualize average number of 
		misclassifications corrected by editing and fine-tuning when applied to 
		an 
		ImageNet-trained VGG16 classifier---where the average is computed 
		over different concept-transformation 
		pairs. }
	\label{fig:app_tradeoff_editing_imagenet_vgg}
\end{figure}

\begin{figure}[!h]
	\centering
	\begin{subfigure}[b]{0.9\textwidth}
		\centering
		\includegraphics[trim=0 0 0 30, clip, 
		width=1\columnwidth]{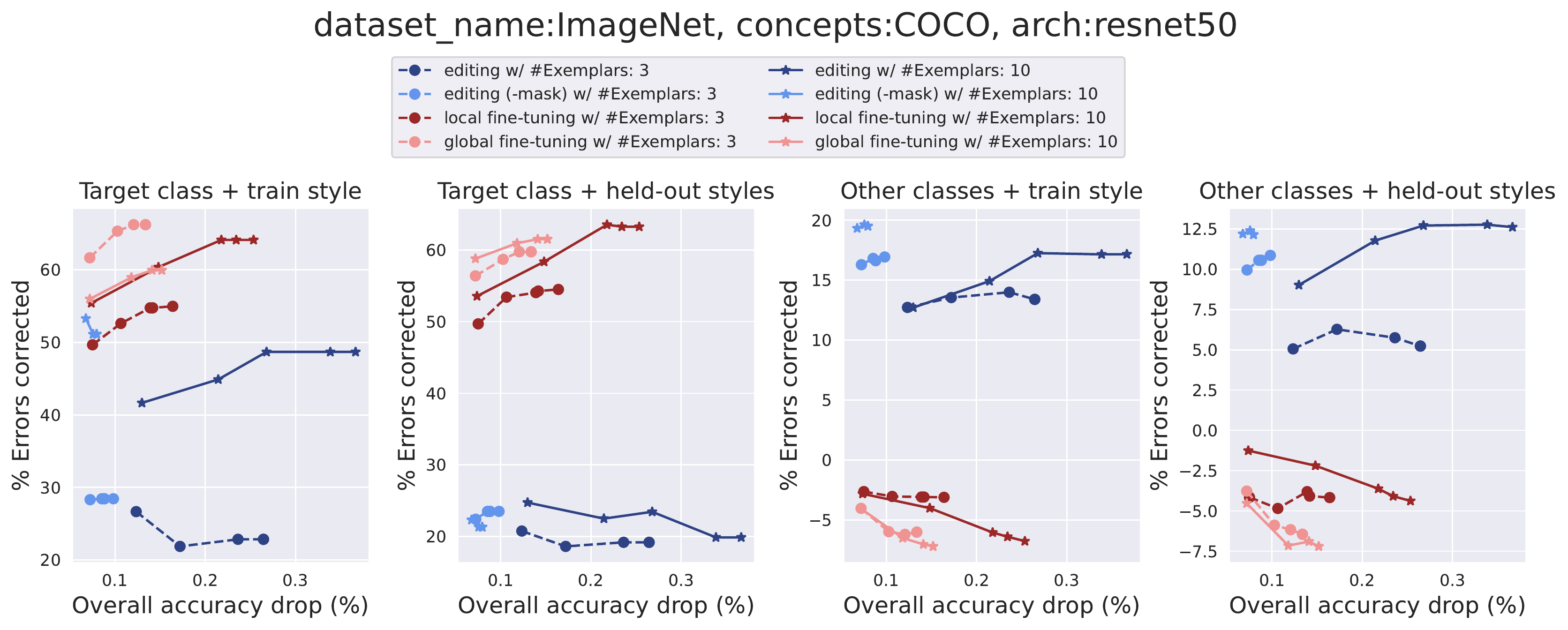}
		\caption{Concepts derived from an instance segmentation model trained 
	on MS-COCO.}
	\end{subfigure}
	\begin{subfigure}[b]{0.9\textwidth}
		\vspace{0.25cm}
		\centering
		\includegraphics[trim=0 0 0 110, clip, 
		width=1\columnwidth]{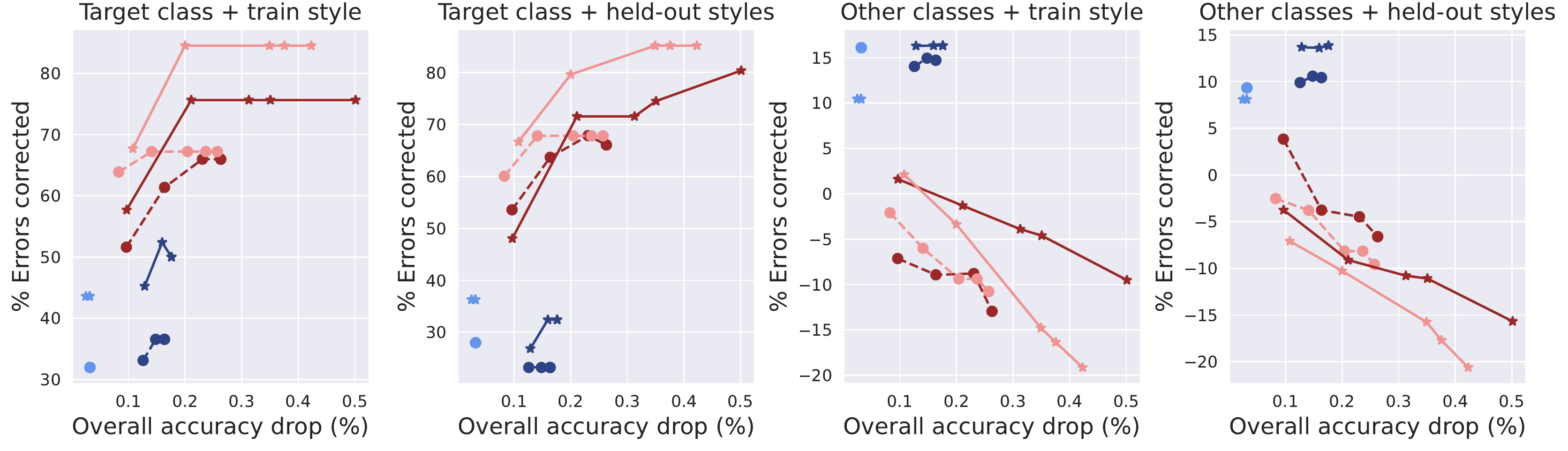}
		\caption{Concepts derived from an instance segmentation model trained 
	on LVIS.}
	\end{subfigure}
	\caption{Repeating the analysis in Appendix 
		Fig.~\ref{fig:app_tradeoff_editing_imagenet_vgg} on an 
		ImageNet-trained ResNet-50 classifier.}
	\label{fig:app_tradeoff_editing_imagenet_resnet}
\end{figure}

\begin{figure}[!h]
	\centering
	\begin{subfigure}[b]{0.9\textwidth}
		\centering
		\includegraphics[trim=0 0 0 30, clip, 
		width=1\columnwidth]{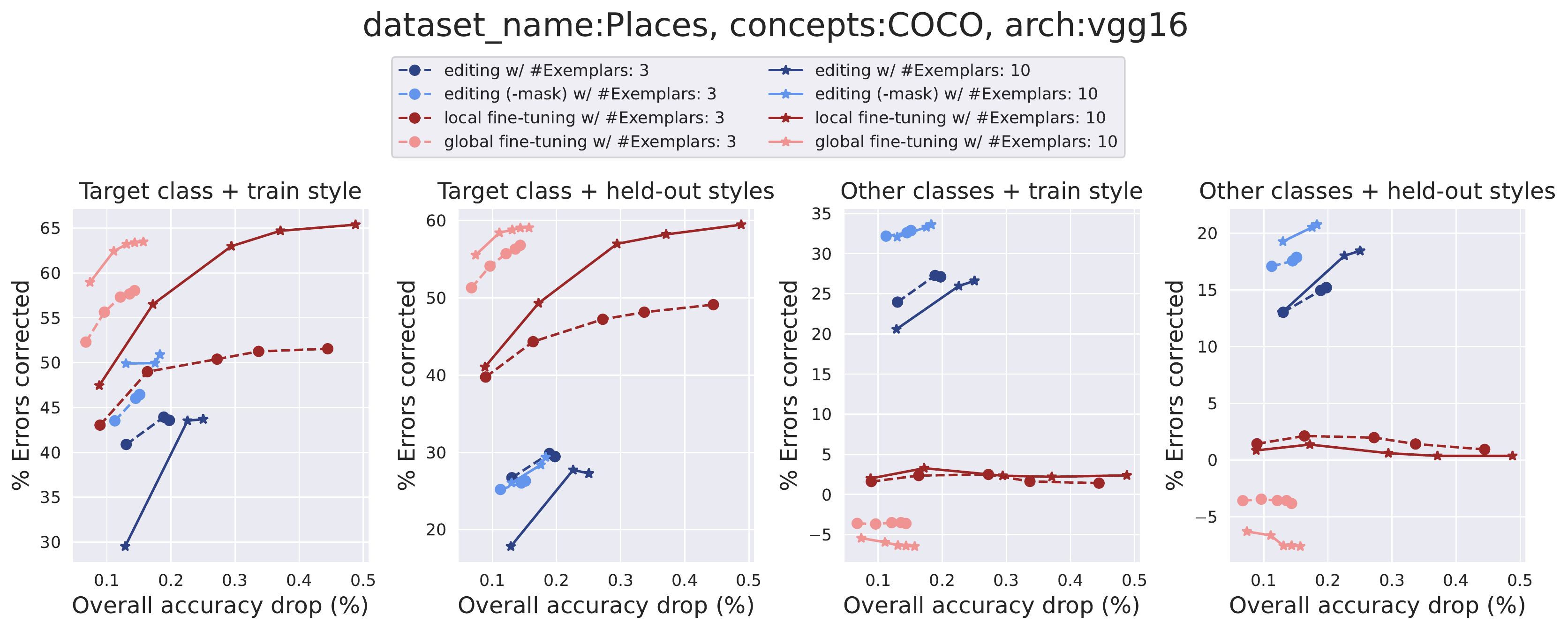}
		\caption{Concepts derived from an instance segmentation model trained 
	on MS-COCO.}
	\end{subfigure}
	\begin{subfigure}[b]{0.9\textwidth}
		\vspace{0.25cm}
		\centering
		\includegraphics[trim=0 0 0 110, clip, 
		width=1\columnwidth]{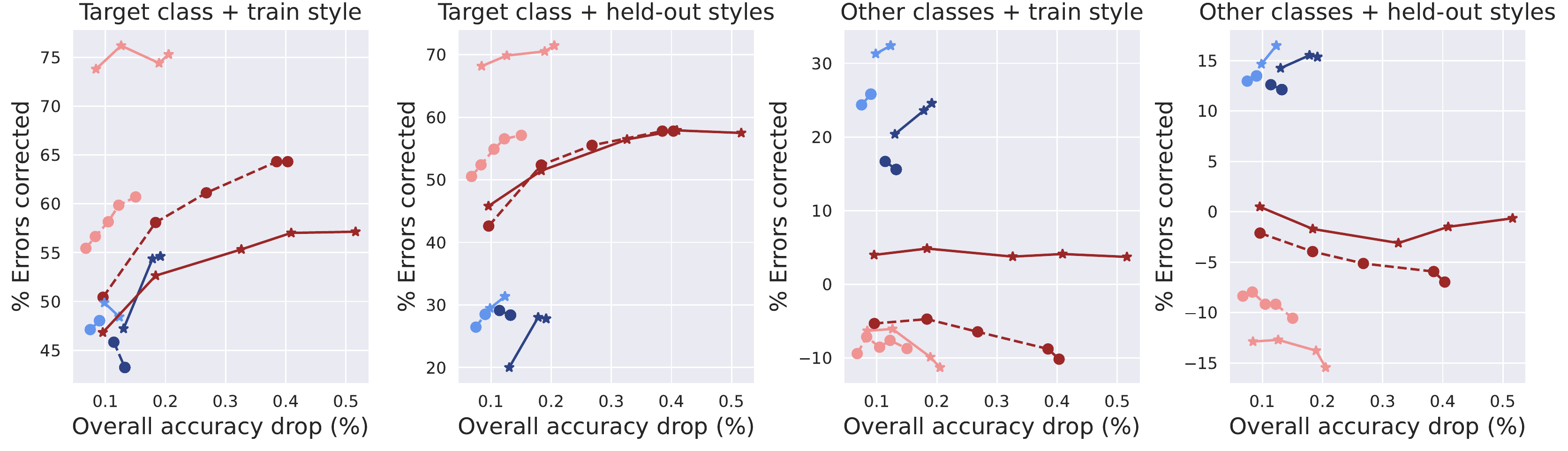}
		\caption{Concepts derived from an instance segmentation model trained 
	on LVIS.}
	\end{subfigure}
	\caption{Repeating the analysis in Appendix 
		Fig.~\ref{fig:app_tradeoff_editing_imagenet_vgg} on an  
		Places365-trained VGG16 classifier.}
	\label{fig:app_tradeoff_editing_places_vgg}
\end{figure}

\begin{figure}[!h]
	\centering
	\begin{subfigure}[b]{0.9\textwidth}
		\centering
		\includegraphics[trim=0 0 0 30, clip, 
		width=1\columnwidth]{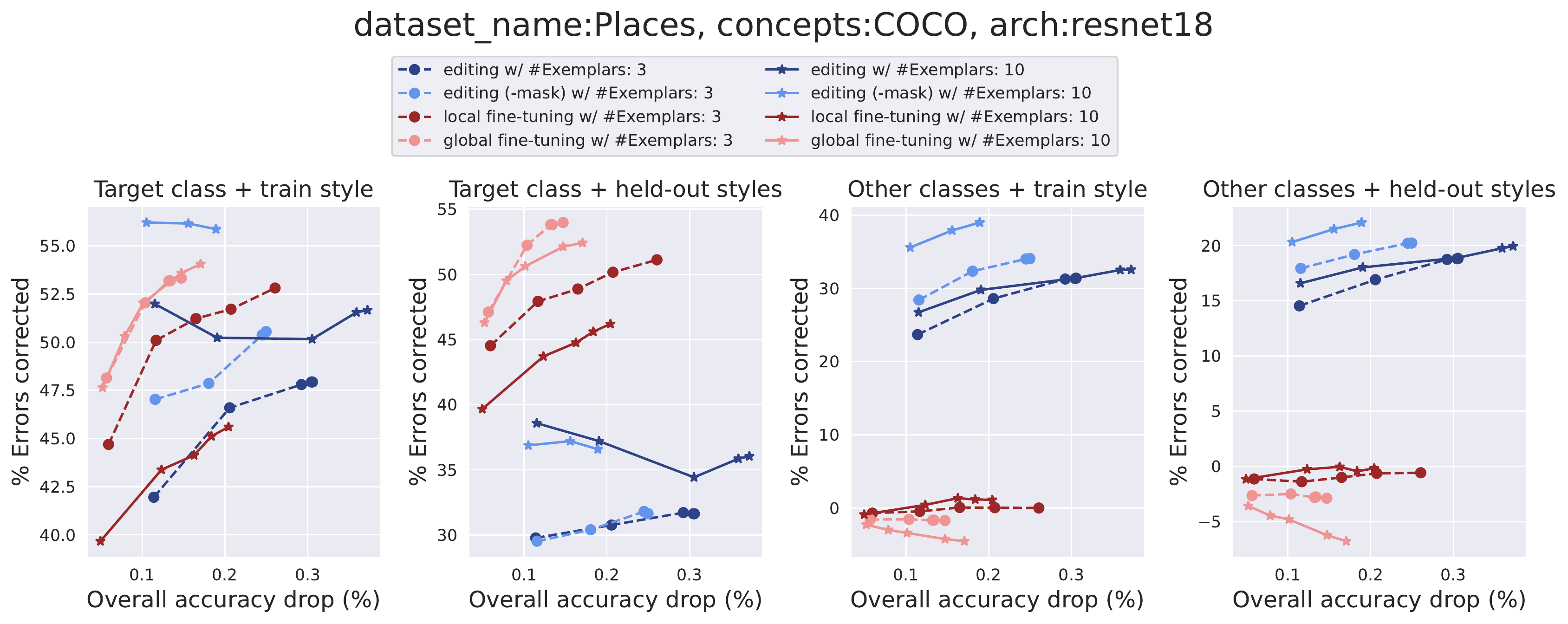}
		\caption{Concepts derived from an instance segmentation model trained 
	on MS-COCO.}
	\end{subfigure}
	\begin{subfigure}[b]{0.9\textwidth}
		\vspace{0.25cm}
		\centering
		\includegraphics[trim=0 0 0 110, clip, 
		width=1\columnwidth]{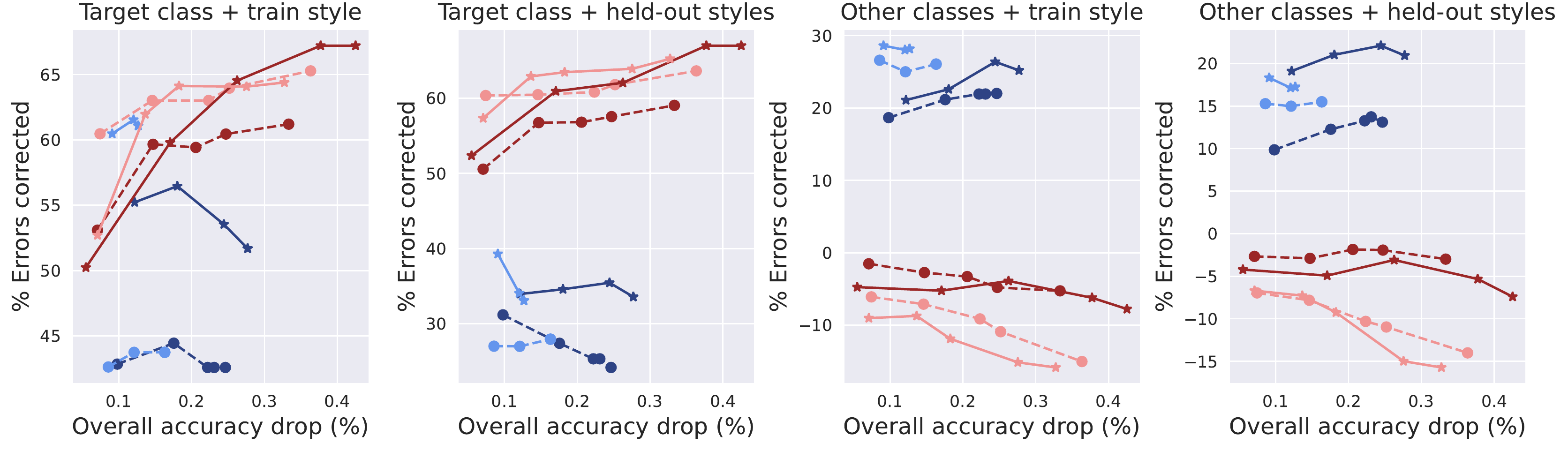}
		\caption{Concepts derived from an instance segmentation model trained 
	on LVIS.}
	\end{subfigure}
	\caption{Repeating the analysis in Appendix 
		Fig.~\ref{fig:app_tradeoff_editing_imagenet_vgg} on an 
		Places365-trained ResNet-18 classifier.}
	\label{fig:app_tradeoff_editing_places_resnet}
\end{figure}

\clearpage
\subsubsection{A fine-grained look at performance improvements}
In 
Appendix Figures~\ref{fig:app_per_concept} and \ref{fig:app_per_style} we 
take a closer look at the performance improvements caused by editing 
(-mask) and (local) 
fine-tuning (with 10 exemplars) on an ImageNet-trained VGG16 classifier.
In particular, we break down the improvements on test examples from 
\emph{non-target} classes with the same transformation as training, 
per-concept and per-style respectively.
As before, we only consider hyperparameters 
that lead to an overall accuracy drop of less 
than 
$0.25\%$.

\begin{figure}[!h]
	\centering
	\begin{subfigure}[b]{1\textwidth}
		\centering
		\includegraphics[width=1\columnwidth]{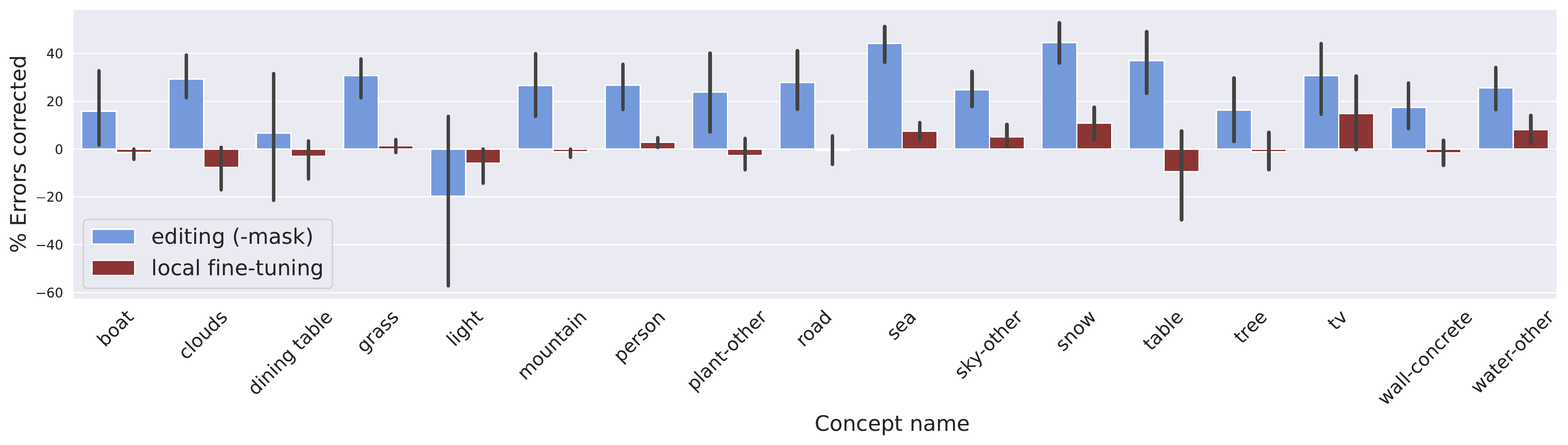}
		\caption{Concepts derived from an instance segmentation model trained 
	on MS-COCO.}
	\end{subfigure}
	\begin{subfigure}[b]{1\textwidth}
		\centering
		\includegraphics[width=1\columnwidth]{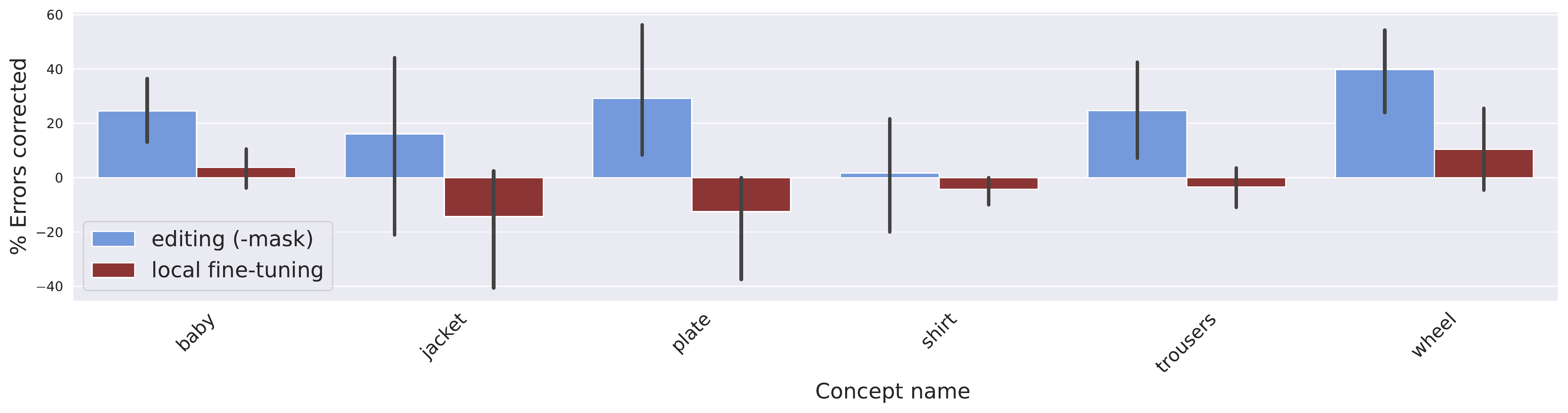}
		\caption{Concepts derived from an instance segmentation model trained 
	on LVIS.}
	\end{subfigure}
	\caption{Performance of editing and fine-tuning on 
	test examples from non-target classes containing a given concept, 
	averaged across transformations (cf. 
	Appendix~\ref{app:editing_setup}).}
	\label{fig:app_per_concept}
\end{figure}

\begin{figure}[!h]
	\centering
	\begin{subfigure}[b]{1\textwidth}
		\centering
		\includegraphics[width=1\columnwidth]{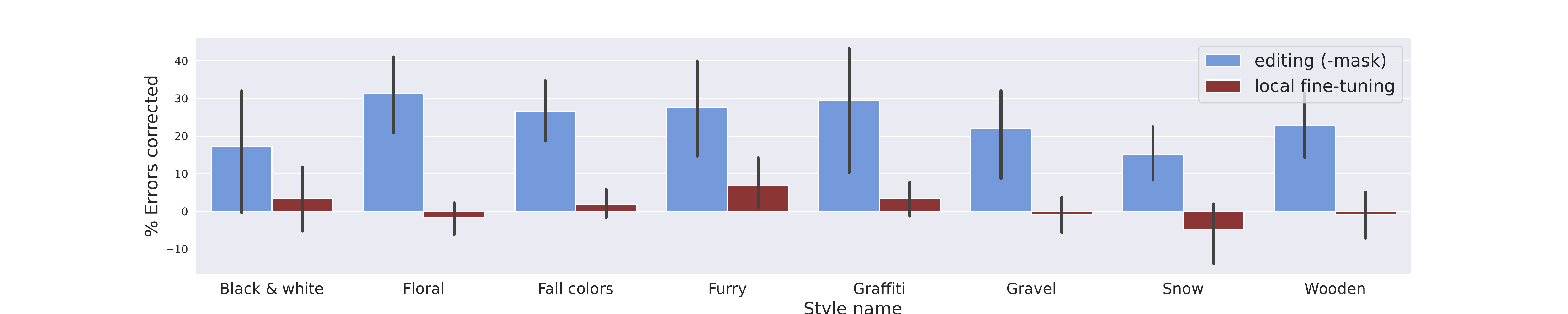}
		\caption{Concepts derived from an instance segmentation model trained 
	on MS-COCO.}
	\end{subfigure}
	\begin{subfigure}[b]{1\textwidth}
		\centering
		\includegraphics[width=1\columnwidth]{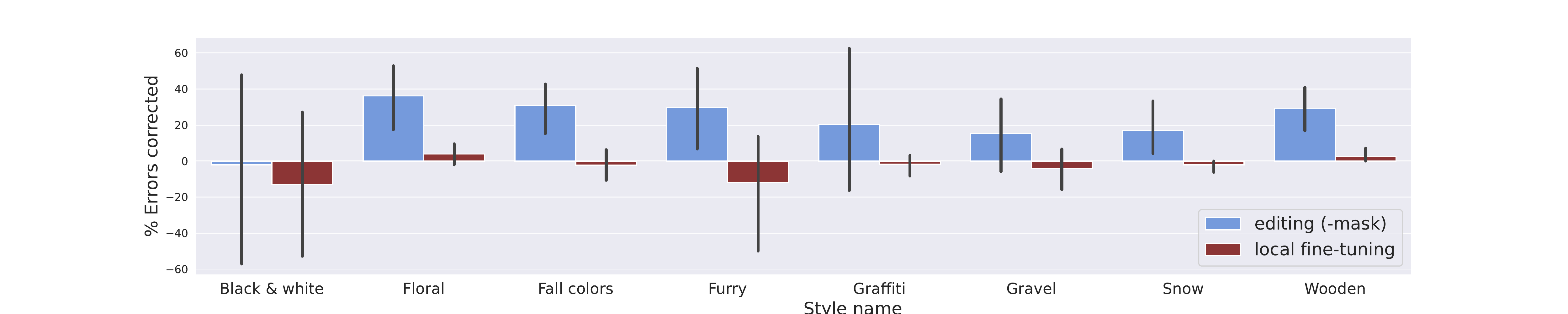}
		\caption{Concepts derived from an instance segmentation model trained 
	on LVIS.}
	\end{subfigure}
	\caption{Performance of editing and fine-tuning on 
	test examples from non-target classes transformed using a given style, 
	averaged across concepts.}
	\label{fig:app_per_style}
\end{figure}

\clearpage
\subsubsection{Ablations}
\label{app:ablation}
In order to get a better understanding of the core factors that affect
performance in this setting, we conduct a set of ablation studies.
Note that we can readily perform these ablations as, in contrast to the setting
of \citet{bau2020rewriting} we have access to a quantitative performance 
metric
that does not rely on human evaluation.
	
	\noindent \textbf{Layer.} We compare both editing and local 
	fine-tuning 
	when
	they are applied to different layers of the model in Appendix
	Figure~\ref{fig:app_layer_ablation}.
	For editing, we find a consistent increase in performance---on examples 
	from both the target and other classes---as we edit deeper
	into the model.
	For (local) fine-tuning, a similar trend is observed with regards to 
	performance on the target class, with the second last layer being optimal 
	overall.
	However, at the same time, the fine-tuned model's performance on 
	examples 
	from other classes 
	containing the concept seems to get worse. 
	
	\begin{figure}[!h]
		\centering
		\includegraphics[width=1\columnwidth]{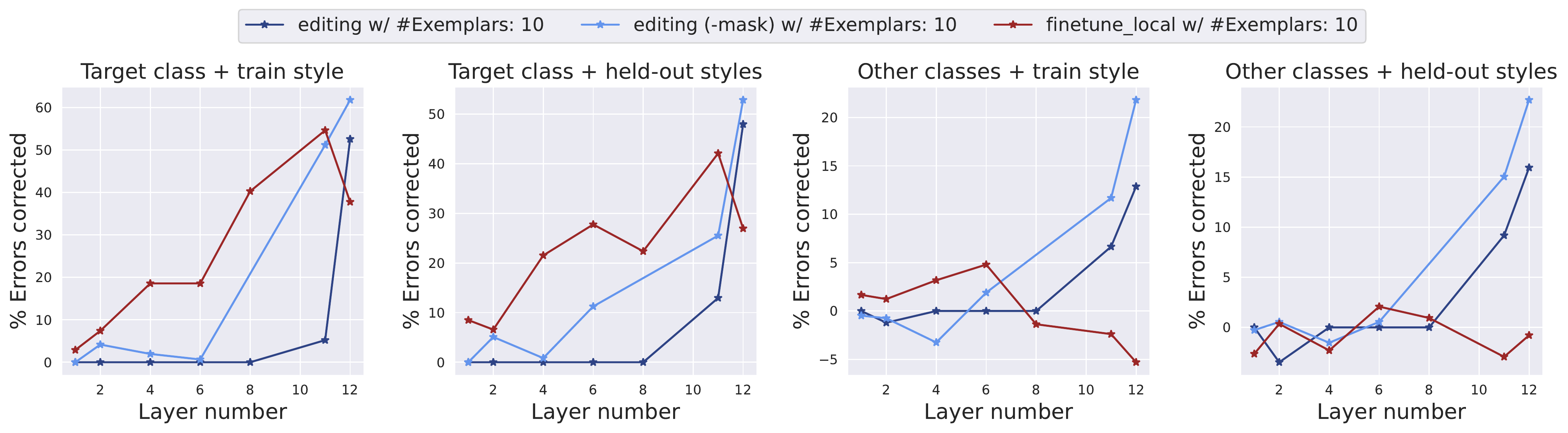}
		\caption{Editing vs. fine-tuning performance (with 10 exemplars) on an 
		ImageNet-trained 
		VGG16 classifier, as a function of the layer that is modified. Here, we 
		visualize the average number of 
			misclassifications corrected over different concept-transformation 
			pairs, with concepts derived from instance 
			segmentation modules 
			trained on MS-COCO; and transformations ``snow'' and ``graffiti''.
			For both editing and fine-tuning, the overall drop in model accuracy 
			is less 
			than 
			$0.25\%$.}
		\label{fig:app_layer_ablation}
	\end{figure}

\begin{figure}[!h]
	\centering
	\includegraphics[width=1\columnwidth]{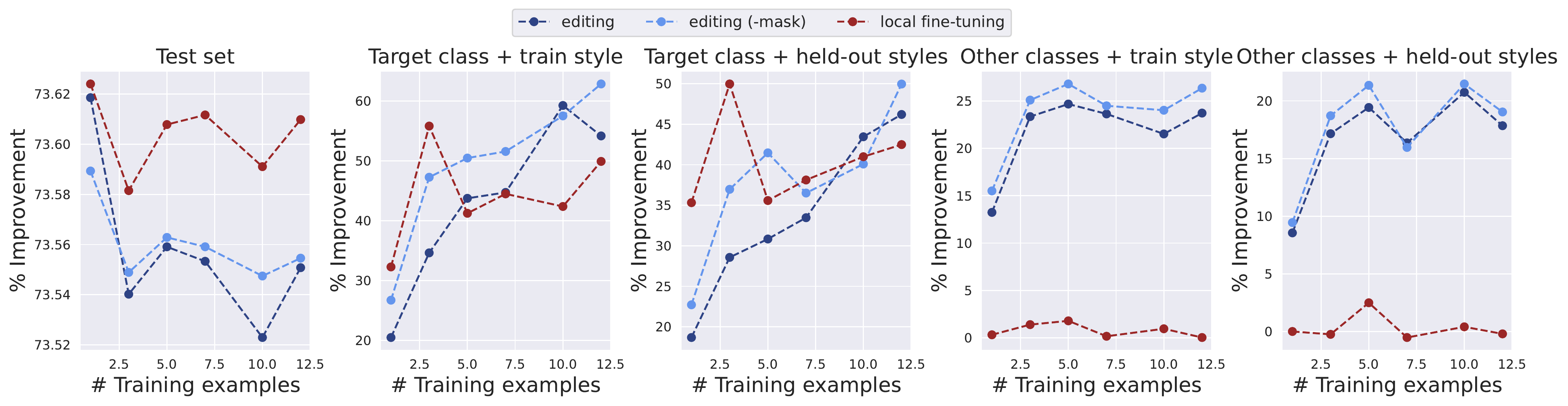}
	\caption{Editing vs. fine-tuning performance on an ImageNet-trained 
		VGG16 classifier, as a function of the number of train exemplars. Here, 
		we visualize the average number of 
		misclassifications corrected over different concept-transformation 
		pairs, with concepts derived from instance 
		segmentation modules 
		trained on MS-COCO; and transformations described in 
		Appendix~\ref{app:editing_setup}.
		For both editing and fine-tuning, the overall drop in model accuracy 
		is less 
		than 
		$0.25\%$.}
	\label{fig:app_n_ablation}
\end{figure}

	\noindent \textbf{Number of exemplars.}
	Increasing the number of exemplars used for each method typically leads 
	to 
	qualitatively the same impact, just more significant, cf. Appendix 
	Figures~\ref{fig:app_tradeoff_editing_imagenet_vgg}-\ref{fig:app_tradeoff_editing_places_resnet}.
	We also perform a more fine-grained ablation for a single model 
	(ImageNet-trained VGG16 on COCO-concepts) in 
	Figure~\ref{fig:app_n_ablation}.
	In general, for editing, using more exemplars tends to improve the number 
	of
	mistakes corrected on both the target and non-target classes.
	For fine-tuning, this improves its effectiveness on the target
	class alone, albeit the trends are more noisy.

\noindent \textbf{Rank restriction.}
We evaluate the performance of editing when the weight update is not
restricted to a rank-one modification.
We find that this change significantly reduces the efficacy of editing on 
examples from
both the target and non-target classes---cf. curves corresponding to `-proj' 
in  Appendix 
Figures~\ref{fig:app_ablation_editing_imagenet_vgg}-\ref{fig:app_ablation_editing_places_resnet}.
This suggests that the rank restriction is necessary to prevent the
model from overfitting to the few exemplars used.

\noindent \textbf{Mask.} During editing,
\citet{bau2020understanding} focus on rewriting only the key-value pairs
that correspond to the concept of interest.
We find, however, that imposing the editing constraints on the entirety of
the image leads to even better performance---cf. curves corresponding to 
`-mask' in Appendix 
Figures~\ref{fig:app_ablation_editing_imagenet_vgg}-\ref{fig:app_ablation_editing_places_resnet}.
We hypothesize that this has a regularizing effect as it constrains the 
weights to preserve the original mapping between keys and values in 
regions that do not contain the concept.

\begin{figure}[!h]
	\centering
	\begin{subfigure}[b]{0.9\textwidth}
		\centering
		\includegraphics[trim=0 0 0 30, clip, 
		width=1\columnwidth]{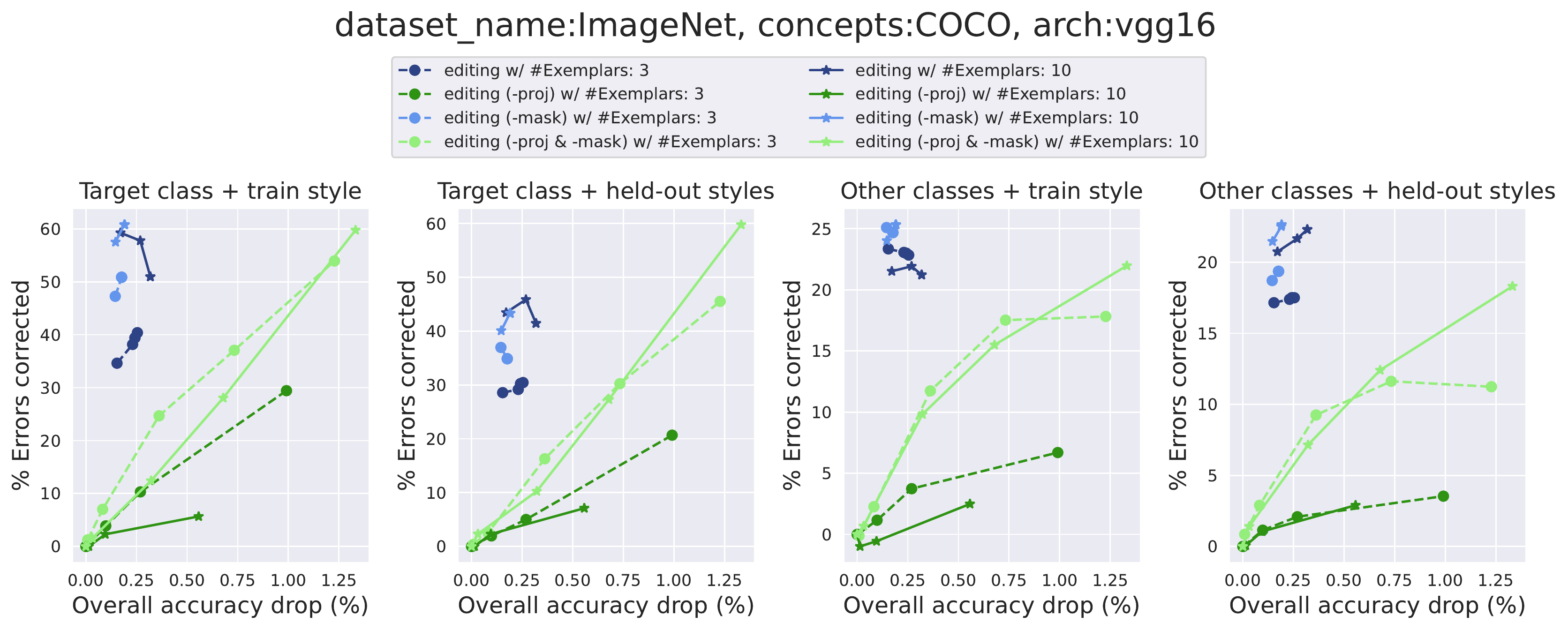}
				\caption{Concepts derived from an instance segmentation model 
				trained 
			on MS-COCO.}
	\end{subfigure}
	\begin{subfigure}[b]{0.9\textwidth}
		\vspace{0.25cm}
		\centering
		\includegraphics[trim=0 0 0 110, clip, 
		width=1\columnwidth]{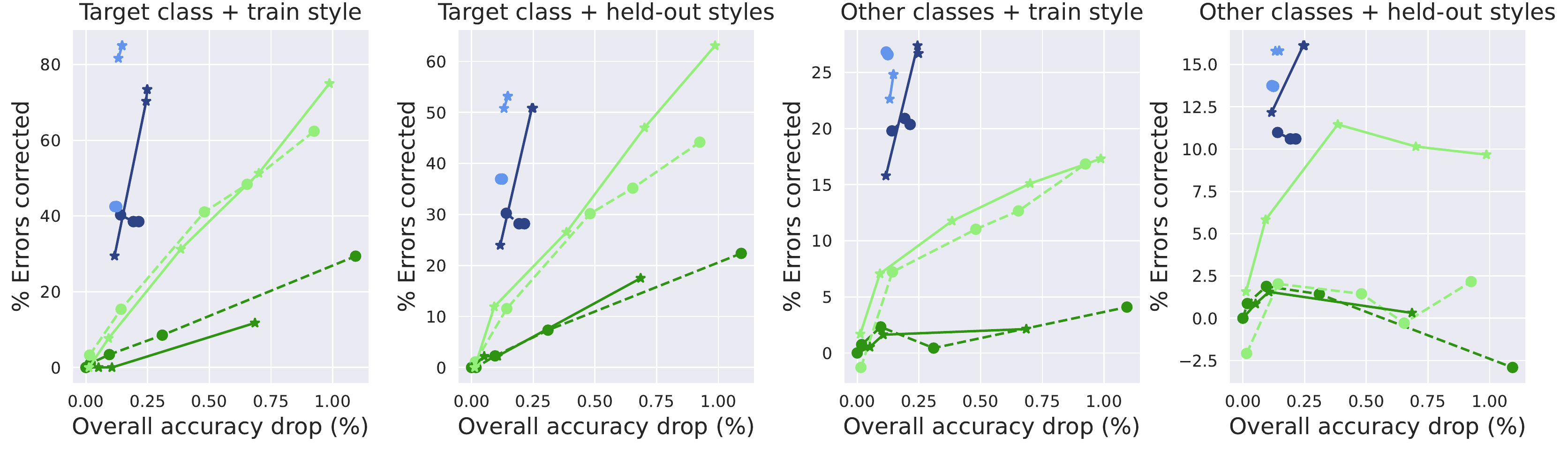}
			\caption{Concepts derived from an instance segmentation model 
			trained 
		on LVIS.}
	\end{subfigure}
	\caption{Performance vs. drop in 
		overall test set accuracy: Here, we visualize average number of 
		misclassifications corrected by editing variants---based on whether or 
		not we use a mask and perform a rank-one update---when applied to 
		an ImageNet-trained VGG16 classifier.  }
\label{fig:app_ablation_editing_imagenet_vgg}
\end{figure}

\begin{figure}[!h]
	\centering
	\begin{subfigure}[b]{0.9\textwidth}
		\centering
		\includegraphics[trim=0 0 0 30, clip, 
		width=1\columnwidth]{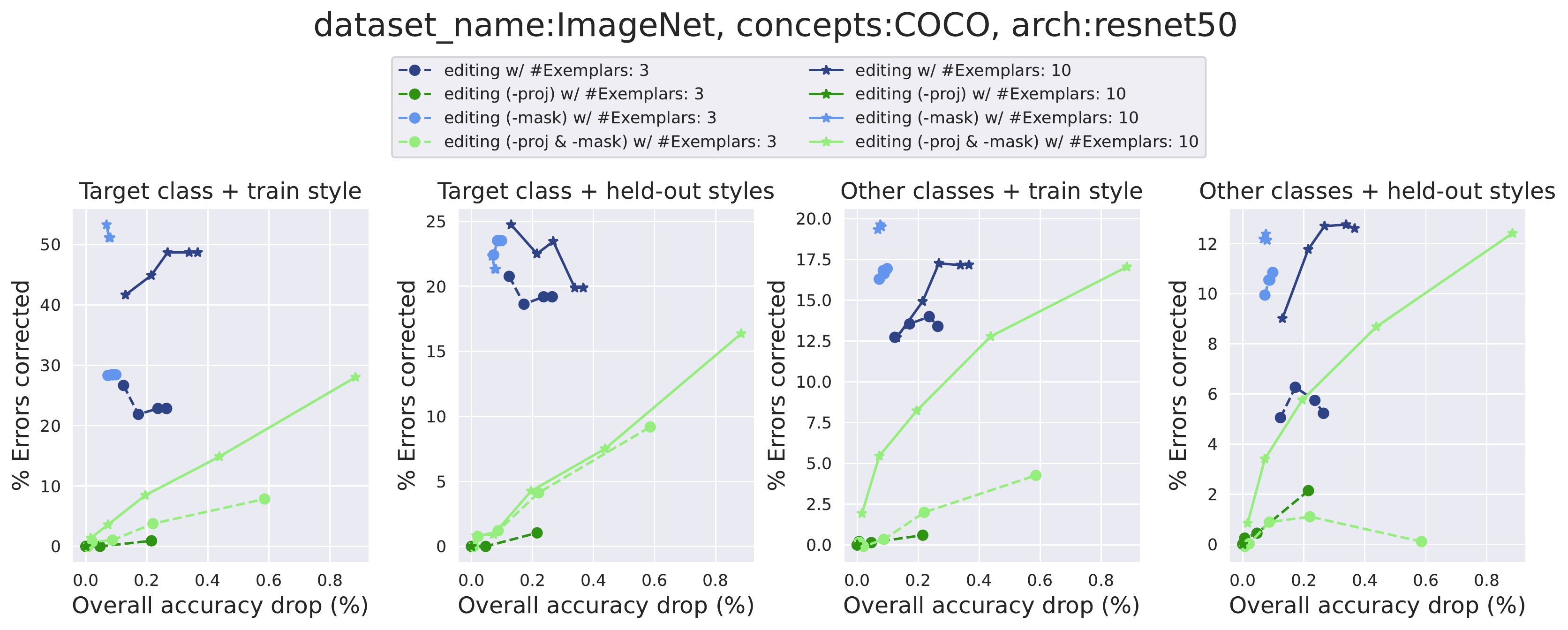}
				\caption{Concepts derived from an instance segmentation model 
		trained 
		on MS-COCO.}
	\end{subfigure}
	\begin{subfigure}[b]{0.9\textwidth}
		\vspace{0.25cm}
		\centering
		\includegraphics[trim=0 0 0 110, clip, 
		width=1\columnwidth]{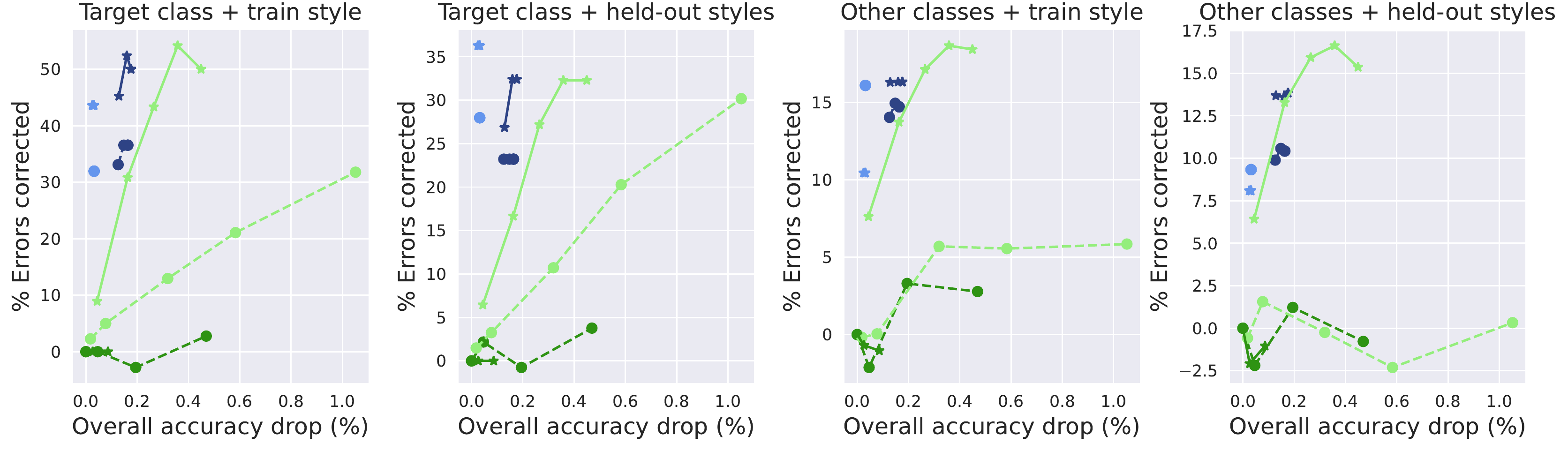}
				\caption{Concepts derived from an instance segmentation model 
		trained 
		on LVIS.}
	\end{subfigure}
	\caption{Repeating the analysis in Appendix 
	Fig.~\ref{fig:app_ablation_editing_imagenet_vgg} on
		an ImageNet-trained ResNet-50 classifier.}
	\label{fig:app_ablation_editing_imagenet_resnet}
\end{figure}

\begin{figure}[!h]
	\centering
	\begin{subfigure}[b]{0.9\textwidth}
		\centering
		\includegraphics[trim=0 0 0 30, clip, 
		width=1\columnwidth]{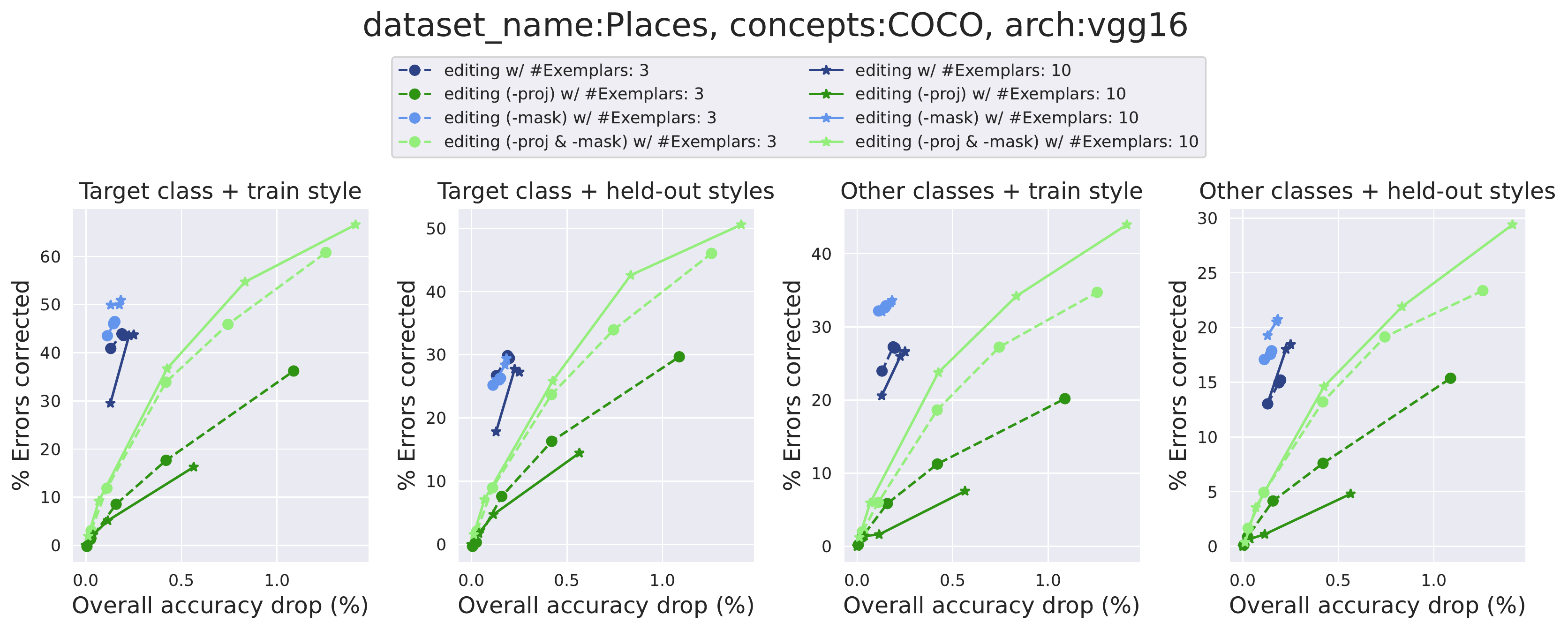}
				\caption{Concepts derived from an instance segmentation model 
		trained 
		on MS-COCO.}
	\end{subfigure}
	\begin{subfigure}[b]{0.9\textwidth}
		\vspace{0.25cm}
		\centering
		\includegraphics[trim=0 0 0 110, clip, 
		width=1\columnwidth]{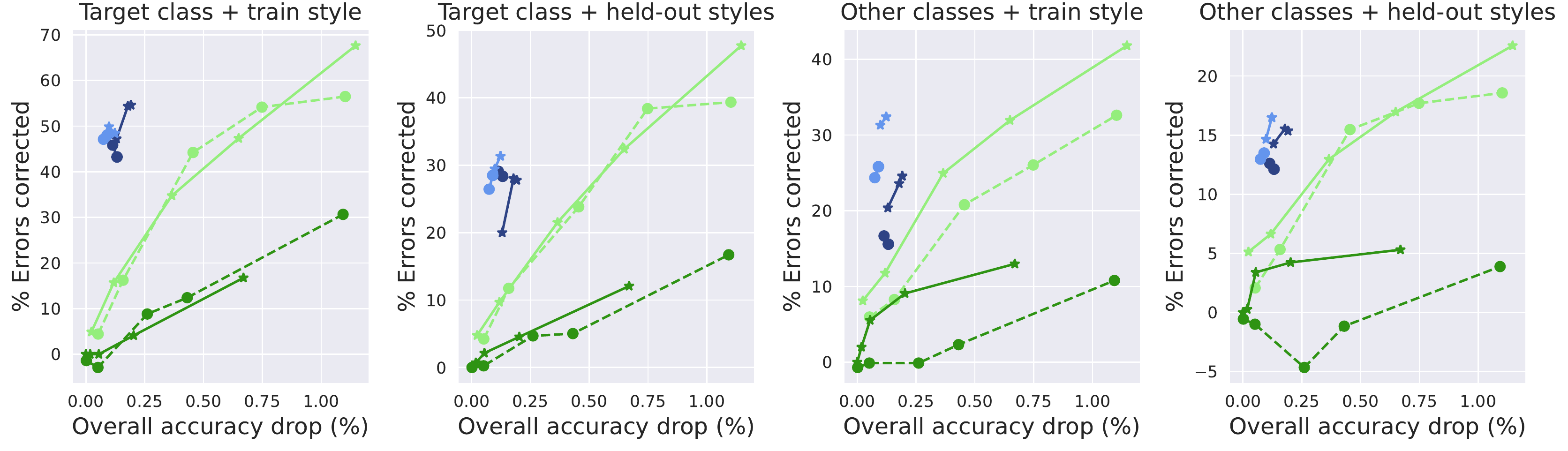}
				\caption{Concepts derived from an instance segmentation model 
		trained 
		on LVIS.}
	\end{subfigure}
	\caption{Repeating the analysis in Appendix 
		Fig.~\ref{fig:app_ablation_editing_imagenet_vgg} on
		an Places365-trained VGG16 classifier.}
	\label{fig:app_ablation_editing_places_vgg}
\end{figure}

\begin{figure}[!h]
	\centering
	\begin{subfigure}[b]{0.9\textwidth}
		\centering
		\includegraphics[trim=0 0 0 30, clip, 
		width=1\columnwidth]{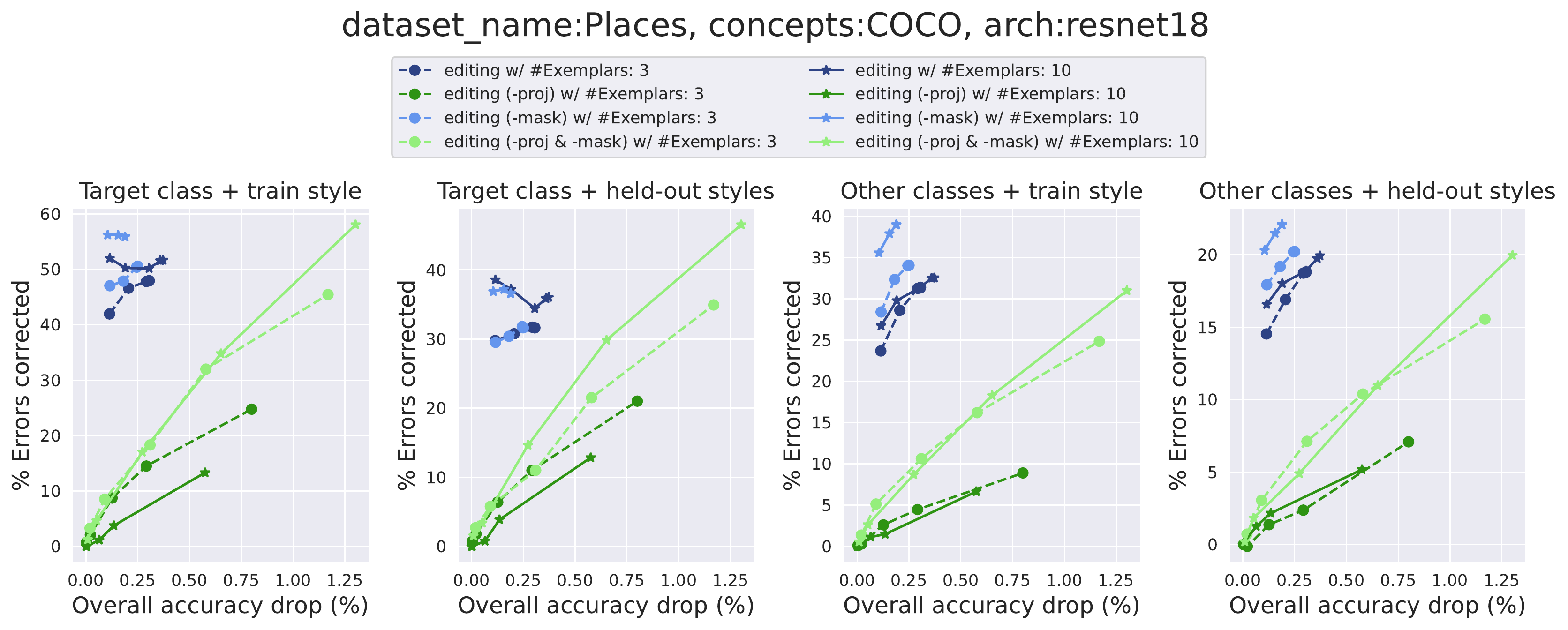}
		\caption{Concepts derived from an instance segmentation model 
		trained 
		on MS-COCO.}
	\end{subfigure}
	\begin{subfigure}[b]{0.9\textwidth}
		\vspace{0.25cm}
		\centering
		\includegraphics[trim=0 0 0 110, clip, 
		width=1\columnwidth]{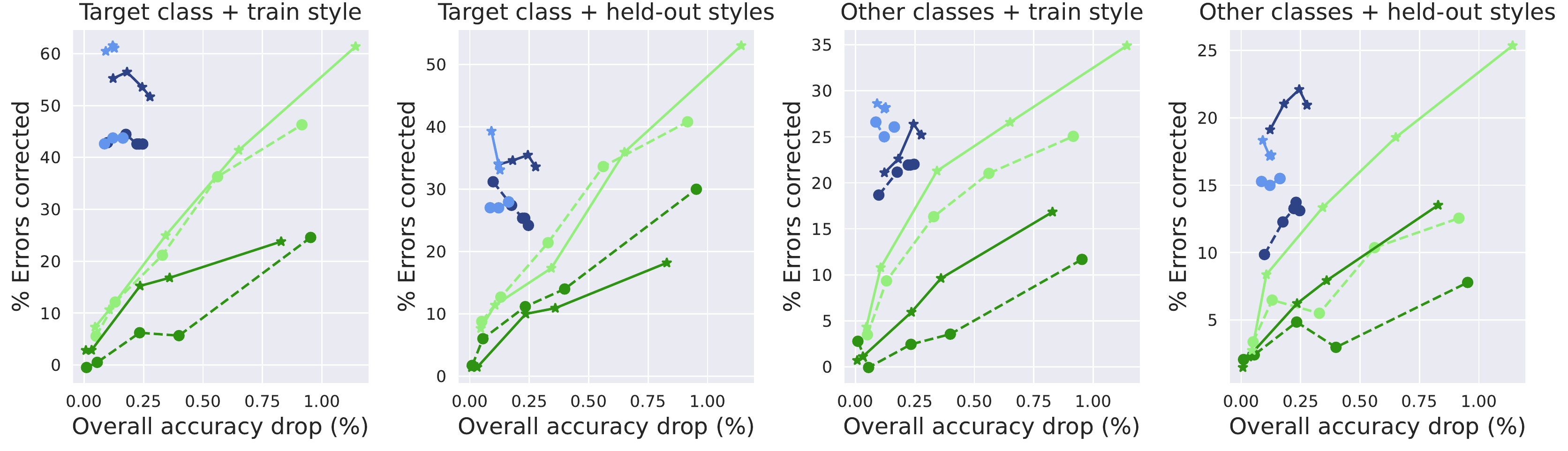}
				\caption{Concepts derived from an instance segmentation model 
		trained 
		on LVIS.}
	\end{subfigure}
	\caption{Repeating the analysis in Appendix 
		Fig.~\ref{fig:app_ablation_editing_imagenet_vgg} on
		an Places365-trained ResNet-18 classifier.}
	\label{fig:app_ablation_editing_places_resnet}
\end{figure}